%% file: aistats.tex
\documentclass[twoside]{article}

\usepackage[accepted]{aistats2024}
\usepackage[round]{natbib}

\usepackage{amsmath}
\usepackage{mathtools}
\usepackage{amssymb}
\usepackage[standard]{ntheorem}
\usepackage{graphicx}
\usepackage{comment}
\usepackage{tikz}
\usetikzlibrary{shapes.symbols}
\usetikzlibrary{fit}
\usetikzlibrary{backgrounds}
\usepackage{wrapfig}
\usepackage{csquotes}
\usepackage{caption}
\usepackage{subcaption}
\usepackage{marvosym}
\usepackage{xcolor, soul}
\usepackage{booktabs}
\usepackage{url}

\usepackage{soul} 
\usepackage{enumitem}
\setitemize{nosep, leftmargin=0.5cm}
\setenumerate{nosep, leftmargin=0.5cm}

\newtheorem{assumption}{Assumption}

\input{my_macros}

\input{tikz_input}

\usepackage{subfiles} 
\usepackage[noabbrev]{cleveref} 
\crefname{assumption}{assumption}{assumptions}
\begin{document}

%
\runningtitle{Self-Compatibility: Evaluating Causal Discovery without Ground Truth}

%
\runningauthor{Faller, Vankadara, Mastakouri, Locatello, Janzing}

\twocolumn[

\aistatstitle{Self-Compatibility: \\Evaluating Causal Discovery without Ground Truth}

\aistatsauthor{ Philipp M. Faller$^{1,2}$ \And Leena Chennuru Vankadara$^2$ \And Atalanti A. Mastakouri$^2$ \AND
	Francesco Locatello$^3$ \And Dominik Janzing$^2$}
\vspace{\baselineskip}
\aistatsaddress{$^1$Karlsruhe Institute of Technology, Germany\\  $^2$Amazon Research Tübingen, Germany \\ $^3$Institute of Science and Technology Austria } ]

\begin{abstract}
  As causal ground truth is incredibly rare, causal discovery algorithms are commonly only evaluated on simulated data.
  This is concerning, given that simulations reflect preconceptions about generating processes regarding noise distributions, model classes, and more.
  In this work, we propose a novel method for \textit{falsifying} the output of a causal discovery algorithm in the \textit{absence of ground truth}. Our key insight is that while statistical learning seeks stability across \textit{subsets of data points}, causal learning should seek stability across \textit{subsets of variables}. Motivated by this insight, our method relies on a notion of {compatibility} between causal graphs learned on different \textit{subsets of variables}.
  We prove that detecting incompatibilities can falsify wrongly inferred causal relations due to violation of assumptions or errors from finite sample effects. 
  Although passing such compatibility tests is only a necessary criterion for good performance, we argue that it provides strong evidence for the causal models whenever compatibility entails strong implications for the joint distribution.  
  We also demonstrate experimentally that detection of incompatibilities can aid in causal model selection.
\end{abstract}

\subfile{sections/intro}

\subfile{sections/compatibility}
\subfile{sections/score}

\subfile{sections/experiments}
\subfile{sections/related}

\section{CONCLUSION}
We have proposed a method 
to falsify the output of causal discovery algorithms which does not rely on causal ground truth.
It is based on compatibility of the output of an algorithm on different sets of variables. While this compatibility 
seems, at first glance, just as a weak sanity check, i.e. a weak necessary condition for providing good causal models, we have argued that there are cases where the compatibility requirement 
results in strong predictions for the joint distribution of 
variables which can be falsified from passive observations
(and thus provide strong evidence in favor of causal models if all attempts of falsification fail).

This approach is limited as we can only \emph{falsify} the outputs of causal discovery.
Even though we argue in \cref{sec:experiments} that our incompatibility score could be used as a proxy for SHD, we have no hard theoretical guarantees that ensure good performance for good scores (and we do not think that such guarantees can be proven without further assumptions).
Further, our work provides no guidance as to which degree of self-compatibility is \enquote{good enough} or when the outputs of an  algorithm should definitely be rejected. 

\subsubsection*{Acknowledgements}
Part of this work was done while Philipp M. Faller was an intern at Amazon Research Tübingen. 
Philipp M. Faller was supported by a doctoral scholarship of the Konrad Adenauer Foundation and the Studienstiftung des deutschen Volkes (German Academic Scholarship Foundation).

\bibliographystyle{abbrvnat}
\bibliography{aistats}

\section*{Checklist}

 \begin{enumerate}

 \item For all models and algorithms presented, check if you include:
 \begin{enumerate}
   \item A clear description of the mathematical setting, assumptions, algorithm, and/or model. Yes. 
   \item An analysis of the properties and complexity (time, space, sample size) of any algorithm. Yes.
   \item (Optional) Anonymized source code, with specification of all dependencies, including external libraries. Yes.
 \end{enumerate}

 \item For any theoretical claim, check if you include:
 \begin{enumerate}
   \item Statements of the full set of assumptions of all theoretical results. Yes.
   \item Complete proofs of all theoretical results. Yes.
   \item Clear explanations of any assumptions. Yes.    
 \end{enumerate}

 \item For all figures and tables that present empirical results, check if you include:
 \begin{enumerate}
   \item The code, data, and instructions needed to reproduce the main experimental results (either in the supplemental material or as a URL). Yes.
   \item All the training details (e.g., data splits, hyperparameters, how they were chosen). Yes.
         \item A clear definition of the specific measure or statistics and error bars (e.g., with respect to the random seed after running experiments multiple times). Yes.
         \item A description of the computing infrastructure used. (e.g., type of GPUs, internal cluster, or cloud provider). Yes.
 \end{enumerate}

 \item If you are using existing assets (e.g., code, data, models) or curating/releasing new assets, check if you include:
 \begin{enumerate}
   \item Citations of the creator If your work uses existing assets. Yes.
   \item The license information of the assets, if applicable. Not applicable.
   \item New assets either in the supplemental material or as a URL, if applicable. Yes.
   \item Information about consent from data providers/curators. Not applicable.
   \item Discussion of sensible content if applicable, e.g., personally identifiable information or offensive content. Not applicable.
 \end{enumerate}

 \item If you used crowdsourcing or conducted research with human subjects, check if you include:
 \begin{enumerate}
   \item The full text of instructions given to participants and screenshots. Not applicable.
   \item Descriptions of potential participant risks, with links to Institutional Review Board (IRB) approvals if applicable. Not applicable.
   \item The estimated hourly wage paid to participants and the total amount spent on participant compensation. Not applicable.
 \end{enumerate}

 \end{enumerate}
 
\renewcommand{\thesection}{A\arabic{section}} 
\subfile{sections/supplement}

\end{document}

%% file: my_macros.tex
\newcommand{\A}{\mathcal A}

\newcommand{\R}{{\mathbb R}}
\newcommand{\N}{{\mathbb N}}

\newcommand{\cA}{{\cal A}}

\newcommand{\cS}{\mathcal{S}}

\newcommand{\cG}{{\cal G}}

\newcommand{\cP}{{\cal P}}

\newcommand{\bX}{{\bf X}}

\newcommand\independent{\protect\mathpalette{\protect\independenT}{\perp}}
\def\independenT#1#2{\mathrel{\rlap{$#1#2$}\mkern2mu{#1#2}}}
\newcommand{\ind}{\independent}

\newtoggle{printcomments}
\togglefalse{printcomments} 

\newcommand\dominik[1]{%
  \iftoggle{printcomments}{%
    \textcolor{cyan}{Dominik: #1}%
  }{}%
}

\newcommand\leena[1]{%
  \iftoggle{printcomments}{%
\textcolor{pink}{Leena: #1} 
  }{}%
}

\newcommand\philipp[1]{%
  \iftoggle{printcomments}{%
\textcolor{red}{Philipp: #1} 
  }{}%
}

%% file: tikz_input.tex
\definecolor{lightgray}{gray}{0.85}
\sethlcolor{Dandelion}

\usetikzlibrary{arrows,shapes,plotmarks,positioning}
\usetikzlibrary{decorations.markings}

\tikzset{>=stealth'} 
\tikzset{node distance=2cm}
\tikzstyle{graphnode} = 
   [circle,draw=black,minimum size=22pt,text centered,text
     width=22pt,inner sep=0pt] 
\tikzstyle{var}   =[graphnode,fill=white]
\tikzstyle{vardashed}   =[graphnode,draw=gray,fill=white]
\tikzstyle{obs}   =[graphnode,fill=black,text=white]
\tikzstyle{obsgrey}   =[graphnode,draw=white,fill=lightgray,text=black]
\tikzstyle{par}    =[graphnode,draw=white,fill=red,text=black] 
 \tikzstyle{crucial} =[graphnode,draw=white,fill=yellow,text=black] 
\tikzstyle{fac}   =[rectangle,draw=black,fill=black!25,minimum size=5pt]
\tikzstyle{facprior} =[rectangle,draw=black,fill=black,text=white,minimum size=5pt]
\tikzstyle{edge}  =[draw=white,double=black,very thick,-]
\tikzstyle{blueedge}  =[draw=white,double=blue,very thick,-]
\tikzstyle{rededge}  =[draw=white,double=red,very thick,-]
\tikzstyle{prior} =[rectangle, draw=black, fill=black, minimum size=
5pt, inner sep=0pt]
\tikzstyle{dirprior} = [circle, draw=black, fill=black, minimum
size=5pt, inner sep=0pt]

\tikzstyle{dot_node}=[draw=black,fill=black,shape=circle]

%% file: sections/intro.tex
	\section{INTRODUCTION}
	
	Causal relationships are often formalized as directed acyclic graphs (DAGs) \citep{pearl2009causality}, or more general graphical models which also account for hidden confounders \citep{spirtes2000causation} and cycles \citep{bongers2021foundations}. Discovering causal relations is an important problem in science, which has led to the development of a diverse range of methods to infer causal graphs from passive observations. These methods are based on various approaches, such as Bayesian priors \citep{Heckerman1995b}, independence testing \citep{spirtes2000causation,lam2022greedy}, additive noise assumptions \citep{shimizu2006linear,hoyer2008nonlinear}, generalizations thereof \citep{Zhang_UAI, strobl2016estimating}, or various implementations of the so-called Independence of Mechanism assumption \citep{deterministic,Marx2017}. 
	
	 Causal discovery algorithms are typically evaluated primarily on
	simulated data. This is because causal ground truth is incredibly rare as it often necessitates real-world experiments. These experiments can be not only expensive and potentially unethical, but also frequently infeasible or even ill-defined from the outset \citep{spirtes2004}.
	Despite promising performance on simulated data, and in spite of numerous results on the 
	identifiability of causal DAGs or parts thereof from passive observations, skepticism about the applicability of these algorithms on real data is warranted. This is because these algorithms are built on assumptions such as faithfulness, additive noise, post-nonlinear models, or independence principles, all of which can be violated in practice. Indeed, recent studies on causal discovery methods reveal a disconcerting reality about their applicability to real-world datasets 
	\citep{huang2021benchmarking,Reisach2021}.
	Accordingly, the value of existing algorithms for downstream causal inference tasks
	is unclear and debated \citep{Imbens2020}. 
	
	{In this paper we propose a novel methodology for the falsification of the outputs of  causal discovery algorithms on real data {\it without access to causal ground truth}. The key idea involves testing the \textit{compatibility} of the causal models inferred by the algorithm when applied to different subsets of variables.  In essence, while statistical learning aims for stability across different subsets of data points, we argue that causal discovery should aim to achieve stability across different subsets of variables. We prove that checking for incompatibilities provides a means of falsifying the outputs of causal discovery, as these incompatibilities indicate either violated causal discovery assumptions or finite sample effects that lead to non-negligible changes in the algorithm's output.
	}
	
	It is natural to ask if one can trust an algorithm if it satisfies such compatibility constraints. To address this question, we align with the theory of science by \citet{Popper1959}, according to which a hypothesis gathers evidence when numerous attempts to falsify it fail. We argue that, under a sufficiently strong notion of compatibility, the algorithm's outputs on various subsets of variables can entail strong implications for the joint distribution, thereby offering numerous opportunities for falsification.
	
	\paragraph{Our contributions.}
	
	In this work we present a novel framework to evaluate causal graphs in the absence of ground truth. Specifically, \leena{We need to make the contributions more concrete and sell our paper more! I will make concrete suggestions once I go through the remaining sections.}
	\begin{itemize}
		\setlength{\itemsep}{0.4em}
		\item we introduce two different notions of compatibility: interventional and graphical (Section \ref{sec:compatibility_notions}). \dominik{these are rather types of compatibility notions, they still admit room for quite different notions} Under these definitions, we prove that if the assumptions of a causal discovery algorithm are met, its outputs are compatible in the population limit. Furthermore, we show for existing algorithms that they admit falsification using this approach (Section \ref{sec:falsification}).
		
		\item We connect compatibility to the causal marginal problem and argue that compatibility can entail strong implications for potential joint distributions of given marginals
        , which can then be falsified statistically 
        (Section \ref{subsec:compatibility_strong}).

		\item We introduce the incompatibility score for causal discovery which quantifies the \textit{level of incompatibility} of the outputs of causal discovery. \dominik{I'm wondering whether we should be upfront here or later about the fact that finding a good score is non-trivial and that ours is not supposed to be the final word}\philipp{I'll add it later as this section is already long.}  We argue based on stability arguments from learning-theory that the incompatibility score could serve as a proxy for measures like structural Hamming distance (SHD) which require access to ground truth (Section \ref{sec:score}). 
		
		\item We demonstrate that our score can potentially be used for model selection in simulation studies and on real-world data where ground truth knowledge is available. Our results show a significant correlation between the score and SHD (Section \ref{sec:experiments}).
		
	\end{itemize}
	
	\subsection{Motivational Example}
	\label{subsec:motivating_example}
	
	\begin{figure}[htp]
		\centering
				\begin{subfigure}[c]{\columnwidth}
			\centering
			\begin{tikzpicture}
				\node[obs,circle] (x) {$X$};
				\node[obs,circle,below of=x] (y) {$Y$};
				\node[obs,circle,above left of=y] (z1) {$Z_1$};
				\node[obs,circle,above right of=y] (z2) {$Z_2$};
				
				\draw (x) edge [->] (y);
				\draw (y) edge[<-] (z1);
				\draw (y)edge[<-] (z2);
				\draw (x) edge[->] (z2);
				
				\begin{pgfonlayer}{background}
					\node[blue!80, left=of y, xshift=1.0cm] (s2) {$\mathbf{S}$};
					\node[teal!80,  right=of y, xshift=-1.0cm] (t2) {$\mathbf{T}$};
					\node[fit=(x) (y) (s2),draw,blue,dashed,inner sep=5pt,fill=blue!20,fill opacity=0.3,rounded corners] {};
					\node[fit=(x) (y) (t2),draw,teal,dashed,inner sep=5pt,fill=teal!20,fill opacity=0.3,rounded corners] {};
				\end{pgfonlayer}
				
			\end{tikzpicture}
			\caption{Compatible DAGs over $S$ and $T$ that admit a joint DAG over $S\cup T$.}
			\label{fig:motivation_SandT}
		\end{subfigure}
		\hfill
            \begin{subfigure}[c]{0.45\textwidth}
			\centering
			\begin{tikzpicture}
				\node[obs,circle] (x) {$X$};
				\node[obs,circle,below of=x] (y) {$Y$};
				\node[obs,circle,above left of=y] (z1) {$Z_1$};
				
				\draw (x) edge [->] (y);
				\draw (y) edge[<-] (z1);
				
				\begin{pgfonlayer}{background}
					\node[blue!80, left=of y, xshift=1.0cm] (s1) {$\mathbf{S}$};
					\node[fit=(x) (y) (s1),draw,blue,dashed,inner sep=5pt,fill=blue!20,fill opacity=0.3,rounded corners] {};
				\end{pgfonlayer}
				
				\node[below right of=x](lightning){\huge\Lightning};
				\node[obs,circle, above right of=lightning] (x1) {$X$};
				\node[obs,circle,below of=x1] (y1) {$Y$};
				\node[obs,circle, above right of=y1] (z2) {$Z_2$};
				
				\draw (x1) edge [<-] (y1);
				\draw (x1) edge[<-] (z2);
				
				\begin{pgfonlayer}{background}
					
					\node[red!80, right=of y1, xshift=-1.cm] (t1) {$\mathbf{T}$};
					\node[fit=(x1) (y1) (t1),draw,red,dashed,inner sep=5pt,fill=red!20,fill opacity=0.3,rounded corners] {};
				\end{pgfonlayer}
			\end{tikzpicture}
			\caption{Incompatible DAGs over $S$ and $T$}
			\label{fig:motivation_SandT_false}
		\end{subfigure}
		
		\caption{Each marginal causal models over $S$ and $T$ graphical implies a constraint for the edge $X-Y$ as it can only be directed in one way.
		}
		\label{fig:graph_v_structure_wrong}
	\end{figure}
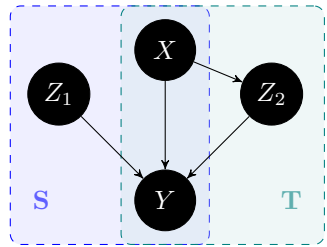
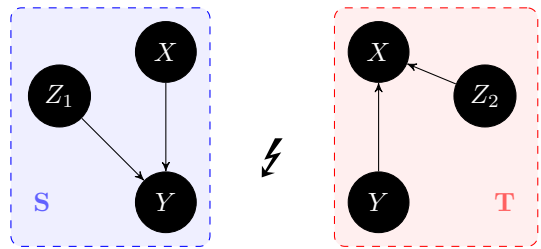

To illustrate our key ideas, we describe a simple example where a causal discovery algorithm results in causal graphs  on different subsets of variables which are incompatible in a sense that we will explain now. 
		Consider the directed acyclic graph (DAG) in 
		\cref{fig:motivation_SandT}.
		Assume we are given the two subsets $S=\{X, Y, Z_1\}$ and $T=\{X, Y, Z_2\}$.
		We will call a causal model that represents only a subset of all relevant variables a \emph{marginal causal model} in analogy to the marginal distribution.
		Assume the PC algorithm\footnote{For simplicity, we use the popular PC algorithm in this example and therefore implicitly assume that $S$ and $T$ are causally sufficient \citep{pearl2009causality} sets of variables. 
			A similar construction can be made for the FCI algorithm \citep{spirtes2000causation}, which does not assume sufficiency (see \cref{sec:proofs} in the appendix). } 
    \citep{spirtes2000causation} is applied to samples from the marginal distributions over $S$ and $T$.
		Suppose the observed distribution violates faithfulness 
		by satisfying $Y\ind Z_2$
		but otherwise only 
		(conditional) independences hold if they are required by the Markov condition.
		PC then outputs the two DAGs in \cref{fig:motivation_SandT_false}, which
		entail the interventional statements:
		$p(y|do(x)) =p(y|x)$ (for $S$) versus $p(y|do(x))=p(y)$ (for $T$). 
		This is a contradiction unless $Y\independent X$, which would be an additional violation of faithfulness in contrast to our assumption. Hence, 
		the outputs of PC cannot both be correct given the observed joint distribution. 
		Perhaps more strikingly, the marginal models in \cref{fig:motivation_SandT_false} have different orientations for the edge between $X$ and $Y$.
		Accordingly, \cref{sec:compatibility_notions} defines an \emph{interventional} and a \emph{graphical} notion of compatibility. Note, that an ordinary statistical cross-validation would 
		not have discovered this inconsistency since
		we assumed $Y\ind Z_2$ to hold in the population case, rather than assuming that a statistical test accepted independence due to a type II error.

%% file: sections/compatibility.tex
\section{COMPATIBILITY OF CAUSAL GRAPHS}
\label{sec:compatibility}
\label{sec:causal_cross_validation}

\paragraph{Notation.}
Throughout this paper we will use $[n] := \{1, \dots, n\}$ for $n\in\N$, and for
$i\in [n]$ we denote random variables with $X_i$ and values of a variable with the respective lower case letter $x_i$.
By slight abuse of notation we denote a vector of random variables in a set $S$ by $X_S$ and a vector of values of these variables with $x_S$.
We denote the matrix containing $m\in \N$ values of all variables in a set $S$ with $\bX^m_S$.
We sometimes omit $m$. 
$\bX$ denotes the matrix containing vectors of values for all 
variables $X_i$ for $i\in [n]$. 
For a probability distribution $P$ over a set of variables $V$ and $S\subset V$ we denote with $P_S$ the marginal distribution over $S$ and $p$ be the probability density function of $P$ where we assume for simplicity that there always exists a density with respect to a product measure.

\paragraph{Causal models.}
\looseness=-1Although our proposed compatibility-based evaluation is in principle not restricted to any kind of causal model, we will focus our exposition on graphical models.
Precisely, we will use acyclic directed mixed graphs (ADMGs) \citep{richardson2003markov,evans2014markovian}, partial ancestral graphs (PAGs) \citep{zhang2008causal} and completed partially directed acyclic graphs (CPDAGs) \citep{spirtes2000causation}. 
In the main paper we focus on ADMGs, which include DAGs as special cases.
Formal definitions of the remaining graphical models and of the causal semantics of graphical models can be found in \cref{sec:formal}. Sections with the prefix \emph{A} are in the appendix.

\begin{definition}[ADMG]
	\label{def:admg_main}
	A \emph{mixed graph} $G$ consists of a finite set of nodes $V$ and a set of directed edges $E\subseteq V\times V$ as well as a set of bidirected edges $B\subseteq\{\{X_i, X_j\} : X_i, X_j \in V\}$. 
	If $(X_i, X_j)\in E$ we say there is a \emph{directed edge} between $X_i$ and $X_j$ and we write $X_i\to X_j$.
	If $\{X_i, X_j\}\in B$ we say there is a \emph{bidirected edge} and write $X_i\leftrightarrow X_j$. 
 Directed and bidirected edges can occur together. 
    A sequence of $l\in \N$ nodes $X_{i_1}, \dots, X_{i_l}\in V^l$ with any edge between $X_{i_j}$ and $X_{i_{j+1}}$ for $j\in [l]$ is called an \emph{undirected path} and it is called a \emph{directed path} if all edges are in $E$ and point towards the same direction.
	A mixed graph is called \emph{acyclic directed mixed graph} (ADMG) if there is no directed path from $X_i$ to itself for all $X_i\in V$.
	An ADMG with no bidirected edges is called \emph{directed acyclic graph} (DAG).
\end{definition}

\begin{definition}[causal discovery algorithm]
	For the purpose of this paper, a \emph{causal discovery algorithm} $\cA$ takes i.i.d. data as input, i.e. a matrix $\bX_{S}$ 
	with $S\subseteq V$ containing samples from $P_S$, and outputs a ADMG, CPDAG or PAG over nodes in $S$, denoted by $G_S$ or a special symbol $\bot$ to indicate that the algorithm itself detected a violation of assumptions.
\footnote{In fact there are few algorithms that can output such a token. E.g. \citet{ramsey2012adjacency} proposed an algorithm that can detect some violated assumptions itself. 
We will use the token in the proof of \cref{thm:idealized_rcd}.}
\end{definition}

Throughout our work, we assume that all data has been generated by a single causal model.
\begin{assumption}[existence of joint model]
	\label{assumption:joint_model}
	Whenever we consider $k\in\N$ sets of variables $S_i$ with $i\in [k]$ and distributions $P_{S_i}$ we assume there is set $V$ and a DAG
	$G = (V, E)$ such that there is a distribution $P_V$ where $G$ is the causal graph\footnote{Note that the requirement that this data generation is formalized as a DAG is not a hard restriction, as CPDAGs, ADMGs, MAGs and PAGs naturally correspond to DAGs (when some variables of the DAG are unobserved).}  of $P_V$ and for all $i\in [k]$ we have $S_i\subseteq V$ and $P_V$ has $P_{S_i}$ as marginal distributions over $S_i$.
\end{assumption}

\subsection{Notions of Compatibility}
\label{sec:compatibility_notions}

We now introduce the concept of 
\textit{compatibility} between causal graphs. We will discuss later how we can 
use  compatibility of the outputs of a causal discovery algorithm (i.e. the \emph{self-compatibility}) to falsify these outputs. 

\begin{definition}[compatibility notion]
	\label{def:abstract:compatibility}
	Let $(\cG_V\cup \{\bot\})^*$ be the space of tuples\footnote{I.e. for any set $A$ define $A^1 = A, A^{i+1} = A^{i} \times A$ for $i\in\N$ and $A^* = \cup_{i\in\N} A^i$.} of the special token $\bot$ or graphs of some type (DAGs, CPDAGs, ADMGs, MAGs, PAGs)
 over subsets of a set $V$.
	Let $\cP_V$ denote the space of probability distributions over $V$.
	A \emph{compatibility notion} is a function
	\[
	c: (\cG_V\cup \{\bot\})^* \times \cP_V \to \{0,1\}.    
	\]
	For $k\in \N$ and $S_1, \dots, S_k\subseteq V$, the graphs $G_{S_1},\dots,G_{S_k}$ are called compatible with respect to $c$ and $P_V$ if 
	$c(G_{S_1},\dots,G_{S_k},P_V) =1$.
	
\end{definition} 
In this work we will discuss two compatibility notions.
We define an \emph{interventional} compatibility notion, as we consider it a natural requirement of a causal discovery algorithm to make contradiction-free interventional statements. 
We also define a \emph{graphical} notion of compatibility that has the advantage that it does not involve statistical decisions (as it does not directly depend on the distribution), but it raises conceptual problems due to implicit genericity assumptions that we discuss in \cref{subsec:critical_graphical_comp}.
We want to emphasize that other notions of compatibility (both, interventional and graphical) are conceivable and might be more appropriate in some situations.

\begin{definition}[interventional compatibility] 
	\label{def:interventional_comp}
	
	Let $S_1, \dots S_k$ be sets of variables for some $k\in\N$ and denote $S:= \bigcup_{i\in[k]} S_i$. 
	Let $G_{S_1}, \dots, G_{S_k}$	be causal models (DAGs, CPDAGs, ADMGs, MAGs, PAGs) and $P_S$ be a probability distribution over $S$. 
	Define the \emph{interventional compatibility} via $c(G_{S_1}, \dots, G_{S_k}, P_S) = 1$
	iff there exists a superset of nodes $V\supseteq S$, a DAG $G_V$ over the variables in $V$ and a distribution $P_V$ such that 
	\begin{enumerate}
		\item $P_V$ has $P_S$ as marginal over $S$,
		\item $P_V$ is Markovian w.r.t. to $G_V$ and
		\item for any $X_i, X_j\in S_l$ with $l\in[k]$ and any identification formula\footnote{As we will describe in more detail in \cref{sec:formal}, an interventional probability can often be expressed by different observational terms. These symbolic terms are a priori only related to the graph, but map a  distribution to a probability. E.g. in \cref{fig:motivation_SandT} $p(y|do(x))$ can be identified by the backdoor formula either  using $\emptyset$ and therefore $p\mapsto p(y|x)$ or with adjustment set $\{Z_1\}$ we get $p\mapsto \int p(y| x, z_1)p(z_1)\,dz_1 $ .} 
		 in $G_{S_l}$ holds:
		if the intervention $p(x_i | do(x_j))$ is identifiable in $G_{S_l}$, then the interventional probabilities coincide with the interventional probabilities $p^{G_V}(x_i | do(x_j))$ derived from $G_V$.
	\end{enumerate}
    Further set $c(\dots, \bot, \dots) = 0$ regardless of the other arguments.
\end{definition}
Interventional compatibility requires that the different causal graphs entail interventional probabilities that could come from a single joint causal model.
Especially, this compatibility is with respect to a specific distribution. 
The second notion of compatibility we discuss in this work is the notion of graphical compatibility.
For simplicity, we will only define graphical compatibility for ADMGs in the main paper, but in \cref{sec:graphical_comp} we will propose detailed definitions for the other graphical models mentioned in this work.

\looseness=1 To define graphical compatibility, we first discuss what \citet{pearl1995theory} called the \emph{latent projection} of a graphical model.
	Precisely,  we will define the latent projection of an ADMG like \citet{richardson2023nested}. 
	\begin{definition}[latent ADMG]
		\label{def:latent_projection_dag}
		Let $G$ be an ADMG with variables $V$ and $S\subset V$.
		The \emph{latent ADMG} $L(G, S)$ is the ADMG that contains all nodes in $S$, all edges between nodes in $S$ and additionally  
		\begin{enumerate}
			\item a directed edge between $X, Y\in S$ if there is a directed path from $X$ to $Y$ where all intermediate nodes are  in $V\setminus S$
			\item  a bidirected edge between $X, Y$ if there is a (undirected) path such that every non-endpoint is a non-collider in $V\setminus S$ and there are arrowheads towards $X$ and $Y$ on the incident edges on the path.
		\end{enumerate}
	\end{definition}


	\begin{definition}[graphical compatibility]
		
		\label{def:graphical_comp}
		Let  $G_{S_1},\dots,G_{S_k}$ be ADMGs  over the $k\in \N$ sets of nodes $S_1, \dots, S_k$ respectively. Then we define \emph{graphical compatibility} via $c(G_{S_1},\dots,G_{S_k}) = 1$ iff there exists a set $V\supseteq \cup_{i=1}^k S_i$, and 
		a DAG $G_V$ such that
		$L(G_V, S_i) = G_{S_i}$ for all $i\in [k]$. 
  Again, $c(\dots, \bot, \dots) = 0$ regardless of the other arguments.
\end{definition} 
Note, that neither interventional compatibility implies graphical compatibility nor vice versa. E.g. the empty graph is graphically compatible with its subgraphs, as graphical compatibility only depends on the distribution via the algorithm. But they are only interventionally compatible if the nodes have no causal effect onto each other in one Markovian joint graph. On the contrary, in \cref{sec:graphical_vs_interventional} we present an example where a non-generic distribution leads to interventionally compatible results that are not graphically compatible.  

\subsection{Falsifiability via Compatibility}
\label{sec:falsification}

In this section we demonstrate how compatibility between the outputs of an algorithm on different variables can be used to detect that either the assumptions of the algorithm are violated or there are finite sample effects in a way that actually change the output of an algorithm.
We will start by defining the terms \emph{observational falsifiability} and \emph{self-compatibility}.

\begin{definition}[observational falsifiability]
	\label{def:falsifiable}
	
	The output of an algorithm $\cA$ is \emph{observationally falsifiable} with respect to a compatibility notion $c$ \dominik{would we still call it observationally falsifiable if $c$ is a purely graphical notion? I see both arguments against and in favor.} if there exists a set of variables $V$, a distribution $P_V$
	and $k\ge 2$ subsets $S_1,\dots,S_k\subseteq V$ such
	that for all $\epsilon > 0$ there is an $m\in\N$ such that for all $m'\ge m$ we have
	\begin{equation}
		\label{eq:self_compatibility}
		c(\cA(\bX^{m'}_{S_1}),\dots,\cA(\bX^{m'}_{S_k}), P_V) = 0
	\end{equation}
	with probability at least $1-\epsilon$,
	where $\bX_V^{m'}$ (and all submatrices) is drawn from\footnote{Note, that we now used finite sample versions of the graphs $\cA(\bX^{m'}_{S_i})$ but still the population version of the distribution $P_V$ as last argument in $c(\dots, P_V)$. With the former, we want to emphasize that the graphs can be subject to finite sample effects. With the latter we want express that the statistical difficulties with testing the equivalence of the interventional distributions are beyond the scope of this work.} 
$P_V$.
	We call the left hand side of \cref{eq:self_compatibility} the \emph{self-compatibility}\footnote{Note that compatibility refers to graphical models while self-compatibility is a property of algorithms.} of $\cA$ w.r.t. $c$, $S_1, \dots, S_k$, $P_V$ and $\bX^{m'}_V$.  
    
\end{definition}
In words, falsifiability means that there exists a joint distribution for which $\cA$'s outputs on the subsets is incompatible according to $c$.
Note, that $c$ does not necessarily have to depend on $P_V$ as the last parameter, like e.g. in \cref{def:graphical_comp}.
To illustrate this definition, recall that in \cref{subsec:motivating_example} we have constructed a distribution such that the PC algorithm produces interventionally incompatible results on $\{Z_1, X, Y\}$ and $\{X, Y, Z_2\}$ in the limit of infinite data. 
Therefore the output of PC is observationally falsifiable.

\begin{remark}
\label{rem:falsifiability_measure_zero}
	One might wonder whether the existence of a \emph{single} incompatible distribution in \cref{def:falsifiable} might be a too weak condition. 
    In \cref{ex:entropy_ordering} we will see, that we can nonetheless find algorithms that are not falsifiable in this sense at all.
    We expect algorithms that are \emph{in principle} falsifiable to also be practically falsifiable.
    Quantifying how many distributions admit falsification is a difficult problem and would drastically increase the scope of this paper.
 \dominik{let us try to rephrase this remark.}
\end{remark}

The following result, while not surprising, ensures that there are no incompatibilities in the limit of infinite data if all assumptions are met. Accordingly, if $G$ is the causal DAG for a distribution $P_V$ over variables in $V$, we require that for any $S\subset V$ for which $P_S$
\emph{meets the assumptions} of $\cA$, 
and all $\epsilon>0$, there exists an $m\in \N$ such that $\cA(X^{m'}_{S}) =L(G,S)$
with probability at least $1-\epsilon$ for all $m'\geq m$.

\begin{lemma}
	\label{lem:stat_consistency}
 	Let $S_1\dots, S_k$  be $k\in \N$ sets of variables and $P_V$ be a  probability distribution over $V\supseteq \bigcup_{i\in [k]} S_i$ such that all $P_{S_i}$ with $i\in [k]$ fulfil the assumptions of $\A$.
	Then for every $\epsilon> 0$ there is an $m\in \N$ such that $\cA(\bX^{m}_{S_1}),\dots, \cA(\bX^{m}_{S_k})$ are interventionally and graphically compatible (w.r.t. to $P$) w.p. at least $1-\epsilon$.
 
\end{lemma}
\looseness=-1All proofs are in \cref{sec:proofs}.
Most causal discovery algorithms come with theoretical guarantees that their output is correct under some assumptions. 
But still, the following theorem shows for two exemplary algorithms that they are  falsifiable (see \cref{sec:proofs} for other algorithms). 
\begin{theorem}
	\label{lem:algos_nonbi}
	The FCI algorithm and the Repetitive Causal Discovery\footnote{RCD \citep{maeda2020rcd}  relies on linear models with non-Gaussian noise like LiNGAM \citep{shimizu2006linear} but does not assume causal sufficiency.} (RCD) algorithm are  falsifiable w.r.t. interventional compatibility.
\end{theorem}
The proof basically consists of constructing two distributions that have the same marginal distribution over two variables but would lead the algorithms to different causal models, similarly as in \cref{fig:graph_v_structure_wrong}.

Now we have established that the algorithms in \cref{lem:algos_nonbi} indeed have falsifiable outputs. But when all of their assumptions are met we can only \enquote{accidentally} falsify their output because of finite-sample effects, as  \cref{lem:stat_consistency} shows.
To illustrate that falsifiability is a non-trivial property, consider the following example.
\begin{example}[Entropy Ordering]
\label{ex:entropy_ordering}
    Define an algorithm $\cA $ that orders nodes according to a simple criterion (e.g. starting from variables with lowest entropy\footnote{Indeed, suggestions like this have been made, motivated by misconceptions on thermodynamics, as criticised by \cite{misconceptions}.}) and outputs an ADMG containing the complete DAG with respect to that order for directed edges and additionally bidirected edges between all nodes.
The outputs of this algorithm on any subset will be graphically compatible, as the marginal models are the latent projection of the models on supersets and also interventionally compatible, as no interventional distribution is identifiable in any of the graphs. 
\end{example}
This raises the question if there are properties that already imply the falsifiability of an algorithm.
Indeed, in \cref{sec:why_bivariate?} we will show that all causal discovery algorithms that are not \enquote{too simple} or \enquote{indecisive} can produce incompatible outputs on different sets of variables and hence, are falsifiable.

\dominik{did you check the proofs again? Not sure they are waterproof and they rely on a different type of outputs agnostic}

\begin{remark}
	\label{rem:gettier}
	Note that our approach does not falsify \emph{particular} causal graphs.
	Especially, an algorithm $\cA$ may well output the ground truth graph on $\bX$ but we can still find incompatible models on some subset.
	Due to \cref{lem:stat_consistency} we know that incompatibility of the outputs indicates that some assumptions are violated or there are errors due to finite sample effects.
	In this sense we would argue, that we cannot \emph{trust} the causal discovery algorithm in this case even if $\cA(\bX)$ happens to be the ground truth graph.\footnote{This situation bears similarity to the Gettier problem in epistemology. \citet{gettier1963knowledge} argues that a person does not \emph{know} a fact $A$ even if $A$ is true but the person based her belief in $A$ on false assumptions.}
\end{remark}

\subsection{Is Self-Compatibility a Strong Condition?}
\label{subsec:compatibility_strong}
We will now show that even though typically more than one joint distribution can have the same marginal distributions over some sets $S$ and $T$, there are cases where the assumptions of an algorithm $\cA$ render the joint distribution unique, as otherwise the outputs of $\cA$ on $S, T$ and $S\cup T$ would be incompatible. 
If the outputs of an algorithm on the marginals predict a unique joint distribution, any other potential joint distribution falsifies the outputs. On the other hand, if the joint data points do not contradict this unique joint, we count this, in the spirit of \citet{Popper1959}, as strong evidence in favor.

\begin{definition}[merging-enabling algorithms]
	\label{def:merging}
	An algorithm $\cA$ is said to \emph{enable merging} distributions 
	with respect to a notion of compatibility  $c$ if for some set of variables $V$ and $k\ge 2$
	there exist sets $S_1,\dots,S_k\subset V$ and 
	distributions 
	$P_{S_1},\dots,P_{S_k}$
	such that
	\begin{enumerate}
	\item there is exactly one 
	joint distribution $P_V$ such that for any $\epsilon>0$ there is an $m\in\N$ such that for all $m'\ge m$ we get 
	$c(\cA(\bX^{m'}_V), \cA(\bX^{m'}_{S_1}),\dots,\cA(\bX^{m'}_{S_k}), P_V) = 1$, 
	with probability at least $1-\epsilon$ and
	\item there exists
	a second distribution $\tilde{P}_V$ whose marginals coincide with all $P_{S_i}$ for $i\in[k]$,
	\end{enumerate} 
	where $\bX^{m'}_{S_j}$ is drawn from $P_{S_j}$ for $j\in[k]$.
\end{definition} 
Condition 1 states that there is only one joint distribution $P_V$ for which $\cA$'s output on $V$ does not contradict $\cA$'s outputs on the subsets. Condition 2 states that there would be more than one possible extension of the marginal probabilities without the graphical information provided by $\cA$.
In other words, the condition of compatibility of causal models implies constraints for the joint distributions that result in a  unique solution of the {\it causal} marginal problem, although the solution to the {\it probabilistic} marginal problem \citep{Vorobev1962} is not unique in this case.  

\begin{theorem}[FCI enables merging] 
	\label{thm:fci_merging}
	The FCI algorithm enables merging w.r.t. graphical compatibility on PAGs.
\end{theorem} 

In \cref{sec:is_compatibitlity_strong?} we provide a similar statement for an idealized version of RCD\footnote{The idealized version can output  $\bot$ if it detects that the given distribution has not been generated by a linear additive noise model. As we defined this token to be incompatible with all ADMGs, condition 1 of \cref{def:merging} allows us to rule out such distributions.}.
In the proof we construct an example where the algorithms output the existence of an \emph{unconfounded} edge. 
This edge rules out the existence of a confounding path and therefore implies a conditional independence that is not observable in the marginals.
In the example of the proof, this independence together with the given marginals suffices to identify the joint distribution.

Although these statements only ensure the \emph{existence} of cases where merging is possible, we think they illustrate the strength of the self-compatibility condition. In addition,  we demonstrate empirically that the condition is often violated in \cref{sec:experiments}. 


\paragraph{Relationship to interventions.} Note, that there is a close relationship between enabling merging and being able to predict the impact of interventions: let the node $X_i$ describes a coin flip, 
triggering the intervention $do(X_k=x_k)$ on another node $X_k$ in a set $S$ when $X_i=1$. When $\cA $ allows the unique reconstruction of $P_{\{X_i,X_j\}}$ after being applied to $S\cup \{X_i\}$ and $S\cup \{X_j\}$, it implicitly provides an interventional probability via $p(x_j|do (x_k))=p(x_j|x_i=1)$ {\it for an observer who knows that $X_i$ controls the intervention on $X_k$}.

\ifSubfilesClassLoaded{
	\bibliographystyle{abbrvnat}
	\bibliography{../aistats}
}{}

\end{document}

%% file: sections/score.tex
	\section{INCOMPATIBILITY SCORE}
	\label{sec:score}
	
	In this section we propose a practical score, that quantifies \enquote{how incompatible} the outputs of an algorithm applied to different subsets of variables are. 
	This score can be seen as a continuous relaxation of the binary notion of compatibility in \cref{def:abstract:compatibility}.
	We propose to use this relaxation, as we showed in \cref{thm:fci_merging} that self-compatibility can be a strong criterion and in practice it is often violated.
	This score is defined such that a perfect score indicates self-compatibility and in this sense the score can be used to \emph{falsify} the outputs of a causal discover algorithm as we described before.
	But moreover, our experimental results in \cref{sec:experiments} suggest that the continuous score could be used to \emph{evaluate} causal discovery algorithms in the sense that it could be used as a \enquote{proxy} for the structural Hamming distance, which cannot be evaluated without ground truth knowledge.
	The first score in this section is based on the interventional compatibility notion (\cref{def:interventional_comp}). We also present an incompatibility score based on graphical compatibility (\cref{def:graphical_comp}) and in \cref{sec:practical_compatibility_scores} we discuss further details of the scores.

\dominik{note, however, that the score does not capture anything about whether the outputs are informative or remain agnostic in large parts. A good score will only increase our trust in the algorithm if the outputs are sufficiently committing.}
 
	\Citet{su2022robustness} proposed a parametric test for whether the interventional distributions of different adjustment sets agree.
	We use this to test whether the parent-adjustment sets derived from  the marginal models $\cA(\bX_{S_1}), \dots, \cA(\bX_{S_k})$ yield the same causal effect as the joint one in $\cA(\bX)$.
	\begin{definition}[interventional score]
		\label{def:interventional_score}
		For $k\in \N$, let $S_1, \dots, S_k\subseteq V$ .
		We define the \emph{interventional incompatibility score} $\kappa^I$ of $\cA$ via
		\begin{equation}
			\label{eq:sb_I}
			\kappa^I(\cA, \bX) := C^{-1}\sum_{\mathclap{\substack{X, Y\in V\\ X\neq Y}}} T\left(X, Y, \cG(\cA, \bX)\right)
		\end{equation}
		where $\cG(\cA, \bX) := \{ \cA(\bX), \cA(\bX_{S_1}), \dots, \cA(\bX_{S_k})\}$ and we define $T(X, Y, \cG(\cA, \bX))=1$ if 
		\begin{enumerate}
			\item \looseness=-1there are (at least two) different valid parent-adjustment sets for $X$ and $Y$ in $\cG(\cA, \bX)$ and the test from \cite{su2022robustness} rejects the hypothesis that they all entail the same causal effect, or  
			\item there is an $i\in [k]$ such that the parent-adjustment for $X$ and $Y$ is valid in $L(\cA(\bX), S_i)$ but not in $\cA(\bX_{S_i})$ or vice versa,
		\end{enumerate}
		 else $T(X, Y, \cG(\cA, \bX)) = 0$ and $C$ is the number of pairs $X, Y$ such that there is at least one graph in $\cG(\cA)$ with a valid parent-adjustment set for $X$ and $Y$ (except for the cases where this graph is $\cA(\bX)$ and the effect is not identifiable in any $L(\cA(\bX), S_i)$ for $ i\in [k]$).
	\end{definition}
	This definition counts the cases where the algorithm makes incompatible statements about any pairwise causal effect on different subsets, normalized by the number of cases where the algorithm does commit to any falsifiable statement on any subset at all.

 The score defined in \cref{def:interventional_score} is only applicable to linear models, as we build on the test of \citet{su2022robustness}.
	As non-linear models are ubiquitous in practice, we also want to propose a score that can be applied in these settings.
	Graphical compatibility notions do not require any statistical test, thus we chose to build on them for this purpose.
	\Cref{def:graphical_comp} always refers to the existence of a joint model. Therefore we found it natural to check the 
	compatibility of each marginal causal graph with the causal graph $\cA(\bX)$ that an algorithm outputs on all available variables (even though, if we do not assume that the observed variables are causally sufficient, $\cA$ can at most output a latent projection of the joint graph in \cref{assumption:joint_model}).
	Further, \cref{def:graphical_comp} requires that the latent projection and the marginal graph are identical.
	We relax this notion by taking the SHD between the joint model $\cA(\bX)$ and a marginal model.
	The SHD is zero iff the graphs are identical.
	
	\begin{definition}[graphical incompatibility score]
		\label{def:graphical_score}
		Let $k\in\N$ and $S_1, \dots, S_k\subseteq V$.
		We define the \emph{graphical incompatibility score} $\kappa^G$ of $\cA$ via
		\begin{equation}
			\label{eq:sb}
			\kappa^G(\cA, \bX) := \frac{1}{k}\sum_{i\in [k]} \operatorname{SHD}\left(L(\cA(\bX), S_i), \cA(\bX_{S_i})\right).
		\end{equation}
		
	\end{definition}
	
	\paragraph{Relationship to stability.}
	While, as previously discussed, it is not possible to guarantee that a causal discovery algorithm that achieves a low incompatibility score will accurately predict system behavior under interventions, we argue that the resulting models are at least \textit{useful} due to their ability to predict statistical properties of {\it unobserved joint distributions}. This perspective is influenced by \citet{janzing2023reinterpreting}, who reconceptualizes causal discovery as a \textit{statistical learning problem}. The key principle underlying this reconceptualization posits that causal models offer predictive value beyond predicting system behavior under interventions; they can also predict statistical properties of unobserved joint distributions. See \cref{sec:stability_results} for the formal setup and description of this idea and the associated learning problem.
	
	Observe that, for a causal discovery method to achieve a low incompatibility score, it's output must remain largely unchanged under small modifications to the variable sets it is applied to. Following this idea, in \cref{sec:stability_results} we define a notion of stability of a causal discovery algorithm. Under this definition, we provide high-probability generalization bounds for causal models generated by stable causal discovery methods. The result demonstrates that stable algorithms, provably, generate useful causal models due to their ability to \textit{generalize statistical predictions across variable sets}. Informally, it provides evidence that a low incompatibility constitutes a useful inductive bias for causal discovery. This is notably distinct from the standard setting in statistical learning, where algorithms that exhibit stability under {small modifications to the data} are known to \textit{generalize across data points}\leena{Add references}.

	\ifSubfilesClassLoaded{
		\bibliographystyle{abbrvnat}
		\bibliography{../aistats}
	}{}
\end{document}

%% file: sections/experiments.tex
	\section{EXPERIMENTS}
	\label{sec:experiments}
	 We now explore the efficacy of the methods described in \cref{sec:score} on real and simulated data.
	The details of the experiments can be found in \cref{sec:exp_details} and the source code under \url{https://github.com/amazon-science/causal-self-compatibility}.
	In the main paper we present results for the RCD algorithm, as it is one of the few available algorithms that does not assume causal sufficiency and its outputs are close\footnote{Note however, that the output of RCD does not strictly describe ADMGs, as the the algorithm does not differentiate between purely confounded relationships and confounding with an additional direct edge.} to the presented formalism based on ADMGs.
 Additional experiments with other algorithms are shown in \cref{sec:additional_experiments}. 

\begin{figure}[htp]
		\centering 
		\includegraphics[width=\columnwidth]{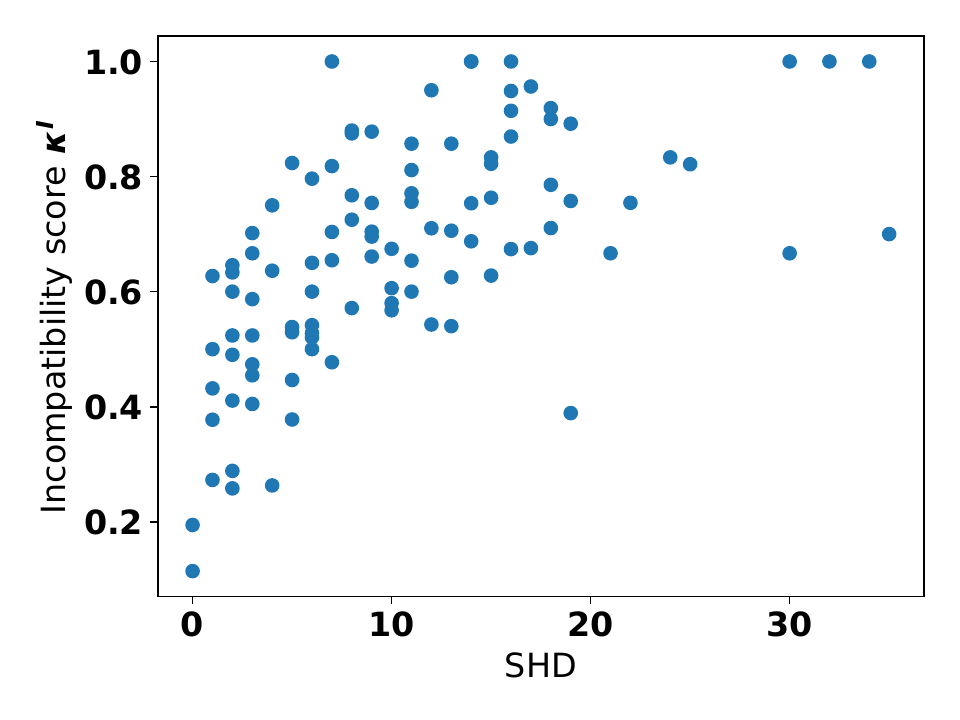}
		\caption{RCD on 100 datasets that fulfill its assumptions. 
			The plot shows structural Hamming distance of estimated graphs $\hat G$ to the respective true graph $G$ versus the interventional incompatibility score $\kappa^I$. 
   As both are influenced by the degree of the true graph, we also calculated the partial correlation given the average node degree of the true graph, which is $0.52$ with $p$-value $3\cdot 10^{-8}$. }
		\label{fig:rcd}
	\end{figure}

	\paragraph{Model evaluation.}
	\label{subsec:falsification}
	 As a first experiment we focus on a setting where we would expect a causal discovery algorithm to work reasonably well.
	We therefore generate 100 datasets with a linear model, uniform noise and potentially hidden variables.
	We use the incompatibility score $\kappa^I$ from \cref{def:interventional_score}.
	The first insight from the plot in \cref{fig:rcd} is that interventional compatibility indeed is a strong condition, in the sense that even in this scenario we find many interventionally incompatible marginal graphs.
	\dominik{let's discuss these statements.}
	Further, the plot in \cref{fig:rcd} shows a significant correlation between $\kappa^I$ and the structural Hamming distance of the joint graph to the true graph. 
	We suspected the density of the ground truth graph to
	influence both, the incompatibility score as well as the SHD.
	We also present the partial correlation between SHD and $\kappa^I$, adjusted for the average node degree of the ground truth graph, which stands at $0.52$ with a $p$-value of $3 \times 10^{-8}$.
	This suggests that $\kappa^I$ might be a useful proxy for SHD, which we cannot calculate in the absence of ground truth. 
	
	\paragraph{Model selection.}
	\label{subsec:model_selection}
\looseness=1	As we have seen that the incompatibility score is correlated with the SHD of the joint model to the ground truth, we now want to investigate whether the score could potentially be used to guide model selection and parameter tuning.
	We used the interventional score to select hyperparameters for the RCD algorithm.
	RCD has three threshold values for different independence tests.
	For simplicity we only checked the two configurations, where all of them are set to either $0.1$ or $0.001$ respectively.\footnote{Indeed this is overly simplistic, as in actual applications one would pick the parameters from a grid.}

\begin{figure}[ht]
		\centering
		\includegraphics[width=\columnwidth]{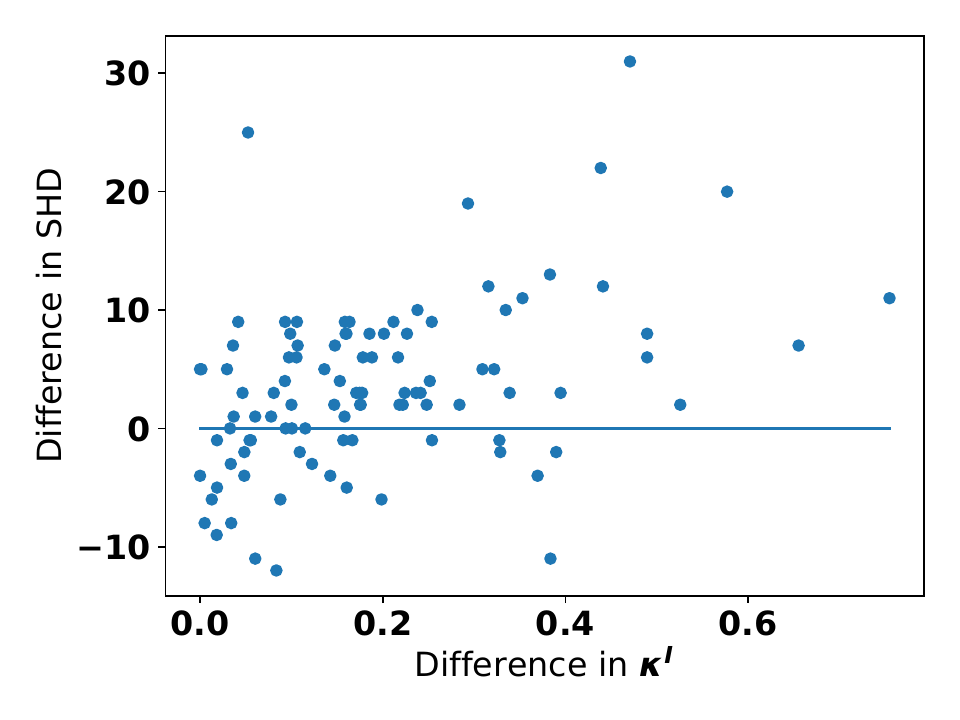}
		\caption{We chose between the hyperparameters $\alpha=0.1$ and $\alpha=0.001$ of RCD according to the incompatibility $\kappa^I$ for 100 datasets. 
    For 72\% of datasets we picked the better model or an equally good model.
    In most cases where we picked the worse model in terms of SHD the difference in $\kappa^I$ is small.}
		\label{fig:model_selection}
	\end{figure}

 \looseness=1 In \cref{fig:model_selection} we plot the difference in SHD between the estimated graphs and the ground truth graph, respectively, on the $y$-axis, where we always subtract the SHD of the algorithm with better $\kappa^I$ from the SHD of the algorithm with worse $\kappa^I$.
 Analogously for $\kappa^I$ on the $x$-axis.
 If the incompatibility score $\kappa^I$ was a \enquote{perfect} selection criterion we would hope to see all points on or above the horizontal line.
 In fact, 68\% are strictly above the line and 28\% are below the line.
	Moreover, we can see, that in most cases where the incompatibility score picked the hyperparameters that produce a worse SHD, the scores of the hyperparameters were close\footnote{This, of course, raises the question which differences should be considered \emph{significant}. The answer may depend on the particular downstream task---just as e.g. for SHD itself. Further, bootstrapping or permutation methods like the one proposed by \cite{eulig2023falsifying} might be helpful to derive a meaningful baseline. We defer this to future work.} to each other.

	\paragraph{Real data.} Finally, we also used the score on the biological dataset presented by \cite{sachs2005causal}. 
	We noted that all causal discovery algorithms that we tried performed quite poorly on this datasets compared to the algorithm's performance on simulated data.
	Our incompatibility score reflects this in the sense that in two out of four cases we get medium to bad incompatibility scores compared to the simulated experiments.
	The cases where the incompatibility score was good were the ones with the best results in terms of SHD and $F_1$ score.
	More details can be found in \cref{subsec:additional_real_data}.

	\ifSubfilesClassLoaded{
		\bibliographystyle{abbrvnat}
		\bibliography{../aistats}
	}{}
\end{document}

%% file: sections/related.tex
	\section{RELATED WORK}
	\label{sec:related_work}
To the best of our knowledge we are the first to leverage compatibility constraints of marginal causal models to falsify the output of causal discovery algorithms.
	
	\paragraph{Robustness of causal effects.}
	\looseness=-1While we proposed a method to falsify the underlying assumptions of causal discovery algorithms, there exists several methods for scenarios where causal directions are given and the goal is to estimate the strength of the treatment effect.
	E.g. \cite{walter2009variable,Lu2014,Oster2017} propose to test whether a regression model is causal by 
	dropping parts of the potential covariates and testing robustness.
	Similarly, \citet{su2022robustness} present the aforementioned parametric test for the case where a hypothetical graph is given.
	
	\paragraph{Evaluating causal models.}
	\looseness=-1The gold standard of evaluating the quality of a causal model would be to conduct a randomized control trial.
	\philipp{Remove this sentence?Minimizing the number of required interventions under different assumptions is an active field of research \cite{hauser2014two,kocaoglu2017experimental,acharya2018learning,jaber2020causal}}
	In contrast, our method 
	allows for falsification in settings where experiments are infeasible.
	There are other methods to judge the quality of causal models that do not rely on ground truth data as well, but they are limited to special cases, such as falsification via Verma-constraints \citep{verma2022equivalence} for instrument variables, tests that require parametric assumptions \citep{bollen1993confirmatory,daley2022experimentally} or the derivation of Bayesian uncertainty estimates \citep{claassen2012bayesian}. 
	A common approach is to count the number of $d$-separation statements that are not reflected in the data \citep{textor2017robust,reynolds2022validating}, although this either requires ground truth again or is also subject to assumptions such as faithfulness.
	\Citet{eulig2023falsifying} propose to reject causal graphs that do not reflect the conditional independences in the data significantly better than a random baseline.
	
	\paragraph{Causal marginal problem.}
	\looseness=-1\citet{tsamardinos2012towards} use causal models on 
	sets of variables $S,T$
	to predict conditional (in)dependences
	in $P_{S\cup T}$.
	For the toy scenario
	of a collider structure  with three binaries, \cite{gresele2022causal} 
	studied compatibility of
	structural equations for 
	the bivariate marginals, which amounts to falsifiability of causal statements of rung 3 in Pearl's ladder of causation
	\citep{Pearl2018}.
	In a scenario where causal directions are \dominik{In Gresele et al causal directions are also known} known, \cite{guo2023out} define \emph{out-of-variable generalization} as the capability of a machine learning algorithm to perform well across environments with different causal features, and 
	use marginal observations 
	to predict joint distributions. 
	\cite{janzing2023reinterpreting} 
	infers a DAG $G$
	on $S:=\bigcup_j S_j$ 
	in order to predict properties of $P_S$ 
	from the set of all $P_{S_j}$, which admits falsifying the DAG without interventions via  
	\enquote{test marginal distributions} $P_{S_{j+1}}$. 
	Despite the connections
 , the key difference of these approaches to our method is, that we do not use marginal distributions to reconstruct an unobservable joint distribution. 
	Instead, we propose to learn a causal model on the joint distribution but to falsify the output of the algorithm by learning marginal models.
	
	\ifSubfilesClassLoaded{
		\bibliographystyle{abbrvnat}
		\bibliography{../aistats}
	}{}
\end{document}

%% file: sections/supplement.tex
%

%

\onecolumn
\aistatstitle{Self-Compatibility: \\Evaluating Causal Discovery without Ground Truth \\
Appendix}

\section{FORMAL DEFINITIONS}
With our formalism we mainly follow \citet{peters2017elements} and \citet{pearl2009causality}.
\label{sec:formal}
\paragraph{Structural causal models}
Causality can be formalized mathematically by saying that all relationships between variables are governed by some deterministic functions except for some genuine, independent sources of randomness. We define structural models like \citet{pearl2009causality}.
\begin{definition}[structural causal model]
	Let $n\in \N$ and $V$ be a set of random variables $X_1, \dots, X_n$, $U$ be the a set of variables $N_1, \dots, N_n$ and $P_U$ be a probability distribution over $N_1, \dots, N_n$ such that all $N_1,\dots, N_n$ are jointly independent.
	Let there be a set of (measurable) functions $F$ such that for all $i\in[n]$ we have
	\begin{displaymath}
		X_i := f_i(PA_i, N_i),
	\end{displaymath}
	where $PA_i$ is some subset of $V\setminus\{X_i\}$ such that there are no cyclic dependencies between variables and $f_i$ depends on all variables in $PA_i$.
	Then we call $(V, U, F, P_U)$ a \emph{structural causal model} (SCM).
\end{definition}
Due to the acyclicity, the distribution $P_U$ also entails a unique joint distribution over $V$, as each value $x_i$ can be solved recursively until it only depends on noise terms.
Accordingly, we define interventions as \citet{pearl2009causality}.
\begin{definition}[intervention]
	Let $\cS = (V, U, F, P_U)$ be an SCM. The \emph{intervention} $do(X_i=x_i^*)$ for some $i\in[n]$ is defined by inserting the fixed value $x_i^*$ in all equations in $F$ that depend on $X_i$, regardless of other variables in the model.
	Denote with $\cS_{do(X_i=x_i)}$ the modified model with these equations.
	The \emph{interventional distribution} is defined as the distribution that canonically arises from $\cS_{do(X_i=x_i)}$ and we denote it with $P(X_S=x_S\mid do(X_i=x_i))$ for any $S\subseteq V$.
\end{definition}
\paragraph{Graphical models} 
Some of the causal aspects of a SCM can be represented  graphically. 
We will first define the graphical structures and then discuss their connection to the causal semantics defined via SCMs.
There are several popular graphical models for causality. We mostly follow \citet{zhang2008causal} and \citet{perkovic2015complete} with our formalism.
First recall the definitions of DAGs, mixed graphs and ADMGs from \cref{def:admg_main}.
We say there is an \emph{arrowhead towards} $X_j$ if there is an edge $X_j\to X_i\in E$ or $X_i\leftarrow X_j \in B$ and a \emph{tail} towards $X_j$ if there is an edge $X_i\leftarrow X_j\in E$.
We further say $X_i$ and $X_j$ are \emph{adjacent} if $X_i\to X_j \in E, X_j\to X_i\in E$ or $X_i\leftrightarrow X_j\in B$.

\begin{definition}[maximal ancestral graphs]
	An \emph{undirected path} is a sequence of nodes $X_0, \dots, X_k$ with $k\in\N$ such that either $X_i\to X_{i+1}\in E$, $X_{i+1}\to X_i\in E$ or $X_i\leftrightarrow X_{i+1} \in B$ for $i\in[k-1]$. 
	A node $X_i$ is called \emph{collider} on an undirected path $p$ if there are two edges with a head towards $X_i$ on $p$.
	We say there is an \emph{almost directed cycle} if there is a directed cycle from $X_i$ to $X_j$ and there is a bidirected edge $X_i\leftrightarrow X_j$.
	Let $L\subset V$ and call the first and the last nodes on a path the \emph{endpoints} of a path. \dominik{is it clear what they are in an almost directed cycle?}\philipp{don't get this remark}
	An \emph{inducing path} relative to $L$ is an undirected path $p$, such that every node on $p$ that is in $V\setminus L$ except for the endpoints is a collider on $p$ and an ancestor of one of the endpoints.
	A mixed graph $G$ is called a \emph{maximal ancestral graph} (MAG) if it contains no almost directed cycles and there is no inducing path between non-adjacent nodes.
\end{definition}

\begin{definition}[$m$-separation]
	Let $G=(V, E, B)$ be a mixed graph. 
	An undirected path $p$ between $X_i$ and $X_j$ in $G$ is called \emph{$m$-connecting} \dominik{wouldn't we rather call $X_i,X_j$ connected by $p$ instead of calling the path connected?} given a set $Z\subset V\setminus\{X_i, X_j\}$ if every non-collider on $p$ is not in $Z$ and every collider on $p$ is an ancestor of a node (or is itself) in $Z$. 
	If no undirected path between $X_i$ and $X_j$ is $m$-connecting given $Z$ we say $X_i$ and $X_j$ are \emph{$m$-separated} by $Z$ and write $X_i \perp_G X_j\mid Z$.
	In a DAG $m$-separation reduces to well-known $d$-separation.
\end{definition}

\paragraph{Graphical models and causality}
We will now connect the graphical models with causal semantics.
\begin{definition}[causal DAG]
	Let $P$ be a probability distribution over variables in $V$ that has been generated by an SCM $S$ as described above.
	We call a DAG $G$ a \emph{causal graph} of $P$ if $G$ contains a node for each variable in $V$ and an edge from $X_i$ to $ X_j\in V$ iff $X_i\in PA_j$ in $S$.
\end{definition}

\begin{definition}[Global Markov condition]
	We say a probability distribution $P$ over $V$ fulfils the \emph{global Markov condition} w.r.t. the DAG (MAG)  $G$ if for every two nodes $X_i, X_j\in V$ and set $Z\subseteq V\setminus\{X_i, X_j\}$ we have that $X_i\perp_G X_j\mid Z$ implies $X_i\ind X_j\mid Z$.
	We also say $P$ is \emph{Markovian} w.r.t. $G$.
	If $X_i\ind X_j\mid Z$ also implies $X_i\perp_G X_j\mid Z$ we say $P$ is \emph{faithful} to $G$.
\end{definition}
A distribution that has been generated by an SCM is always Markovian w.r.t. its causal DAG.

\begin{definition}[Markov equivalence]
	Two DAGs (MAGs) $G_1, G_2$ are \emph{Markov equivalent} if for all nodes $X_i, X_j \in V$ and sets of nodes $Z\subseteq V\setminus\{X_i, X_j\}$ we have $X_i\perp_{G_1} X_j\mid Z$ iff $X_i\perp_{G_2} X_j\mid Z$.
	We call 
	\begin{displaymath}
		[G]:=\{G' : G \text{ and } G' \text{ are Markov equivalent}\}
	\end{displaymath}
	the \emph{Markov equivalence class} of $G$.
\end{definition}

\begin{definition}[CPDAG]
	Let $G$ be a DAG.
	The \emph{completed partially directed graph} $C$ of $[G]$ is a mixed graph that contains a directed edge $X_i\to X_j$ iff this edge exists in all DAGs in $[G]$ and a bidirected edge $X_i\leftrightarrow X_j$ iff there is a DAG in $[G]$ with the edge $X_i\to X_j$  and a DAG in $[G]$ with the edge $X_i\leftarrow X_j$.
	We call $C$ a \emph{causal CPDAG} if $G$ is a causal DAG.
\end{definition}
The partial ancestral graph that we now introduce can also represent the presence of selection bias. 
In this paper we omit this part of the formalism (similarly to \citet{zhang2008causal}).
\begin{definition}[PAG]
	Let $G$ be a MAG over variables in a set $V$.
	The \emph{partial ancestral graph} of $[G]$ is a graph $H = (V, E)$ with a symmetric set of edges $E\subseteq V\times V$ and a map $\operatorname{end}: E\to \{>, -, \circ\}$ such that for $X_i, X_j\in V$
	\begin{itemize}
		\item $(X_i, X_j)\in E$ and $(X_j, X_i)\in E$ if $X_i, X_j$ are adjacent in any graph in $[G]$
		\item $\operatorname{end}(X_i, X_j) =$\enquote{$>$} iff there is an arrowhead from $X_i$ to $X_j$ in every graph in $[G]$
		\item $\operatorname{end}(X_i, X_j) =$\enquote{$-$} 
  iff there is \emph{no} arrowhead from $X_i$ to $X_j$ in every graph in $[G]$
		\item $\operatorname{end}(X_i, X_j)=$\enquote{$\circ$} else.
	\end{itemize}
	$X_i$ and $X_j$ are called adjacent if $(X_i, X_j)\in E$.
	We say there is a \emph{directed edge} from $X_i$ to $X_j$ if $\operatorname{end}(X_i, X_j) =$\enquote{$>$} and $\operatorname{end}(X_j, X_i) = $\enquote{$-$} and then define directed paths if there are paths of directed edges. 
	We say there is a \emph{possibly directed path} from $X_1$ to $X_k$ if there are nodes $X_1, \dots, X_{k}$ such that $X_i$ is adjacent to $X_{i+1}$ and there is no arrowhead towards $X_i$ for $i\in[k-1]$.
	A node $X_j$ is a \emph{possible descendant} of $X_i$ if there is a possibly directed path from $X_i$ to $X_j$.
	A node $X_i$ is a collider on a path if there are two arrowheads towards $X_i$ on that path.
	A collider path is a path, such that every non-endpoint is a collider.
	A direct edge is also a (trivial) collider path.
	A \emph{definite non-collider} on a path is a node $X_i$ that has at least one tail towards $X_i$ on the path and a \emph{definite status path} is path such that every node on the path is either a collider or a definite non-collider on this path.
\end{definition}

\begin{definition}[visible edge]
	\label{def:visible}
	Every edge in a DAG or CPDAG is \emph{visible}.
	A directed edge $X_i\to X_j$ in a MAG (PAG) $G$ is visible if there is a node $X_k$ not adjacent to $X_i$ and there is a collider path from $X_k$ to $X_i$ such that every non-endpoint of the path is a parent of $X_j$ and the last edge towards $X_i$ has an arrowhead at $X_i$.
\end{definition}
Intuitively, a visible edge indicates an \emph{unconfounded} edge, i.e. that there is no hidden confounder between the start and the endpoint of the edge.

\paragraph{Identification}
We now want to define what we mean with \emph{identifiability}.
The following definition is from \citet{tian2002general}.
\begin{definition}[identifiability]
	\label{def:identifiability}
	Let $\cS$ be a set of SCMs and $S, S'\in\cS$ be SCMs with the same causal graph. Any quantity of $Q(S)$ is \emph{identifiable} in $\cS$ if $Q(S) = Q(S')$ whenever $P_S$ coincides with $P_{S'}$, where $P_S$ denotes the probability distribution entailed by $S$.
\end{definition}
Often one assumes that the causal graph $G$ is known and then tries to derive the property $Q(S)$ from $G$ and the distribution $P_S$.
We want to make the distinction between the graphical properties of $G$ and the distribution $P_S$ a bit more explicit, as in our self-compatibility framework we deal with estimated graphs that might not correctly represent the underlying data generation.
In such a case, the identification formulae derived from such a graph may be incorrect and different identification formulae may even lead to contradicting results for the quantities of interest.
Note, that the following formalism assumes for an interventional probability $p(x_j|do(x_i))$ that $x_i, x_j$ are fixed.
\begin{definition}[identification formula]
	Let $S$ be an SCM with causal graph $G$ and identifiable quantity $Q(S)$.
	A \emph{identification formula} in $G$ is a map $g:\cP\to \R$, such that for a probability distribution $P_S$ of $S$ we have $g(P) = Q(S)$, where $\cP$ denotes the space of probability distributions.
\end{definition}
\dominik{interventional probabilities only fit into this definition if we look at $p(x_j|do(x_i))$ for fixed $x_i,X_j$.} 
Note that if we have two different graphs $G, G'$, we may have identification formulae $g,g'$ for $Q$ in $G$ and in $G'$ respectively but for a distribution $P$ we get $g(P)\neq g'(P)$, as $G$ and $G'$ may be different graphs.

\dominik{do we need the general definition of identifiability for some arbitrary quantity or shouldn't we just talk about identification of interventional probs?} 

\paragraph{Latent projection}
In \cref{def:latent_projection_dag} we have already defined how to project ADMGs (and therefore also DAGs, as they are ADMGs without hidden confounders) to an ADMG that represent only a subset of variables.
We will now define similar projections for MAGs, PAGs and CPDAGs,
and start be defining a projection of a DAG to a MAG (see also \citep{zhang2008causal}).
We mainly do this to connect MAGs and PAGs to the SCM formalism. 

\begin{definition}[latent MAG]
	\label{def:latent_MAG}
	Let $G$ be a DAG with variables in $V$ and $S\subseteq V$.
	The \emph{latent MAG} $L^{\text{MAG}}(G, S)$ is the MAG that contains
	all nodes in $S$ and and for $X_i, X_j\in S$
	\begin{enumerate}
		\item an edge between $X_i$ and $X_j$ iff there is an inducing path in $G$ between them
		\item an arrowhead at $X_i$ (or $X_j$) iff the last edge of the inducing path has an arrowhead at $X_i$ ($X_j$).
	\end{enumerate}
	When no confusion with \cref{def:latent_projection_dag} can arise, we also just write $L(G, S)$.
	We call a MAG $M$ a \emph{causal MAG} if it is the latent projection of a causal DAG $G$.
\end{definition}

\begin{definition}[latent PAG]
	\label{def:latent_PAG}
	Let $H$ be a PAG with variables in $V$, $S\subseteq V$ and $M$ be some MAG in the equivalence class described by $H$.
	The \emph{latent PAG} $L^{\text{PAG}}(H, S)$ is the PAG of $[L^{\text{MAG}}(M, S)]$.
	We call a PAG $H$ a \emph{causal PAG} if it is the latent projection of a causal MAG.
\end{definition}
The latent PAG is well-defined, as all MAGs in the same equivalence class also have the same independence statements in $S$.
Therefore, these independences are represented by the same PAG, i.e. defining the latent PAG via \emph{some} MAG does not introduce arbitrariness.

It is not possible to define a projection operator for CPDAGs without assumptions about the subset $S$, as this model class cannot represent the presence of latent confounders.
Nonetheless, we wanted to include them in our framework as popular causal discovery algorithms like PC and GES output CPDAGs.
We chose to only define the projection operator for sets $S$ that fulfil the causal sufficiency assumption, i.e. sets that do not contain two nodes $X_i, X_j\in S$ with common ancestor $L\in V\setminus S$ such that any intermediate nodes on paths from $L$ to $X_i$ or $X_j$ are also in $V\setminus S$. 
In other words, the latent ADMG does not contain birected edges: 
\begin{definition}[latent CPDAG]
	\label{def:latent_projection_cpdag}
	Let $G$ be a DAG with variables $V$ and $S\subset V$ be a subset such that the latent ADMG $L(G, S)$ contains no bidirected edges.
	Then the \emph{latent CPDAG} $L(G, S)$ is the CPDAG that represents the equivalence class $[L(G, S)]$.
\end{definition}

\paragraph{Adjustment criteria}
The following theorems from the literature provide graphical criteria to identify interventional probabilities.
The first one is Theorem 1 in \citep{tian2002general}.
\begin{theorem}[parent adjustment in ADMGs]
	\label{thm:parent_adj_admg}
	Let $G$ be a causal ADMG over discrete variables in $V$ and $X_i, X_j\in V$.
	If there is no bidirected edge connected to $X_j$, we have 
	\begin{displaymath}
		p(x_i| do(x_j)) = \sum_{pa_i} p(x_i| x_j, pa_j) p(pa_j).
	\end{displaymath}
\end{theorem}
The sum can easily be replaced by an integral for continuous variables with positive densities. \philipp{is that right?}
\citet{tian2002general} also show the more general result in their Theorem 3:
\begin{theorem}[generalised identifiability in ADMGs]
	\label{thm:identifiability_admgs}
	Let $G$ be a causal ADMG over discrete variables in $V$ and $X_i, X_j\in V$.
	The probability $p(x_i|do(x_j))$ is identifiable iff there is no bidirected path between $X_j$ and any of $X_j$'s children.
\end{theorem}
For the other graphical models we considered, \citet{perkovic2015complete} provided the following result in their Theorem 3.4, where we restrict ourselves to the case of adjustment between single variables. First we define the \emph{forbidden set}.
\begin{definition}[forbidden set]
    Let $G$ be a causal DAG, CPDAG, MAG or PAG for the probability distribution $P$ over $V$, $X_i, X_j\in V$.
    Denote with $\operatorname{Forb}(X, Y, G)$ the set of possible descendant in $G$ of any $W\in V\setminus\{X_j\}$ that lies on a possibly directed path from $X_j$ to $X_i$ and call this set the \emph{forbidden set}.
\end{definition}
\begin{theorem}[generalised adjustment criterion]
	\label{thm:generalised_identification}
	Let $G$ be a causal DAG, CPDAG, MAG or PAG for the probability distribution $P$ over $V$, $X_i, X_j\in V$ and $Z\subseteq V\setminus\{X_i, X_j\}$.
	Then we have
	\begin{displaymath}
		p(x_i|do(x_j)) = \begin{cases}
			p(x_i|x_j), \text{ if } Z = \emptyset\\
			\int p(x_i| x_j, x_Z) p(x_Z) dx_Z \text{, else}
		\end{cases}
	\end{displaymath}
	if and only if 
	\begin{enumerate}
		\item every possibly directed path in $G$ from $X_j$ to $X_i$ starts with a visible edge (see \cref{def:visible})
		\item $Z \cap \operatorname{Forb}(X, Y, G) = \emptyset$ 
		\item all definite status paths that are not directed are $m$-separated by $Z$.
	\end{enumerate}
\end{theorem}

\citet{perkovic2015complete} also give a definition for a set that is a valid adjustment set, iff there is a valid adjustment set.
\begin{definition}[canonical adjustment set]
    Let $G$ be a causal DAG, CPDAG, MAG or PAG for the probability distribution $P$ over $V$, $X_i, X_j\in V$.
    We call the set
    \begin{displaymath}
        \operatorname{Adjust}(X, Y, G) = \operatorname{PossAnc}(\{X, Y\}, G) \setminus (\operatorname{Forb}(X, Y, G) \cup \{X, Y\})
    \end{displaymath}
    the \emph{canonical adjustment set}, where $\operatorname{PossAnc}(\{X, Y\}, G)$ is the set of possible ancestors of $X$ and $Y$.
\end{definition}

\section{PROOFS FOR THE MAIN PAPER}
\label{sec:proofs}
\subsection{For \Cref{lem:stat_consistency}}
\begin{proof}
Let $S_1\dots, S_k$  be $k\in \N$ sets of variables and $P_V$ be a  probability distribution over $V\supseteq \bigcup_{i\in [k]} S_i$ such that all $P_{S_i}$ with $i\in [k]$ fulfil the assumptions of $\A$.
Further, let $\epsilon > 0$.
Set 
\begin{displaymath}
    \delta := 1 - \sqrt[k]{1-\epsilon}.
\end{displaymath}
As every marginal distribution fulfils the assumptions of $\cA$, we know that for every $i\in [k]$ there is a $m_i\in \N$ such that for all $m'>m_i$ we get $\cA(X^{m'}_{S_i}) =L(G,S_i)$ with probability at least $1-\delta$, where again $G$ is the true causal DAG.
Set $m^* = \max_{i\in [k]} m_i$.
Then we get
\begin{displaymath}
    P(\exists i\in [k] : \cA(X^{m*}_{S_i}) \neq L(G,S_i)) = 1 - \prod_{i\in [k]}P(\cA(X^{m*}_{S_i}) = L(G,S_i)) 
    \le 1 - (1 - \delta)^k
     = \epsilon.
\end{displaymath}

The graphical compatibility follows directly from \cref{def:graphical_comp}.
Similarly, the interventional compatibility follows from the fact that the algorithms find the latent projections of $G$.
This renders them causal
in the sense of \cref{thm:identifiability_admgs,thm:generalised_identification}
 and therefore the interventional probabilities coincide with the ones in $G$ if they are identifiable.
\end{proof}

\subsection{For \Cref{lem:algos_nonbi}}
\label{subsec:proof_falsifiable}
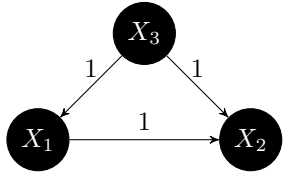
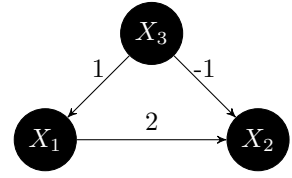
\begin{figure}[htp]
		\centering
		
		\begin{subfigure}[c]{0.45\textwidth}
			\centering
			\begin{tikzpicture}
				\node[obs,circle] (x1) {$X_1$};
				\node[obs,circle,above right of=x1] (x3){$X_3$}; 
				\node[obs,circle,below right of=x3] (x2) {$X_2$};
				
				\draw (x1) edge [->] node[above]{1} (x2);
				\draw (x1) edge[<-] node[above]{1} (x3);
				\draw (x2) edge[<-] node[above]{1} (x3);

			\end{tikzpicture}
			\caption{Graph $G_1$ with structural coefficient 1 between $X_1$ and $X_2$.}
			\label{fig:proof_lingam_nonbi_first}
		\end{subfigure}
		\hfill
		\begin{subfigure}[c]{0.45\textwidth}
			\centering
			\begin{tikzpicture}
				\node[obs,circle] (x1) {$X_1$};
				\node[obs,circle,above right of=x1] (x3){$X_3$}; 
				\node[obs,circle,below right of=x3] (x2) {$X_2$};
				
				\draw (x1) edge [->] node[above]{2} (x2);
				\draw (x1) edge[<-] node[above]{1} (x3);
				\draw (x2) edge[<-] node[above]{-1} (x3);

			\end{tikzpicture}
			\caption{Graph $G_2$ with structural coefficient 2 between $X_1$ and $X_2$.}
			\label{fig:proof_lingam_nonbi_second}
		\end{subfigure}
		
		\caption{Two causal models that fulfil the LiNGAM assumption, have the same marginal over $X_1$ and $X_2$ and different coefficient from $X_1$ to $X_2$.
		}
		\label{fig:proof_lingam_nonbi}
	\end{figure}

\begin{proof}
	We will prove the statement separately for the different algorithms. 
	For FCI 
    we explicitly construct a joint distribution such that the algorithms make contradicting interventional statements on different subsets, as our motivating example from \cref{subsec:motivating_example} almost suffices to show the statement.
	For RCD we show an example where two LiNGAM models with different linear coefficient between $X_i$ and $X_j$ generate the same marginal for $X_i$ and $X_j$.
    In section \cref{sec:why_bivariate?} we will show that this suffices to render RCD falsifiable.
	Precisely, in this proof we will show that RCD is non-bivariate (\cref{def:nonbi}) and therefore \cref{thm:icommitt} implies that it is falsifiable.
	
	For FCI we have to slightly modify the example from \cref{subsec:motivating_example} to render the edge $X\to Y$ visible (and therefore the interventional probability identifiable).
	To this end, assume we have the graph $G$ shown in \cref{fig:fci_falsifiable_true_graph}.
	Assume that as before, all independences are given by $d$-separation in $G$ except for the additional independence $Y\ind Z_2$.
	On the set $S'=\{X, Y, Z_1, Z_3, Z_4\}$, FCI will find all edges between nodes in $S'$ that are in $G$ as shown in \cref{fig:fci_falsifiable_s} (except for some circle marks).
	In this subgraph, the edge $X\to Y$ is visible as $Z_3$ (or $Z_4$) has an edge towards $X$ but is not adjacent to $X$.
	Further, there are no non-causal paths 
    (i.e. backdoor-paths) between $X$ and $Y$.
	Then the empty set is a valid adjustment set according to \cref{thm:generalised_identification}.
	On $T$, FCI will find the marginal model as in \cref{fig:fci_falsifiable_t} for the same reason as in \cref{subsec:motivating_example}.
	As there is an arrowhead towards $X$, the there is no effect from $X$ to $Y$.
	Now we get the same interventional probabilities $p^{S'}(y|do(x)) = p(y|x) \neq p(y) =  p^T(y|do(x))$ as in \cref{subsec:motivating_example}, where $S' = S\cup \{Z_3, Z_4\}$.
 Therefore we have constructed a joint distribution such that we can find incompatible results on some subsets in the limit of infinite data.

 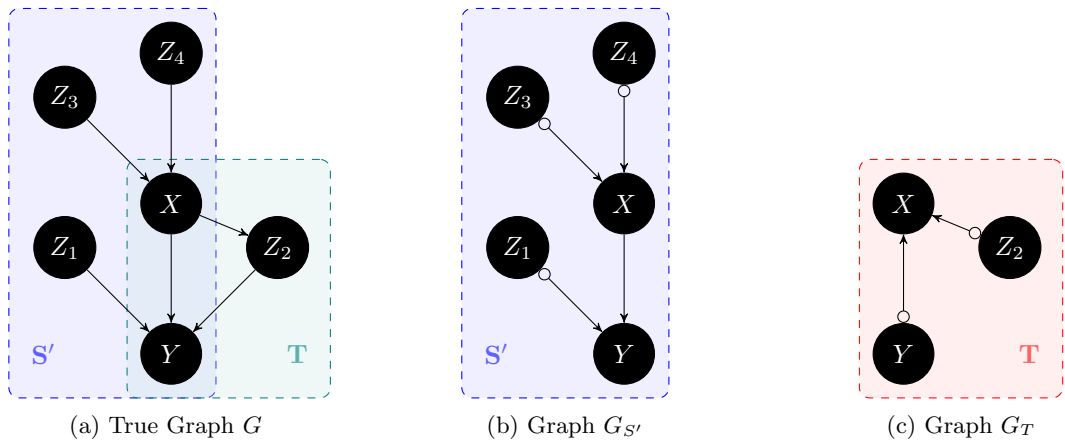
\begin{figure}[htp]
  \subcaptionbox{True Graph $G$
			\label{fig:fci_falsifiable_true_graph}}[0.3\textwidth]{
			\centering
			\begin{tikzpicture}
		\node[obs,circle] (x) {$X$};
		\node[obs,circle,below of=x] (y) {$Y$};
		\node[obs,circle,above left of=y] (z1) {$Z_1$};
		\node[obs,circle,above right of=y] (z2) {$Z_2$};
		\node[obs,circle,above left of=x] (z3) {$Z_3$};
		\node[obs,circle,above  of=x] (z4) {$Z_4$};
		
		\draw (x) edge [->] (y);
		\draw (y) edge[<-] (z1);
		\draw (y)edge[<-] (z2);
		\draw (x) edge[->] (z2);
		\draw (z3) edge[->] (x);
		\draw (z4) edge[->] (x);
		
		\begin{pgfonlayer}{background}
			\node[blue!80, left=of y, xshift=1.0cm] (s2) {$\mathbf{S'}$};
			\node[teal!80,  right=of y, xshift=-1.0cm] (t2) {$\mathbf{T}$};
			\node[fit=(z4) (y) (s2),draw,blue,dashed,inner sep=5pt,fill=blue!20,fill opacity=0.3,rounded corners] {};
			\node[fit=(x) (y) (t2),draw,teal,dashed,inner sep=5pt,fill=teal!20,fill opacity=0.3,rounded corners] {};
		\end{pgfonlayer}
		
	\end{tikzpicture}
   }
    \subcaptionbox{Graph $G_{S'}$\label{fig:fci_falsifiable_s}}[0.3\textwidth]{
			\centering
			\begin{tikzpicture}
		\node[obs,circle] (x) {$X$};
		\node[obs,circle,below of=x] (y) {$Y$};
		\node[obs,circle,above left of=y] (z1) {$Z_1$};
		\node[obs,circle,above left of=x] (z3) {$Z_3$};
		\node[obs,circle,above  of=x] (z4) {$Z_4$};
		
		\draw (x) edge [->] (y);
		\draw (y) edge[<-o] (z1);
		\draw (z3) edge[o->] (x);
		\draw (z4) edge[o->] (x);
		
		\begin{pgfonlayer}{background}
			\node[blue!80, left=of y, xshift=1.0cm] (s2) {$\mathbf{S'}$};
			\node[fit=(z4) (y) (s2),draw,blue,dashed,inner sep=5pt,fill=blue!20,fill opacity=0.3,rounded corners] {};
		\end{pgfonlayer}
		
	\end{tikzpicture}
   }
    \subcaptionbox{Graph $G_T$\label{fig:fci_falsifiable_t}}[0.3\textwidth]{
			\centering
			\begin{tikzpicture}
		\node[obs,circle] (x) {$X$};
		\node[obs,circle,below of=x] (y) {$Y$};
		\node[obs,circle,above right of=y] (z2) {$Z_2$};
		
		\draw (x) edge [<-o] (y);
		\draw (x) edge[<-o] (z2);
		
		\begin{pgfonlayer}{background}
			\node[red!80,  right=of y, xshift=-1.0cm] (t2) {$\mathbf{T}$};
			\node[fit=(x) (y) (t2),draw,red,dashed,inner sep=5pt,fill=red!20,fill opacity=0.3,rounded corners] {};
		\end{pgfonlayer}
		
	\end{tikzpicture}
   }	
	\caption{This modification of \cref{fig:motivation_SandT_false} renders the edge $X\to Y$ visible if FCI is applied to $S'$ and thus shows that FCI is falsifiable.}
	\label{fig:proof_nonbi_fci}
\end{figure}

	For RCD 
    we modify an example from \citet{hoyer2008estimation}.
	Precisely, their example V consists of a linear SCM with three nodes defined via 
	\begin{displaymath}
		X_3 := N_3, \quad
		X_1 := \beta X_3 + N_1, \quad
		X_2 := \alpha X_1 + \gamma X_3 + N_2,
	\end{displaymath}
	where $N_1, N_2, N_3$ are jointly independent non-Gaussian noise variables.
	\Cref{fig:proof_lingam_nonbi} shows two such models.
	\citet{hoyer2008estimation} show that there are two models that cannot be distinguished on the marginal over $\{X_1, X_2\}$, one with structural coefficient $\alpha$ and one with $(\alpha\beta+\gamma)/\beta$ between $X_1$ and $X_2$.
	For example, we can set $\alpha=\beta=\gamma=1$ and assume all noise variables to have the same distribution with zero mean and unit variance as visualized in \cref{fig:proof_lingam_nonbi_first}.
     This model generates the distribution $P$.
	Analogously, we define another distribution $\tilde P$ via a model with $\alpha=2, \beta=1, \gamma=-1$  (we require the noise terms to have the same distributions as in the previous model) shown in \cref{fig:proof_lingam_nonbi_second}.
    Then the joint behaviour between $X_1$ and $X_2$ can be described via the vector
    \begin{displaymath}
        \begin{pmatrix}
            X_1\\ X_2
        \end{pmatrix} = 
        \begin{pmatrix}
            1 & 0 & 1\\
            1 & 1 & 2
        \end{pmatrix} \cdot 
        \begin{pmatrix}
            N_1\\ N_2\\ N_3
        \end{pmatrix} = 
        \begin{pmatrix}
            N_1 + N_3\\
            N_1 + N_2 + 2 N_3
        \end{pmatrix}
    \end{displaymath}
    for $P$, while for $\tilde P$ we get
        \begin{displaymath}
        \begin{pmatrix}
            X_1\\ X_2
        \end{pmatrix} = 
        \begin{pmatrix}
            1 & 0 & 1\\
            2 & 1 & -1
        \end{pmatrix}\cdot 
        \begin{pmatrix}
            N_1\\ N_2\\ N_3
        \end{pmatrix} = 
        \begin{pmatrix}
            N_1 + N_3\\
            2 N_1 + N_2 + N_3
        \end{pmatrix}.
    \end{displaymath}
    Since all noise terms have the same distribution, the vector $(X_1, X_2)^T$ has the same distribution in both cases.
    
	Further, RCD will identify the respective model when all three nodes are observed.  	
	In other words, there are two distributions $P, \tilde P$ over $X_1, X_2, X_3$ (generated by the SCMs described above) that have identical marginals over $X_1, X_2$ but RCD results in different interventional probabilities $p(x_2|do(x_1)) \neq p'(x_2| do(x_1))$.
	Therefore  RCD is non-bivariate (\cref{def:nonbi}) and via \cref{thm:icommitt} it is falsifiable.
\end{proof}

\subsection{For \Cref{thm:fci_merging}}
\begin{proof}
	The idea is that
	we apply FCI to data generated from the DAG in \cref{fig:fci_merging_true_graph} on $S\cup\{i\}$ and $S\cup\{j\}$. 
	In the first case, FCI identifies the direction $i\to k$. 
    Likewise,  FCI infers that $k\to j$ is visible, i.e. that $k$ is an unconfounded cause of $j$ in the second case. This unconfoundedness excludes a direct link between $i$ and $j$, as otherwise $i$ would be a confounder for $k$ and $j$.
    Further, $S$ also contains no common child of $i$ and $j$.
    Consequently, $S$ $m$-separates $i$ and $j$.
	From these independences, we can reconstruct the entire joint distribution. 

 \begin{figure}[htp]
  \subcaptionbox{True Graph $G$
			\label{fig:fci_merging_true_graph}}[0.3\textwidth]{
			\centering
			\begin{tikzpicture}
			
			\node[ obs] (p1) {$p_1$};
			\node[ obs] (p2) [below of=p1] {$p_2$};
			\node[ obs] (i) [above right of=p1] {$i$};
			\node[ obs] (k) [below right of=i] {$k$};
			\node[ obs] (j) [below right of=k] {$j$};
			
			
			\draw (p1) edge[->] (i);
			\draw (p2) edge[->] (i);  
			\draw (i) edge[->] (k);
			\draw (k) edge[->] (j);
			
			\begin{pgfonlayer}{background}
				\node[blue!80, left=of i, xshift=2.0cm] (s2) {$\mathbf{S\cup\{i\}}$};
				\node[teal!80,  below=of j, yshift=1.8cm] (t2) {$\mathbf{S\cup\{j\}}$};
				\node[fit=(p1) (i) (p2) (k),draw,blue,dashed,inner sep=5pt,fill=blue!20,fill opacity=0.3,rounded corners] {};
				\node[fit=(p1) (j) (p2),draw,teal,dashed,inner sep=5pt,fill=teal!20,fill opacity=0.3,rounded corners] {};
			\end{pgfonlayer}

		\end{tikzpicture}
   }
    \subcaptionbox{Graph $G_{S\cup \{i\}}$\label{fig:fci_merging_s_i}}[0.3\textwidth]{
			\centering
			\begin{tikzpicture}
			
			\node[ obs] (p1) {$p_1$};
			\node[ obs] (p2) [below of=p1] {$p_2$};
			\node[ obs] (i) [above right of=p1] {$i$};
			\node[ obs] (k) [below right of=i] {$k$};
			
			
			\draw (p1) edge[o->] (i);
			\draw (p2) edge[o->] (i);  
			\draw (i) edge[->] (k);
			
			\begin{pgfonlayer}{background}
				\node[blue!80, left=of i, xshift=2.0cm] (s2) {$\mathbf{S\cup\{i\}}$};
				\node[fit=(p1) (i) (p2) (k),draw,blue,dashed,inner sep=5pt,fill=blue!20,fill opacity=0.3,rounded corners] {};
			\end{pgfonlayer}

		\end{tikzpicture}
   }
    \subcaptionbox{Graph $G_{S \cup \{j\}}$\label{fig:fci_merging_s_j}}[0.3\textwidth]{
			\centering
			\begin{tikzpicture}
			
			\node[ obs] (p1) {$p_1$};
			\node[ obs] (p2) [below of=p1] {$p_2$};
			\node[ obs] (k) [below right of=i] {$k$};
			\node[ obs] (j) [below right of=k] {$j$};
			
			  
			\draw (p1) edge[o->] (k);
			\draw (p2) edge[o->] (k);
			\draw (k) edge[->, red] (j);
			
			\begin{pgfonlayer}{background}
				\node[teal!80,  below=of j, yshift=1.8cm] (t2) {$\mathbf{S\cup\{j\}}$};
				\node[fit=(p1) (j) (p2),draw,teal,dashed,inner sep=5pt,fill=teal!20,fill opacity=0.3,rounded corners] {};
			\end{pgfonlayer}

		\end{tikzpicture}
   }
   
    \label{fig:proof_merging_fci}
 \caption{True DAG $G$ and PAGs $G_1$ and $G_2$ over $S\cup\{i\}$ and $S\cup\{j\}$ respectively. The red edge indicate a visible edge. There cannot be any edge between $i$ and $j$ due to the visible edge between $k$ and $j$.}
		\end{figure}
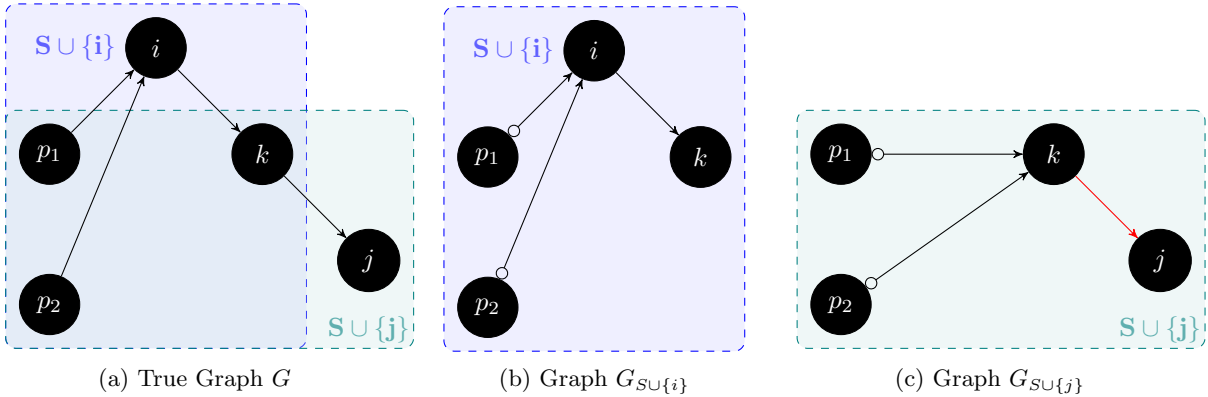
	
	More precisely, let $S:=\{k,p_1,p_2\}$ be a set of nodes with $i,j \not\in S$. Let $G_{1}$ 
	and $G_2$ denote the PAGs given by the asymptotic outputs 
	when FCI is applied to $S\cup \{i\}$
	and $S\cup \{j\}$, respectively, where we assume that the distribution is Markovian to the joint PAG in \cref{fig:fci_merging_true_graph}. 
	As visualized in \cref{fig:fci_merging_s_i}, $G_1$ consists of the edges $p_1 \to i$, $p_2 \to i$ (with circle marks) and $i\to k$ (without circle, since this link is recognized as visible). 
	Likewise, $G_2$ consists of 
	the edges $p_1 \to j$, $p_2 \to j$ (with circle marks) and $k\to j$ (without circle), where the latter is a visible edge, as e.g. the edge between $p_1$ and $k$ has an arrow head towards $k$ and $p_1$ is not adjacent to $j$.
    This graph is shown in  \cref{fig:fci_merging_s_j}. 
	We now conclude that there cannot be a direct link between $i$ and $j$ as follows: $i\to j$ and $i\to k$ would create an inducing path w.r.t. $i$ between $k$ and $j$ (which is ruled out by the visible edge), while 
	$j\to i$ would be a directed cycle. 
	We
	thus obtain $X_i\independent X_j\,|X_S$, which enables constructing the joint distribution from the two marginals.
    Precisely, we get
    \begin{displaymath}
        p(i, k, j, p_1, p_2) = p(i | k, p_1, p_2) p(j | k, p_1, p_2) p(k, p_1, p_2),
    \end{displaymath}
    where every factor on the right hand side is already given by the marginal distributions.

    \looseness=-1 We now need to show that there is another distribution $\tilde P$ that has the same marginals, but is ruled out by the self-compatibility constraint.
    We start by restricting the example from \cref{fig:fci_merging_true_graph} further to the linear Gaussian case.
    If we assume that all noise terms adhere to independent standard normal distributions and all structural coefficients are one, we get a distribution that is Markovian and faithful to $P$ and its covariance matrix is given via
    \begin{displaymath}
    \Sigma = ((I - A) (I - A)^T) ^{-1} = 
        \begin{pmatrix}
            1 & 0 & 1 & 1 & 1 \\
            0 & 1 & 1 & 1 & 1 \\
            1 & 1 & 3 & 3 & 3 \\
            1 & 1 & 3 & 4 & 4 \\
            1 & 1 & 3 & 4 & 5
        \end{pmatrix},
    \end{displaymath}
    where $A$ is the adjacency matrix of the graph $G$.
    The marginal covariance matrices follow directly form this:
    \begin{displaymath}
        \Sigma_{\{p_1, p_2, i, k\}} = 
        \begin{pmatrix}
            1 & 0 & 1 & 1 \\
            0 & 1 & 1 & 1  \\
            1 & 1 & 3 & 3 \\
            1 & 1 & 3 & 4  
        \end{pmatrix} \quad
        \Sigma_{\{p_1, p_2, k, j\}} = 
        \begin{pmatrix}
            1 & 0  & 1 & 1 \\
            0 & 1  & 1 & 1 \\
            1 & 1  & 4 & 4 \\
            1 & 1  & 4 & 5
        \end{pmatrix} 
    \end{displaymath}
    Finally, we define the matrix
    \begin{displaymath}
        \widetilde  \Sigma = 
        \begin{pmatrix}
            1 & 0 & 1 & 1 & 1 \\
            0 & 1 & 1 & 1 & 1 \\
            1 & 1 & 3 & 3 & 7/2 \\
            1 & 1 & 3 & 4 & 4 \\
            1 & 1 & 7/2 & 4 & 5
        \end{pmatrix}.
    \end{displaymath}
    $\widetilde \Sigma$ is also a symmetric, positive definite matrix and is therefore a valid covariance matrix.
    Further, it has the same marginal covariances over $\{p_1, p_2, i, k\}$ and $\{p_1, p_2, k, j\}$ as $\Sigma$.
    Therefore, the normal distribution with zero means and covariance matrix $\widetilde\Sigma$ is an example for a distribution $\tilde P$ that we were looking for.
    
\end{proof}

\section{GRAPHICAL COMPATIBILITY}
\label{sec:graphical_comp}
\subsection{Definitions for Further Graphical Models}
Now we want to define graphical notions for the remaining types of graphical models.

\begin{definition}[purely graphical compatibility] 
	A compatibility notion $c$ is called purely graphical if 
	it does not depend on $P_V$, that is, it can also be written as a function
	\[
	c: \cG_V^*  \to \{0,1\}.
	\]
\end{definition} 

We will present several purely graphical compatibility notions for different classes of graphical models respectively.
In \cref{def:graphical_comp} we have already defined graphical compatibility for ADMGs.
We will now define graphical compatibility for the other graphical models that we considered.

\begin{definition}[graphical compatibility of MAGs]
	Let  $S_1, \dots, S_k$ be $k\in \N$ sets of nodes and $G_{S_1},\dots,G_{S_k}$ be MAGs.
	Then we define graphical compatibility by the function $c$ with $c(G_{S_1},\dots,G_{S_k}) = 1$ iff there exists a set $V\supseteq \cup_{j=1}^k S_j$, and 
	an DAG $G_V$ such that
	$L(G_V, S_j) = G_{S_j}$, where $L$ is the latent projection from DAGs to MAGs as defined in \cref{def:latent_MAG}. 
\end{definition}

\begin{definition}[graphical compatibility of PAGs]
	Let  $S_1, \dots, S_k$ be $k\in\N$ sets of nodes and $G_{S_1},\dots,G_{S_k}$ be PAGs.
	Then we define graphical compatibility by the function $c$ with $c(G_{S_1},\dots,G_{S_k}) = 1$ iff there exists a set $V\supseteq \cup_{j=1}^k S_j$, and 
	a DAG $G_V$ such that
	$L^\text{PAG}(L^\text{MAG}(G_V, V), S_j) = G_{S_j}$, i.e. if $G_{S_j}$ represents the conditional independences of the DAG $G_V$ over $S_j$. 
\end{definition}
The definition of latent PAGs is centered around conditional independences and the causal semantics is not as obvious as for ADMGs. 
Note, that this is due to the fact that PAGs represent the causal statements that are consistent for \emph{all} models that entail the same independence structure (via \cref{thm:generalised_identification}).
Therefore we can also \enquote{safely} marginalise PAGs by only referencing the independence structure.

\begin{definition}[graphical compatibility of CPDAGs]
	Let  $S_1, \dots, S_k$ be $k\in\N$ sets of nodes and $G_{S_1},\dots,G_{S_k}$ be CPDAGs. Then we define graphical compatibility by the function $c$ with $c(G_{S_1},\dots,G_{S_k}) = 1$ iff there exists a set $V\supseteq \cup_{j=1}^k S_j$, and 
	an DAG $G_V$ such that
	$L(G_V, S_j) = G_{S_j}$, where $L$ is the latent CPDAG as defined in \cref{def:latent_projection_cpdag}. 
\end{definition}
\subsection{Critical Discussion of Graphical Compatibility}
\label{subsec:critical_graphical_comp}

Here we show that a purely graphical criterion implicitly relies on a genericity condition which is close to faithfulness in spirit. 
Consider, the causal model $Y\to Z$ and recall that ADMG compatibility excludes that an additional variable $X$ influences 
both $Y$ and $Z$. This conclusion can be criticised, depending on the precise interpretation of 
$Y\to Z$. Assume our interpretation of $Y\to Z$ reads: ``$Y$ is an unconfounded cause of $Z$ in the sense that $p(z|do(y))=p(z|y)$''. 
One can then construct a causal model with the complete DAG on $X,Y,Z$ in which the relation between $X$ and $Z$ is confounded by a hidden variable $W$. In a linear Gaussian model, we can easily tune parameters such that 
the confounding bias that $W$ induces  on $P(Y,Z)$ cancels out with the confounding bias induced by $Z$. This way, we still have  
$p(z|do(y))=p(z|y)$ despite the confounding paths. 
Excluding such a non-generic choice of parameters follows from faithfulness in a DAG 
with $X,Y,Z,W,F_Y$ where $F_Y$ controls  randomized interventions on $Y$ (the so-called `regime-indicator variable' of \cite{Dawid2021}). 
In our example,
vanishing confounding bias
corresponds to 
\[
F_Y \independent Z\,|Y, 
\]
without $d$-separation. 
In other words, the conclusion that $X\to Y$ together with the unconfounded causal relation $Y\to Z$ is incompatible with the complete DAG on $X,Y,Z$ relies
on a genericity condition against which one may raise doubts. 
After all, it is problematic to benchmark 
causal discovery algorithms
via methods that implicitly rely on principles close to faithfulness, if we on the other hand argue that assumptions like faithfulness are often violated. We cannot resolve this counter argument entirely. 
It may be reassuring, however, that also  model classes that do not rely on faithfulness come
to the same conclusion, that is, also exclude
the direct link from $X$ to $Z$. 
This will be shown in the proof of \cref{thm:idealized_rcd}. 

\section{Relationship between Interventional and Graphical Compatibility}
\label{sec:graphical_vs_interventional}
The next example shows a case, where interventional compatibility does not imply graphical compatibility.
\begin{example}[Non-generic confounder]
		Let there be an ADMG $G_1$ with variables $X, Y, Z$. Let $X$ consist of two components\footnote{One might ask, whether it makes more sense to treat $X_1$ and $X_2$ as separate variables, instead of a single variables with two components.
        Indeed, summarizing $X_1$ and $X_2$ as a single variable may seem a bit artificial.
		Though we want to note, that this is merely an illustrative example.
		The same phenomenon would e.g. also occur for a scalar variable, where $Y$ only depends on the first bit of the binary encoding and $Z$ only depends on the second bit.} $X_1, X_2$ with $X_1\ind X_2$ as in \cref{fig:two_components}. 
		If $Y$ only depends on $X_1$ and $Z$ only on $X_2$ we get
		\begin{align*}
			p(z\mid do(y)) &= \sum_x p(x) p(z\mid y, x) \\
			&= \sum_{x_1, x_2} p(x_1)p(x_2) p(z\mid y, x_2) \\
			&= \sum_{x_2} p(x_2) p(z\mid y, x_2) \\
			&= \sum_{x_2} p(x_2\mid y) p(z\mid y, x_2)= p(z\mid y).
		\end{align*}
        Now assume we have an ADMG $G_2$  that only contains $X\to Y$, i.e. implicity rules out confounding.
        Especially, $G_2$ implies 
        \begin{displaymath}
			p(z\mid do(y)) =  p(z\mid y).
		\end{displaymath}
        Therefore, the two models entail the same interventional statements and are interventionally compatible.
        Yet, $G_1$ and $G_2$ are not graphically compatible, as $G_2$ is different from the latent projection $L(G_1, \{X, Y\})$.
		\begin{figure}[htp]
  \subcaptionbox{Graph $G_1$
			\label{fig:two_components}}[0.4\textwidth]{
			\centering
			\begin{tikzpicture}
				\node[vardashed, minimum size=4cm] (X) {} ;
				\node[obs, left=of X.east] (X1) {$X_1$} ;
				\node[obs, right=of X.west] (X2) {$X_2$} ;
				\node[obs, below left=of X] (Y) {$Y$} edge[<-] (X1) ;
				\node[obs, below right=of X] (Z) {$Z$} edge[<-] (Y) ;
				\draw (X2) edge[->]  (Z);
			\end{tikzpicture}
   }
    \subcaptionbox{Graph $G_2$}[0.3\textwidth]{
			\centering
			\begin{tikzpicture}
				\node[obs] (Y) {$Y$} ;
				\node[obs, right=of Y] (Z) {$Z$} edge[<-] (Y) ;
			\end{tikzpicture}
   }
    \subcaptionbox{Graph $L(G_1, \{X, Y\}$}[0.3\textwidth]{
			\centering
			\begin{tikzpicture}
				\node[obs] (Y) {$Y$} ;
				\node[obs, right=of Y] (Z) {$Z$} edge[<-] (Y) ;
				\draw (Y) edge[<->, bend left=30]  (Z);
			\end{tikzpicture}
   }
  \caption{Non-generic case where interventional compatibility does not imply graphical compatibility.}
		\end{figure}
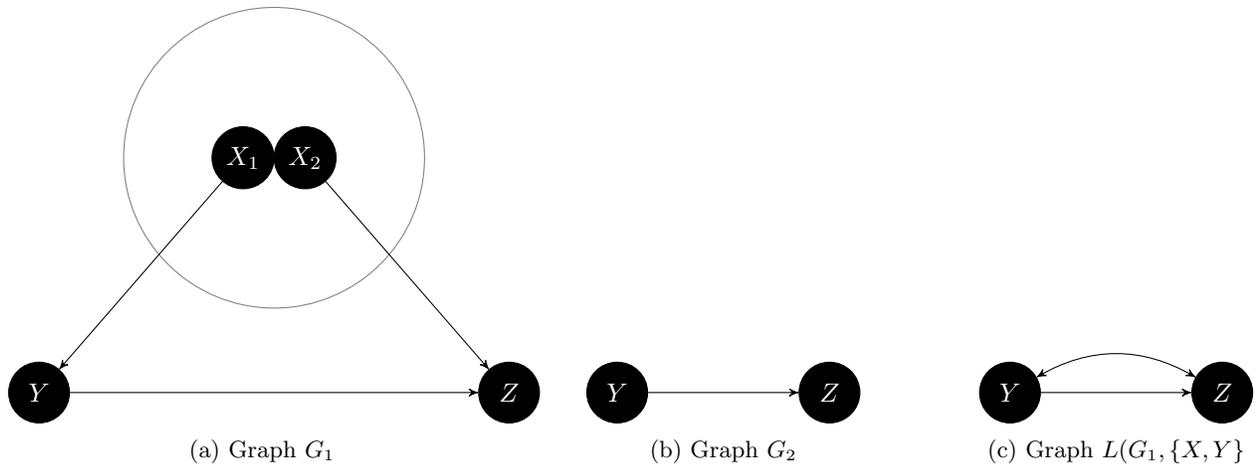
\end{example}
In the main paper we already mentioned that graphical compatibility also does not imply interventional compatibility.
In the following example we want to illustrate this in more detail.
\begin{example}[Empty graphs]
Let $G$ from \cref{fig:empty_graphs_true_dag} be the true underlying DAG for some distribution.
Assume the distribution is faithful to the DAG.
Further, let $\hat G$ and $\hat G_{X, Y}$ be the DAGs in \cref{fig:empty_graph_joint,fig:empty_graph_marginal}, respectively.
Clearly, $\hat G$ and $\hat G_{X, Y}$ are graphically compatible.
They imply
\begin{displaymath}
    p(y| do(x)) = p(y).
\end{displaymath}
But further, $\hat G$ also entails (as $Z$ is a valid conditioning set)
\begin{displaymath}
    p(y| do(x)) = \sum_z p(y | x, z) p(z),
\end{displaymath}
which is not equal to $p(y)$ for any distribution that is Markovian and faithful to $G$.
Thus, $\hat G$ and $\hat G_{X, Y}$ are not interventionally compatible.
		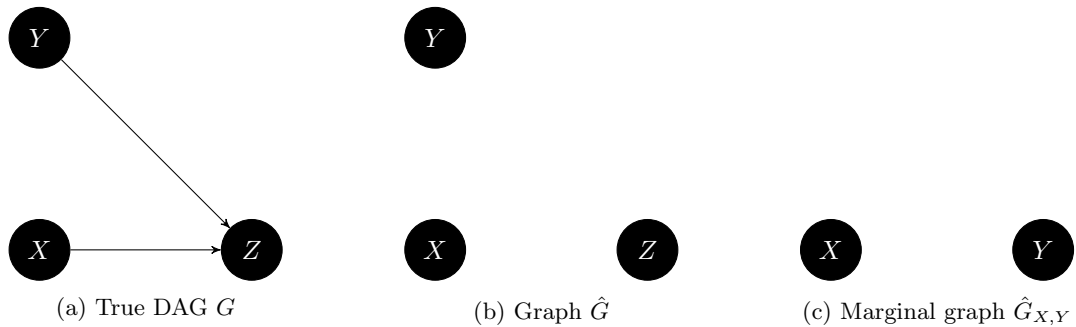
\begin{figure}[htp]
  \subcaptionbox{True DAG $G$\label{fig:empty_graphs_true_dag}}[0.3\textwidth]{
			\centering
   
			\begin{tikzpicture}
				\node[obs] (X) {$X$} ;
				\node[obs, above=of X] (Y) {$Y$} edge[->] (Z) ;
				\node[obs, right=of X] (Z) {$Z$} edge[<-] (X) ;
			\end{tikzpicture}
   }
    \subcaptionbox{Graph $\hat G$\label{fig:empty_graph_joint}}[0.3\textwidth]{
			\centering
			\begin{tikzpicture}
				\node[obs] (X) {$X$} ;
				\node[obs, above=of X] (Z) {$Y$};
				\node[obs, right=of X] (Z) {$Z$};
			\end{tikzpicture}
   }
    \subcaptionbox{Marginal graph $\hat G_{X, Y}$\label{fig:empty_graph_marginal}}[0.3\textwidth]{
			\centering
			\begin{tikzpicture}
				\node[obs] (Y) {$X$} ;
				\node[obs, right=of Y] (Z) {$Y$};
			\end{tikzpicture}
   }
  \caption{The graphical compatibility only depends indirectly on the data. For non-Markovian graphs, their graphical compatibility does not necessarily imply interventional compatibility.}
		\end{figure}
\end{example}
Indeed, if we require that the graphs are Markovian w.r.t. to the distribution at hand, we get that graphical compatibility implies interventional compatibility.
\begin{lemma}
    Let $S_1, \dots S_k$ be sets of variables for some $k\in\N$ and denote $S:= \bigcup_{i\in[k]} S_i$. 
	Let $G_{S_1}, \dots, G_{S_k}$	be an ADMGs or PAGs and $P_S$ be a probability distribution over $S$. 
    Let further $G_{S_i}$ be Markovian w.r.t. $P_{S_i}$ for all $i\in [k]$.
	Then, if $G_{S_1}, \dots, G_{S_k}$ are graphically compatible, they are also interventionally compatible w.r.t. $P_S$.
\end{lemma}
This follows directly from the construction of the latent projection in \cref{def:latent_projection_dag,def:latent_PAG}.

In \cref{def:interventional_comp} we did not require the marginal graphs to be Markovian w.r.t. the distribution.
With the following example we want to show why.
\begin{example}[Non-Markovian model]
Consider the linear Gaussian model with variables $X, Y, Z$ and structural coefficients as in \cref{fig:non_markovian_true_dag}.
The graph $\hat G$ in \cref{fig:non_markovian_marginal} is clearly not the latent projection of $G$ to $\{X, Z\}$.
Yet, both, $G$ (with the structural coefficients) and $\hat G$ imply 
\begin{displaymath}
    p(z| do(x)) = p(z).
\end{displaymath}
Therefore they are interventionally compatible.
With \cref{def:interventional_comp} we want to allow such non-generic cases, as long as the marginal model gets the interventional distributions \enquote{right} in the sense that they could have been created by a single joint model.
As we have mentioned before, this might not be the only appropriate choice, depending on the downstream task of the causal model.

    		\begin{figure}[htp]
  \subcaptionbox{True DAG $G$\label{fig:non_markovian_true_dag}}[0.5\textwidth]{
			\centering
   
			\begin{tikzpicture}
				\node[obs] (X) {$X$} ;
				\node[obs, right=of X] (Y) {$Y$};
				\node[obs, right=of Y] (Z) {$Z$};
				\draw (X) edge [->] node[above]{1} (Y);
				\draw (Y) edge [->] node[above]{-1} (Z);
				\draw (X) edge [->, bend left=30] node[above]{1} (Z);
			\end{tikzpicture}
   }
    \subcaptionbox{Graph $\hat G$\label{fig:non_markovian_marginal}}[0.5\textwidth]{
			\centering
			\begin{tikzpicture}
				\node[obs] (X) {$X$} ;
				\node[obs, right=of X] (Y) {$Z$};
			\end{tikzpicture}
   }

  \caption{A marginal model that is not Markovian w.r.t. to the distribution might still be interventionally compatible.}
		\end{figure}
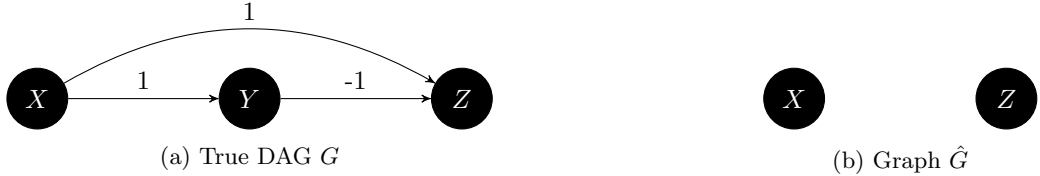
\end{example}




\section{WHICH ALGORITHMS ARE FALSIFIABLE?}
\label{sec:why_bivariate?}

\philipp{This section is WIP.}
In \cref{lem:algos_nonbi} we have seen that FCI and RCD are observationally falsifiable.
Other popular causal discovery algorithms that do not assume causal sufficiency are falsifiable as well.
We deferred the following lemma to the appendix to keep the presentation in the main paper more concise.
\begin{lemma}
\label{lem:falsifiable_alt_algos}
    The algorithms PC, GES \citep{chickering2002optimal} and DirectLiNGAM \citep{shimizu2011directlingam} are  falsifiable.
\end{lemma}

\begin{proof}
    Recall the graph from \cref{fig:motivation_SandT} and assume that this is the ground truth graph $G$.
	Also recall that we constructed a distribution that is Markovian and faithful to $G$ except for the additional independence $Y\ind Z_2$.
	We have already discussed that the PC algorithm will find the graphs in \cref{fig:motivation_SandT_false} in the population limit, as these are the only graphs that capture the conditional independences on the the subsets $X, Y, Z_1$ and $X, Y, Z_2$ under the assumption of no hidden confounder.
 
	Similarly, GES will find the same graphs in the limit of infinite data, as a graph that reflects exactly the independences in the distribution will have an optimal score\footnote{For simplicity we assume that the greedy search procedure always finds the optimal score.} (Proposition 8 \citep{chickering2002optimal}).
	Denote the graph that PC and GES find on $S$ with $G_S$ and the one over $T$ with $G_T$.
	In these graphs we get $p^S(y|do(x)) = p(y|x) \neq p(y) =  p^T(y|do(x))$ where $p^S(y|do(x))$ denotes the causal effect derived from $G_S$ using the identification formula from \cref{thm:generalised_identification} and analogously for $T$. 
	The inequality follows since we had  $Y\independent X$ otherwise, in contradiction to the assumption that $Y\ind Z_2$ is the only independence that is not entailed by the true DAG.\footnote{In this case we showed the falsifiability directly by constructing a joint distribution. But the example also shows that PC and GES are non-bivariate (as the marginal distribution over $S$ and $T$ could be interpreted as $P$ and $\widetilde P$ from \cref{def:nonbi}) and then \cref{thm:icommitt} could be applied.}

    For DirectLiNGAM, the same argument as in the previous proof for RCD (visualized in \cref{fig:proof_lingam_nonbi}) suffices.
\end{proof}

As the unfalsifiable algorithm in \cref{ex:entropy_ordering} showed, falsifiability is not a trivial property.
This raises the question if the falsifiable algorithms can be characterised differently.
With the following definition we want to exclude algorithms like the aforementioned ordering by entropy and indeed show that this is a sufficient criterion for falsifiability.
The definition has two aspects: 1) a non-bivariate algorithm must be able to produce an output that allows identification of a causal effect 2) this output does not only depend on bivariate properties of $X_i$ and $X_j$ but also depends on the distribution of the other nodes. 
\begin{definition}[non-bivariate causal discovery]\label{def:nonbi} 
	A causal discovery algorithm $\cA$ 	is \emph{non-bivariate} if there exists a set of variables $V$ with $|V| > 2$, as well as probability distributions $P_V$ and $\widetilde{P}_V$ over $V$ whose marginalisations to a subset of two variables $\{X_i, X_j\}\subseteq V$ coincide (i.e. $P_{\{X_i, X_j\}} = \widetilde{P}_{\{X_i, X_j\}}$) such that for every $\epsilon > 0$ there is a $m\in \N$ where for all $m' \ge m$  the following conditions hold with probability at least $1-\epsilon$:
	\begin{enumerate}
		\item $p(x_i | do(x_j))$ is identifiable in the estimated graphs $\cA(\bX^{m'})$ and $\cA(\widetilde{\bX}^{m'})$ and
		\item   there are identification formulae in $\cA(\bX^{m'})$ and $\cA(\widetilde{\bX}^{m'})$ respectively such that $p^{\cA(\bX^{m'})}(x_i | do(x_j)) \neq p^{\cA(\widetilde{\bX}^{m'})}(x_i | do(x_j))$, 
	\end{enumerate}
	where $\bX^{m'}$ denotes a data matrix with $m'$ samples drawn from $P_V$ and $\widetilde\bX^{m'}$ contains samples from $\widetilde P_V$ and $p^G(x_i\mid do(x_j))$ denotes the identification formula from $G$ applied to probability distribution $P$.
\end{definition} 
Note that we did not require the distributions to fulfil the causal discovery assumptions of $\cA$.

The following theorem asserts that for every non-bivariate causal discovery algorithm $\cA$ there is at least one distribution $P$ for which we can detect that $P$ does not fulfil the causal discovery assumptions of $\cA$ by applying $\cA$ to a subset of variables.
We observe:
\begin{theorem}[non-bivariate implies falsifiable]
	\label{thm:icommitt}  
	If $\cA$ is 
	non-bivariate,
	it is observationally falsifiable with respect to interventional compatibility.
\end{theorem}
\begin{proof}
	Let the set of variables $S$ and the distributions $P, \widetilde P$ be like in \ref{def:nonbi}. Let $\tilde{S}$ with $S\cap \tilde{S} = \{i,j\}$ 
	be such that $\tilde{S}\setminus \{X_i,X_j\}$ are variables of the same type as $S \setminus \{X_i,X_j\}$ with
	canonical one-to-one correspondence.
    We will assign different distributions to them now: define $P'$ to be such that $\widetilde P$ is \enquote{copied} to $\tilde S$, i.e. $X_i$ and $X_j$ have the same distribution like in $\widetilde P$ and all nodes in $\tilde S\setminus\{X_i, X_j\}$ have the same distributions as their corresponding node in $S$ has in $\widetilde P$. 
	Then define the joint distribution via
	\begin{equation}\label{eq:inconsjoint} 
		P(\bX_{S\cup \tilde{S}}) := P(\bX_{S\setminus \{i,j\}}|X_i,X_j)P(X_i,X_j) P'(\bX_{\tilde{S} \setminus \{i,j\}}|X_i,X_j).
	\end{equation}     
	One checks easily that its restrictions to $S$ and $\tilde{S}$ coincide with $P(\bX_S)$ and
	$\widetilde P(\bX_S)$, respectively. 
	By construction, the algorithm $\cA$ will result on contradictory statements 
	about the interventional distribution $p(x_i| do(x_j))$  when applied to $\bX_S$, versus when applied to $\bX_{\tilde{S}}$ in the limit of infinite data.  
	Thus we have $c(\cA(\bX_S), \cA(\bX_{\tilde S}), P_{S\cup \tilde S}) = 0$. 
\end{proof}

We now want to consider an example that is similar to the one in \cref{ex:entropy_ordering} but with very different conclusions.
\begin{example}[DAG Entropy Ordering]
    Define an algorithm $\cA $ that orders nodes according to their entropy and outputs the complete DAG with respect to that order.
\end{example}

This example is interesting for multiple reason.
First, note that the algorithm in this example is non-bivariate in the sense of our definition.
This illustrates that we refer to the non-bivariateness of interventional statements and not e.g. the resulting edges.
Second, it highlights the importance of condition 1) of \cref{def:nonbi}, as this is basically the only difference between the example above and \cref{ex:entropy_ordering}.
Third, the following lemma shows that this na\"ive algorithm is indeed falsified for almost all distributions.

\begin{lemma}[DAG via entropy order is almost always falsified]\label{lem:entralways} 
	Let $T:=\{1,2,3\}$ and 
	$P_{\{1,2,3\}}$ be generic 
	in the sense that
	the entropies of all variables are different. Assume 
	$H(X_1)< H(X_2)< H(X_3)$, without loss of generality, and
	assume the further genericity condition
	\begin{equation}\label{eq:do32} 
		\sum_{x_1} p(x_3|x_2,x_1)p(x_1)\neq p(x_3|x_2).
	\end{equation}
	Then the complete DAGs 
	$G_{\{1,2\}}$ and $G_{\{1,2,3\}}$, obtained by entropy ordering of nodes,
	are interventionally incompatible.
\end{lemma}

\proof{The left hand side of
	\eqref{eq:do32} is 
	$p(x_3|do(x_2))$ in $\cA(P_{\{1,2,3\}})$, while the right hand side is the same interventional probability in $\cA(P_{\{2,3\}})$. }
 \dominik{here we apply }
	
	Note that 
	$P_T$ in \cref{lem:entralways} 
	is always Markovian to 
	$\cA(P_{\{1,2,3\}})$ because this is a complete DAG. It is thus notable that the algorithm falsifies itself although $P_T$ is a distribution that is allowed by $\cA(P_{\{1,2,3\}})$. 
	
\section{FURTHER DISCUSSION OF THE RELATIONSHIP TO STABILITY}
	\label{sec:stability_results}
	In this section, we motivate our incompatibility score using stability arguments from learning theory \citep{shalev2010learnability}. Observe that, for a causal discovery method to achieve a low incompatibility score, it's output must remain largely unchanged under small modifications to the variable sets it is applied to. Following this idea, under some notion of stability of a causal discovery algorithm, we now show that stable algorithms, provably, generate \textit{useful causal models} in the sense described below.
	
	While, it is not possible to guarantee that a causal discovery algorithm that achieves a low incompatibility score will accurately predict system behavior under interventions, we argue that the resulting models are at least \textit{useful} due to their ability to predict statistical properties of unobserved joint distributions. This perspective is influenced by \citet{janzing2023reinterpreting}, who reconceptualizes causal discovery as a \textit{statistical learning problem}. The key principle underlying this reconceptualization posits that causal models offer predictive value beyond predicting system behavior under interventions; they can also predict statistical properties of unobserved joint distributions. To illustrate the main idea, consider the following example.

	Consider a set of variables $\mathcal{X} = \left \{ X_1, X_2, \cdots X_n \right \}$, and let $\mathcal{S}$ represent a collection of subsets of $\mathcal{X}$. A statistical property $Q$ can be defined as a mapping from $\mathcal{S}$ to $\mathbb{R}$. For a given tuple $S_i = (X_{l_1}, X_{l_2}, \cdots X_{l_k})$, $Q$ might indicate whether the conditional independence $X_{l_1} \independent X_{l_2} \vert X_{l_3}, \cdots X_{l_k}$ holds, represented as $Q(S_i) = 1$ (if true) or $Q(S_i) = 0$ (if false).\footnote{Empirically, it can also indicate whether a given test $T$ detects the conditional independence under consideration based on datasets of observations corresponding to the variable set.} 
	
	Causal models such as Directed Acyclic Graphs (DAGs) can thus be viewed as predictors of statistical properties, like conditional independences. The statistical property predicted by a causal model $\mathcal{M}$ is denoted as $\widehat{Q}_{\mathcal{M}}: \mathcal{S} \rightarrow \mathbb{R}$. \footnote{For a more detailed discussion, formalism, and justification of this problem we refer the reader to \citet{janzing2023reinterpreting}.}
	
	The statistical prediction problem can now be outlined as follows: Given a set of $m$ observations $\mathbb{S} = \left \{ S_1, S_2, \cdots S_m) \right \} \in \mathcal{S}^m$,
	and a loss function $l: Q(\mathcal{S}) \times Q(\mathcal{S}) \rightarrow [0, 1]$, the goal of causal discovery is to learn a joint causal model over $\mathcal{X}$ that minimizes the loss $l(\widehat{Q}_{\mathcal{M}}(S), {Q}(S))$ on unobserved subsets $S \in \mathcal{S}$. As an illustrative example, the PC algorithm constructs a joint causal model across all variables—encoding conditional independences for any subset of $\mathcal{X}$—by evaluating conditional independences within a selected set of smaller subsets.

	If we assume that the subsets are drawn according to a distribution over $\mathcal{S}$, we can invoke stability arguments from learning theory to guarantee generalization from observed to unobserved sets of variables for causal discovery algorithms that demonstrate {stability under small modifications to variable sets}. This is different from the standard setting in statistical learning, where algorithms that exhibit stability under {small modifications to the data} are known to \textit{generalize across data points} \citep{shalev2010learnability}.

	In order to formalize our discussion, let's define the concept of 'stability' in the context of causal discovery. While there are many related notions of stability in statistical learning theory, their relevance varies depending on the context. In our case, Leave-one-out (LOO) stability \citep{mukherjee2006learning}, which requires stability of the loss when a single data point  (or a subset of data points) is included or excluded from the training set, is particularly applicable. Nonetheless, to maintain a high-level perspective, we will introduce a strong, distribution-independent form of stability: uniform stability \citep{bousquet2002stability}, which encompasses other weaker forms of stability like LOO stability. \dominik{not clear what uniform stability means in the traditional setting}
	
	\begin{definition}[\textbf{Uniform Stability}]
		A causal discovery algorithm $A$ is said to be $\gamma-$uniformly stable with respect to the loss function $l$ if, for any $S \in \mathcal{S}$, any subset $\mathbb{S} = \left \{ S_1, S_2, \cdots S_m \right \} \subset \mathcal{S}^m$, and for any index $i \in [m]$, the following inequality holds:
		\begin{equation*}
			\vert l(\widehat{Q}_{A_{\mathbb{S}}}(S), {Q}(S)) - l(\widehat{Q}_{A_{\mathbb{S}^{/i}}}(S), {Q}(S))\vert \leq \gamma,
		\end{equation*}
		where $S^{/i}$ denotes the set $\mathbb{S}$ replacing the element $S_i$ by some $S'_i \in \mathcal{S}$.
	\end{definition}
	
	With this definition, we can formally prove that stable causal discovery algorithms generate useful causal hypotheses in the sense of their ability to generalize statistical predictions across variable sets. To this end, we first introduce the notions of empirical and true risks and state the key result in Theorem \ref{thm:stability}.
	
	\begin{definition}[\textbf{Empirical and Expected Risks}]
		Under the reconceptualization of causal discovery as a learning problem, assuming that the subsets are drawn according to a distribution $\mathbb{L}$ over $\mathcal{S}$, the empirical ($\widehat{R}(\mathcal{M})$) and expected risks ($R(\mathcal{M})$) incurred by a causal model $\mathcal{M}$ are defined as:
		
		\begin{equation*}
			\widehat{R}(\mathcal{M}) \coloneqq \frac{1}{n} \sum \limits_{i=1}^m  l(Q_\mathcal{M}(S_i), Q(S_i)), \quad R(\mathcal{M}) \coloneqq \mathbb{E}_{\mathbb{L}} [ l(Q_\mathcal{M}(S_i), Q(S_i))].
		\end{equation*}
	\end{definition}
	
	Under this definition, one can leverage recent results from \citet{bousquet2020sharper} to derive the following high probability generalization bounds for uniformly stable causal discovery algorithms.
	\begin{theorem}[\textbf{Generalization bounds for uniformly stable causal discovery}]
		\label{thm:stability}
		Let $A$ denote a causal discovery algorithm that is $\gamma-$uniformly stable with respect to the loss function $l$ in the sense described above. Then there exists a constant c such that for any probability distribution $\mathbb{L}$ over $\mathcal{S}$ and for any $\delta \in (0, 1)$, the following holds with probability at least $1 - \delta$ over the draw of $m$ subsets $S_1, S_2, \ldots, S_m$ according to $\mathbb{L}$:
		\begin{equation*}
			R(A(\mathbb{S})) - \widehat{R}(A(\mathbb{S}))  \leq c \Big ( \gamma \log (n) \log (1/\delta) + \frac{\sqrt{\log (1/\delta)}}{\sqrt{n}} \Big),
		\end{equation*}
		
		where $A(\mathcal{S})$ denotes the causal model output by the causal discovery algorithm $A$ applied on the set $\mathbb{S} = \left \{ S_1, S_2, \cdots, S_m \right \}$.
	\end{theorem}
	
	The result demonstrates that stable algorithms, provably, generate useful causal models due to their ability to \textit{generalize statistical predictions across variable sets}. Informally, it provides evidence that a low incompatibility constitutes a useful inductive bias for causal discovery. This is notably distinct 
 from the standard setting in statistical learning, where algorithms that exhibit stability under {small modifications to the data} are known to \textit{generalize across data points}.
	
\section{ADDITIONAL RESULTS ABOUT MERGING}
	\label{sec:is_compatibitlity_strong?}
	
	In \cref{thm:fci_merging} we have seen that FCI enables what we have called \emph{merging} in \cref{def:merging}.
	We now want to show that an idealized version of RCD also has this property. 
	Repetitive Causal Discovery (RCD) \citep{maeda2020rcd} is 
	a based on the LiNGAM-assumptions and 
	is able to also infer the presence of latent confounders.
	It assumes a linear model with independent non-Gaussian noise, where 
	confounding is modelled by some shared noise-variables. Explicitly, it reads:
	\begin{equation}\label{eq:rcd}
		X_i = \sum_{j}  \beta_{ij} X_j + \sum_j \epsilon_{ij} W_j+  N_i,   
	\end{equation}
	where all $N_i,W_j$ are independent noise variables (with non-zero variance). The variables $W_j$, which are shared by at least two $X_i$, describe the confounding.   
	\begin{definition}[idealized RCD]
		We define \emph{idealized RCD} to be a causal discovery algorithm that has an additional output token $\bot$, that indicates that idealized RCD 'abstain from a decision', i.e. it indicates that (idealized RCD estimated\footnote{We assume that the idealized algorithm estimates this in an \enquote{oracle}-fashion, i.e. we do not discuss how this could be estimated.} that) the distribution cannot be generated via \cref{eq:rcd}. 
		Otherwise, it
		draws an arrow $X_i\to X_j$ whenever $\beta_{ij}\neq 0$, and a bidirected
		link whenever they share a variable $W_k$, that is, there exists a $k$
		such that $\epsilon_{ik}\neq 0$
		and  $\epsilon_{jk}\neq 0$ according to the usual RCD algorithm.
		
	\end{definition}
	\begin{theorem}[idealized RCD enables merging] 
        \label{thm:idealized_rcd}
		Idealized RCD enables merging with respect to graphical consistency. 
	\end{theorem}
	
	\proof{
		For the variables $X_1,X_2,X_3$,
		assume that idealized RCD outputs
		$X_1\to X_2$ (without confounding) asymptotically when applied
		to data from $P_{\{1,2\}}$. Assume further, that it outputs
		$X_2\to X_3$ (without confounding) when applied to data from $P_{\{2,3\}}$. We now show that
		applying RCD to all three variables
		can only yield compatible results 
		if $X_1\independent X_3\,|X_2$
		and thus 
		\[
		P(X_1,X_2,X_3) = P(X_1,X_2) P(X_3|X_2).
		\]
		Consequently, the joint distribution follows
		uniquely from the two marginal distributions. 
		The proof builds heavily on the theorem of Darmoir-Skitovic \citep{Darmois,Skitovic}, which entails that
		for any set of independent non-Gaussian variables $Y_1,\dots,Y_d$ with non-zero variance, we have 
		\[
		\Big(\sum_j a_j Y_j \Big) \quad \independent \quad \Big(\sum_j b_j Y_j \Big)\quad \Rightarrow
		a_j \cdot b_j =0 \quad \forall j.  
		\] 
		In other words, two linear combinations can only be independent if they share none of the variables. 
		
		If the output of RCD on
		the joint distribution $P_{\{1,2,3\}}$ is compatible
		with the two marginal models, it needs to 
		be described 
		by a model of the form
		\eqref{eq:rcd} with causal ordering
		$1,2,3$. For any different order, causal directions would not be compatible with the marginal models.
		Further, RCD would output $\bot$ if the joint model was not of the form \eqref{eq:rcd}, which we also count as incompatibility.
		Hence, compatibility entails a joint model of the following form:
		\begin{eqnarray}\label{eq:rcd1} 
			X_1 &=&  \sum_j \epsilon_{1j} W_j + N_1\\ \label{eq:rcd2}
			X_2 &=& \beta_{21} X_1 + \sum_{j} \epsilon_{2j} W_j +  N_2 \\
			X_3 &=& \beta_{31} X_1 + \beta_{32} X_2 + \sum_j \epsilon_{3j} W_j + N_3.
			\label{eq:rcd3}
		\end{eqnarray}
		For all $j$ we have
		$\epsilon_{1j} \epsilon_{2j} =0$
		otherwise the causal relation $X_1\to X_2$ would be confounded and RCD could not output an unconfounded link 
		as bivariate model.
		This can be seen as follows. 
		RCD only outputs 
		an unconfounded link $X_1\to X_2$
		if $X_2- \alpha X_1$ is 
		independent of $X_1$ for
		some $\alpha$. 
		This can only be true for 
		$\alpha=\beta_{21}$, otherwise both expressions contain $N_1$. 
        Further, $X_2-\beta_{21} X_1= \sum_{j} \epsilon_{2j} W_j+ N_2$ can
		only be independent of 
		$\sum_j \epsilon_{1j} W_j + N_1$ if the linear combinations share no $W_j$. 
		In a similar way we conclude
		that $\epsilon_{2j} \epsilon_{3j} =0$:
		We first check that
		$X_3 - \alpha X_2$ can only be independent of $X_2$ for $\alpha =\beta_{32}$, otherwise $N_2$ appears
		in both expressions. 
		Further,
		$X_3 - \beta_{32}X_2 $
		can only be independent from $X_2$ if they share none of the variables $W_j$. 
		
		The variables $W_j$ thus fall into three classes: those that appear in only one of the variables $X_1,X_2,X_3$ and those that are shared by $X_1$ and $X_3$. The former ones can be absorbed into the noise variables $N_j$. We thus simplify 
		\eqref{eq:rcd1} to \eqref{eq:rcd3} 
		to 
		\begin{eqnarray}\label{eq:rcd1s} 
			X_1 &=&  \sum_j \epsilon_{1j} W_j + N_1\\ \label{eq:rcd2s}
			X_2 &=& \beta_{21} X_1  +  N_2 \\
			X_3 &=& \beta_{31} X_1 + \beta_{32} X_2 + \sum_j \epsilon_{3j} W_j + N_3.
			\label{eq:rcd3s}
		\end{eqnarray}
		In a similar way as we have repeatedly argued, $X_3-\alpha X_2$ can only be independent of $X_2$ if $\alpha=\beta_{31}$.
		We obtain
		\begin{eqnarray}
			X_3 - \beta_{32} X_2 &=&  \beta_{31} \sum_j \epsilon_{1j} W_j + \sum_j \epsilon_{31} W_j +  \beta_{31} N_1  + N_3.   
		\end{eqnarray}
		This expression can only be independent of $X_1$ if $\beta_{31}=0$,
		otherwise $N_1$ appears in both expressions.
		The remaining term $\sum_j \epsilon_{31} W_j+N_3$
		can only be independent of $X_1$ if 
		there is no $W_j$ shared by $X_1$ and $X_2$, i.e., $\epsilon_{1j} \epsilon_{3j}=0$ for all $j$. 
		However, then the respective linear combination of all $W_j$ that appear only in $X_1$ can be merged with $N_1$, and the others with $N_3$.
		Thus, we end up with the structural equations
		\begin{eqnarray}
			X_1 &=&  N_1\\ 
			X_2 &=& \beta_{21} X_1  +  N_2 \\
			X_3 &=&  \beta_{32} X_2 +  N_3,
		\end{eqnarray}
		which implies $X_1 \independent X_3\,|X_2$. 

        We now want to construct a different distribution $\tilde P$ that has the same marginals as $P$, to show that without the self-compatibility constraint the solution to the statistical marginal problem is not unique. 
        We will construct an SCM with the edges $X_1 \leftarrow X_2 \to X_3$ and $X_1\to X_3$ and violate the assumption of a linear model with additive noise.

        We first reconstruct the marginal $P_{1, 2}$ using a construction from proposition 4.1 by \citet{peters2017elements}.
        Define the conditional cumulative distribution function 
        \begin{displaymath}
            F_{Y|x}(y) := P(Y\le | y \mid X = x)
        \end{displaymath}
         and further define
        \begin{displaymath}
            F^{-1}_{Y|x}(n_Y) := \inf\{y\in \R : F_{Y|x}(y) \ge n_Y\}.
        \end{displaymath}
	}
        Now set $X_1$ via the structural equation
        \begin{displaymath}
            X_1 = f(X_2, \tilde N_1),
        \end{displaymath}
        where $f(x_2, n_1) = F^{-1}_{Y|x}(n_1)$ and $\tilde N_1$ is uniformly distributed over $[0, 1]$ and independent from $X_2$ and $P_2 = \tilde P_2$.
        By construction we get $\tilde P_{1, 2} = P_{1, 2}$.

        We now set $X_3$ via the structural equation
        \begin{displaymath}
            X_3 = \beta_{32} X_2 + F^{-1}_{N_3}(\tilde N_1),
        \end{displaymath}
        where $\beta_{32}$ is the underlying structural coefficient between $X_2$ and $X_3$ from $P$ and $F^{-1}_{N_3}$ is the quantile function of the noise term $N_3$.
        Again, by construction we get the same marginal as in $P$.
        But clearly, in $\tilde P$ we do not have $X_1\ind X_3| X_2$, as the noise term of $X_3$ is a deterministic function of the noise term of $X_1$.

\section{PRACTICAL EVALUATION OF COMPATIBILITY}
	\label{sec:practical_compatibility_scores}
	\subsection{Details of Interventional Incompatibility Score}
	We have already introduced our interventional compatibility score in \cref{def:interventional_score}.
	In this section we want to shortly elaborate on this score to avoid confusion.
	
	\citet{su2022robustness} propose to falsify interventional statements of a linear causal model by comparing the interventional distributions entailed by different adjustment sets.
	More precisely: in many cases a causal DAG $G$ implies several sets $C_1, \dots, C_k$ for $k>2$ satisfying the backdoor criterion \cite{pearl2009causality} such that 
	\begin{displaymath}
		p(y\mid do(x)) = \sum_{c_i} p(y\mid x, c_i) p(c_i),
	\end{displaymath}
	for all $i\in[k]$.
	\Cref{fig:motivation_SandT} is a (rather trivial) example, as we could use $\emptyset$ and  $\{Z_1\}$.
	The set of parents of $X$ is always such an adjustment set if the effect is identifiable \citep{tian2002general} in an ADMG.
	If $G$ is the true causal model, we can decide whether a set $C$ is a valid adjustment set by graphical criteria (see \cref{thm:generalised_identification}).
	\cite{su2022robustness} now propose a statistical test to reject the null hypothesis $H_0$ that all sets under consideration yield the same interventional distribution.
	Precisely, for  sets of variables $C_1, \dots, C_k$  the hypothesis reads
	\begin{displaymath}
		H_0:\quad \beta_{X_i, X_j\cdot C_1} =  \dots = \beta_{X_i, X_j\cdot C_k},
	\end{displaymath}
	where $\beta_{X_i, X_j\cdot C_l}$ denotes the partial regression coefficient for $X_i$ regressed on $X_j$ and all variables in $C_l$ for some $l\in[k]$. Their test assumes linear causal models and Gaussian noise.
	
	We build on their work in the following sense: instead of using one causal model $G$ to derive multiple adjustment sets, we use adjustment sets of \emph{different} causal models, learned on different subsets of variables.
	Just as in their work, they should all yield the same interventional distributions.
	For simplicity, we use parent-adjustment for each marginal model in the case of RCD and the canonical adjustment set for FCI.
	Note, that in the case where all models are correct, each adjustment set in a marginal model will be a valid adjustment set in the joint model.
	But that the marginal parent sets are valid adjustment sets in the joint model is neither sufficient nor necessary for the test to accept.

	\subsection{Further Practical Considerations}
	In the following we want to specify some aspects of \cref{def:interventional_score,def:graphical_score} that have not been discussed in detail and also slightly modify \cref{def:graphical_score} further to tackle some practical issues.
	
	\paragraph{Sampling of subsets.}
	In \cref{def:interventional_score,def:graphical_score} we have assumed the subsets are given. 
	In practice, we sample them randomly.
	But we did not sample from the set of all possible substes for the following reason:
	if a subset is very small, some algorithms like the FCI algorithm often give uninformative outputs in the sense that most edges are unoriented.
	On the other hand, if subsets are large and differ only by few nodes, the resulting marginal models usually do not differ much.
	Therefore, in all experiments, we uniformly drew subsets $S_i$  from the set of subsets with $|S_i| = \lceil |V| / 2\rceil$ for $i=1, \dots, l$ for some $l\in \N$.

    \paragraph{Canonical adjustment and false positives}
    In our experiments with FCI we noted that many tests indicated 
    a difference in the interventional distributions, despite the fact that the graphical models seemed to be Markovian and graphically compatible.
    This seemed to be due to the fact that the canonical adjustment set is usually quite large, rendering the problem statistically hard.
    In cases where there was no possibly directed causal path (and therefore the model graphically implied no dependence for any adjustment set) we chose to replace the canonical adjustment set with a randomly chosen subset of size one. 

    \paragraph{Marginalisation of RCD outputs}
    As we have noted before, the outputs of RCD are not ADMGs, as RCD cannot differentiate between the case, where a node $X$ has a direct edge to $Y$ \emph{and} they have an unobserved hidden confounder or where they are confounded without any directed edge between them.
    During marginalisation we treated these bidirected edges accordingly, and drew additional edges if one node is a \emph{potential} ancestor or an additional bidirected edge if two nodes are \emph{potentially} confounded.

    	\paragraph{Causal sufficiency.}
	In practice, many algorithms like the PC algorithm or GES rely on the assumption that all causally relevant variables are observed, i.e. that there is no common cause between two observed variables that is itself unobserved and all directed paths from this common cause to the observed nodes only contains unobserved nodes. 
	Clearly, if we sample the subsets uniformly from all subsets with the same size, causal sufficiency will often be violated for these subsets, even if it holds for $V$. 
	Consequently, if we detect an incompatibility between $\cA(\bX)$ and $\cA(\bX_S)$ we cannot know whether this indicates an actual error  or it is simply due to the newly introduced hidden confounder.
	Still, in \cref{sec:experiments} we will show experiments with PC and GES that indicate that the graphical incompatibility score might help in model selection for these algorithms.
    The interventional criterion seems to be more sensitive to these violations of sufficiency.
	
	\paragraph{CPDAGs and hidden confounders.}
	In \cref{def:latent_projection_cpdag} we have only defined the latent projection of  DAG to a CPDAG on causally sufficient subsets.
	But as we have just discussed, we will not only look at causally sufficient subsets.
	In the experiments, we simply calculated the latent ADMG and deleted its bidirected edges.
	The resulting graph is a DAG again and we proceeded with its respective CPDAG.

\subsection{Runtime of the Incompatibility Scores}
Let $\cA$ be an algorithm, $k$ be the number of considered subsets, $m$ be the number of samples, $f(n, m)$ be the worst-case run time of $\cA$ on $n$ nodes and $m$ samples, $g(k, n, m)$ be the time of the test by \citet{su2022robustness} and $h(n)$ be the time to calculate the latent projection from a graph with $n$ nodes.
We then get a run time in 
\begin{displaymath}
    \mathcal{O}(f(n, m) + k \cdot f(\lceil n / 2\rceil, m) + n^2 \cdot (g(k, n, m) + k\cdot h(n))), \quad
    \mathcal{O}(f(n, m) + k \cdot (f(\lceil n / 2\rceil, m) + n^2 + h(n))),
\end{displaymath}
for $\kappa^I$ and $\kappa^G$ respectively. 
With our simple implementation, $h(n) \in \mathcal{O}(n^3)$.
As $g$ is polynomial in $k, n$ and $m$, the run time in our main experimental setting  is dominated by $f$ which is exponential in $n$ for FCI and RCD.

\section{EXPERIMENTAL DETAILS}
	\label{sec:exp_details}
	\subsection{Data Generation}
	For our first experiments, we generated synthetic data.   
	We first sampled a random ground truth graph using the Erdos-Renyi model (with number of nodes $n+h$ and expected degree $d$, where $h$ is the number of potential hidden confounders).
	For each node $X$ we then define a functional model from the class of linear models. 
	For the linear model we drew the parameters uniformly from $[-1, -0.1] \cup [0.1, 1]$.
	We then apply an additive noise term which is either drawn from a standard normal distribution or a uniform distribution with zero mean and unit variance.
    We then randomly pick $d$ nodes as observed nodes and marginalise out the others.
	
	In all experiments we used graphs with 10 nodes and expected degree 2, as well as linear Gaussian and linear uniform structural equations.
	For all incompatibility scores we drew 40 subsets uniformly from the set of all subsets of size 5. For experiments with RCD and FCI we set $h=3$, otherwise $h=0$.
 For all experiments we draw 1000 samples from the SCM.
	
	\subsection{Other Details}
	We evaluated the SHD of CPDAGs and PAGs with respect to the respective CPDAG and PAG of the ground truth graph and for PAGs, according to the definition of \citet{triantafillou2016score}.
	For the test of \citet{su2022robustness} in the interventional compatibility score we always chose the confidence level $0.001$.
    For RCD and FCI we also set all confidence thresholds to $0.001$, unless stated otherwise.
    Similarly, for PC and GES we set the parameters of the algorithms to $\alpha=0.01$  and $\lambda = 0.01$, respectively, unless stated otherwise.
    For all algorithms, we used the implementation from the \texttt{causal-learn} python package \citep{causallearn}.

	The computations were done on an Intel Core i5-5200U CPU with 8 GB RAM or an Apple M1 Pro with 32 GB of RAM. All experiments can be run in less than a day.
	
\section{ADDITIONAL EXPERIMENTS}
	\label{sec:additional_experiments}
\subsection{Additional Plots}
\begin{figure}
		\centering
		\includegraphics[width=.4\textwidth]{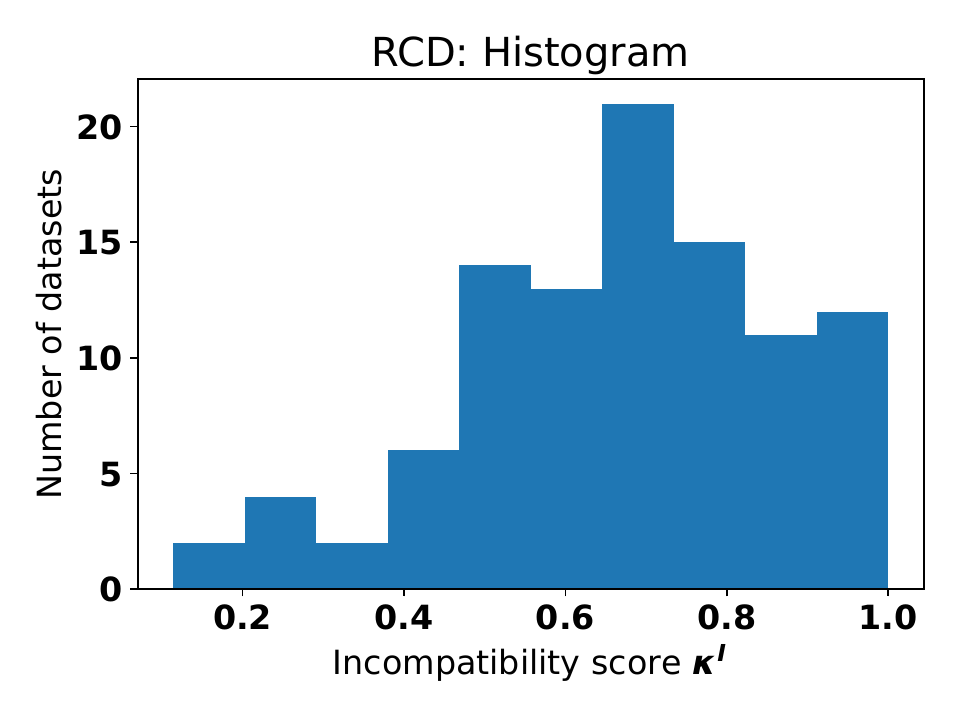}
		\includegraphics[width=.4\textwidth]{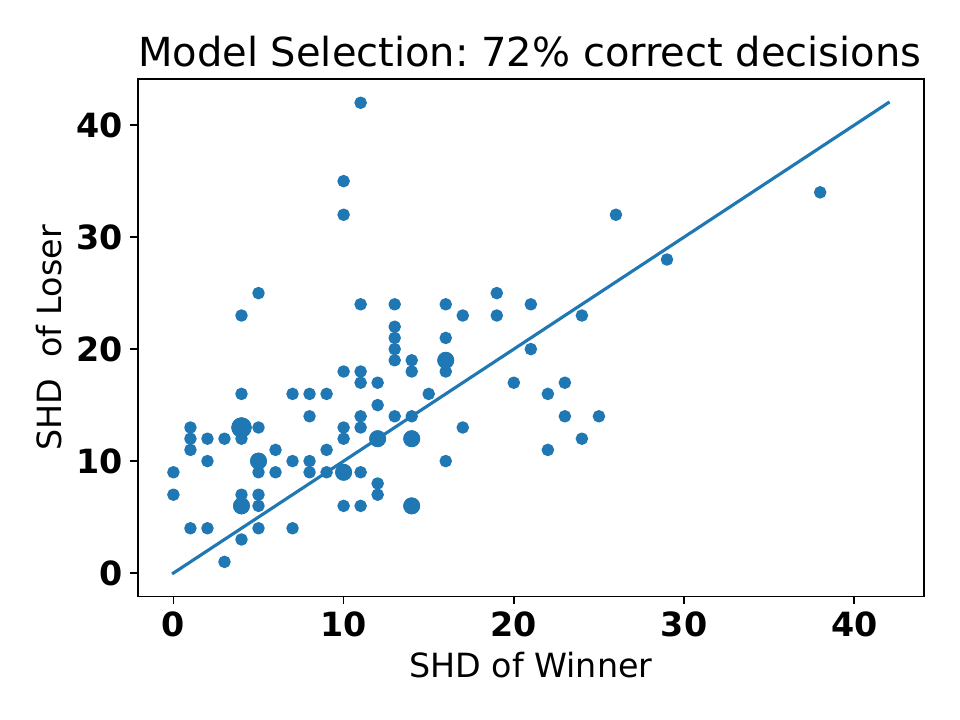}
		\caption{(Left) A histogram of $\kappa^I$ values for RCD on 100 linear uniform datasets. (Right) The incompatibility score $\kappa^I$ as metric for model selection with RCD and $\alpha=0.1$ or $\alpha=0.001$. We picked strictly better parameters for 68\%
of datasets and for 28\% we picked strictly worse parameters. Overall, in 72\% of datasets we picked better or equally good parameters.}
		
		\label{fig:additional_main_plots}
	\end{figure}

In the experiments in \cref{fig:rcd,fig:model_selection} we have studied the behaviour of the RCD algorithm and the interventional incompatibility score $\kappa^I$.
The plots in \cref{fig:additional_main_plots} do not contain novel insights about the experiments.
Yet, the visualization emphasises slightly different aspects.
For comparability, we also report them for the following experiments.
The \enquote{Winners} in the right plot are determined as follows:
\begin{displaymath}
		\operatorname{Winner} = \operatorname{argmax}_{{\cA\in \{\operatorname{RCD}^{\alpha=0.1}, \operatorname{RCD}^{\alpha=0.001}\}}} \kappa^I(\cA, \bX).
\end{displaymath}
The loser are the respective other parameters. The defenitions of \enquote{Winners} and \enquote{Losers} in the following plots is analogous.

   	\begin{figure}
		\centering
		\includegraphics[width=\textwidth]{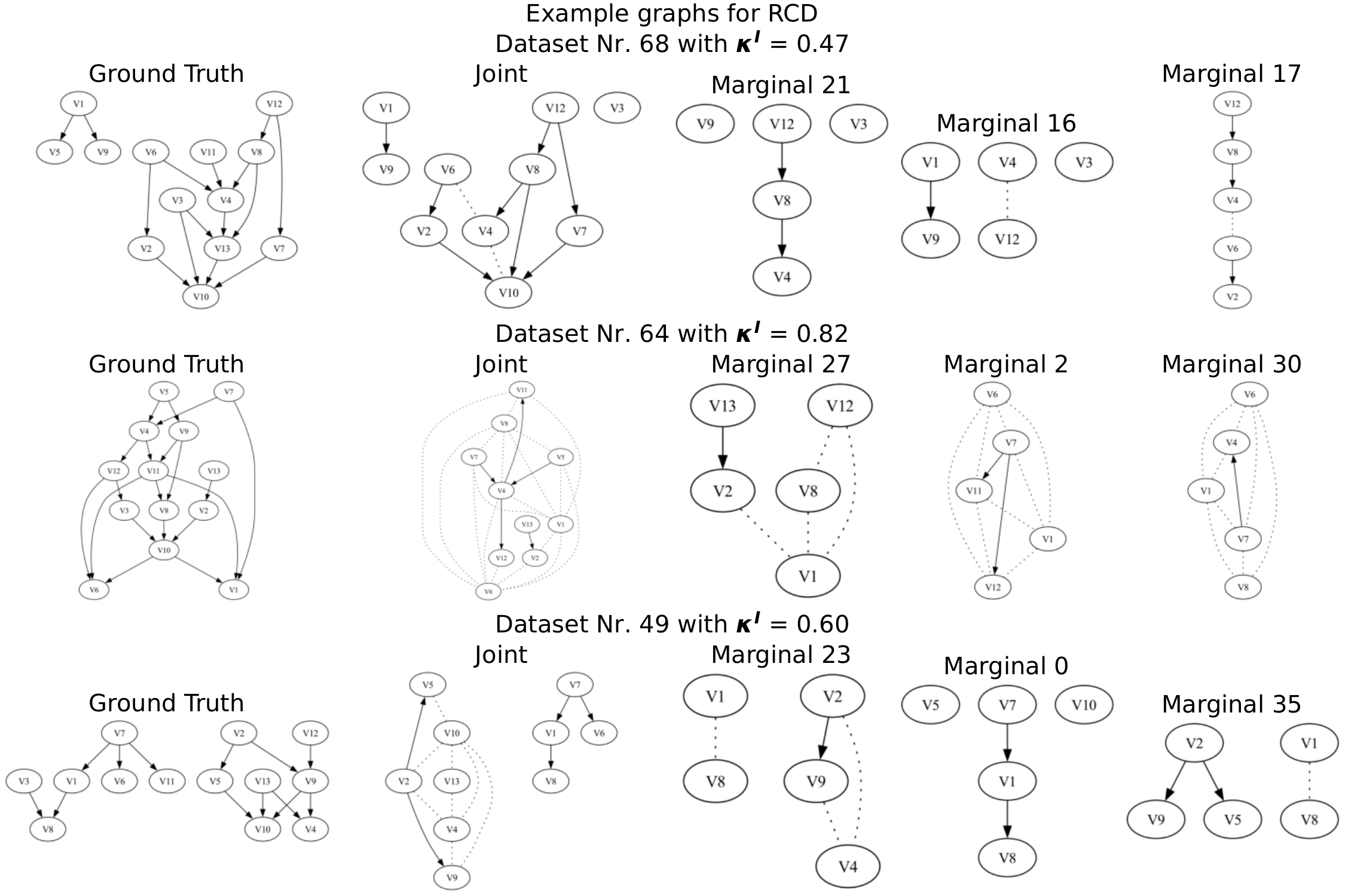}
		\caption{Randomly drawn example graphs from the experiment shown in \cref{fig:rcd}. The figure shows for three randomly picked datasets the ground truth graph (including hidden variables), the joint graph found be the algorithm on all variables and some randomly drawn marginal graphs, i.e. graphs that the algorithm found on subsets of variables. Dotted edges indicate that RCD could not infer the direction of the edge.}
		
		\label{fig:example_graphs_rcd}
	\end{figure}

	\subsection{Model Evaluation}
\begin{figure}
		\centering
		\includegraphics[width=.4\textwidth]{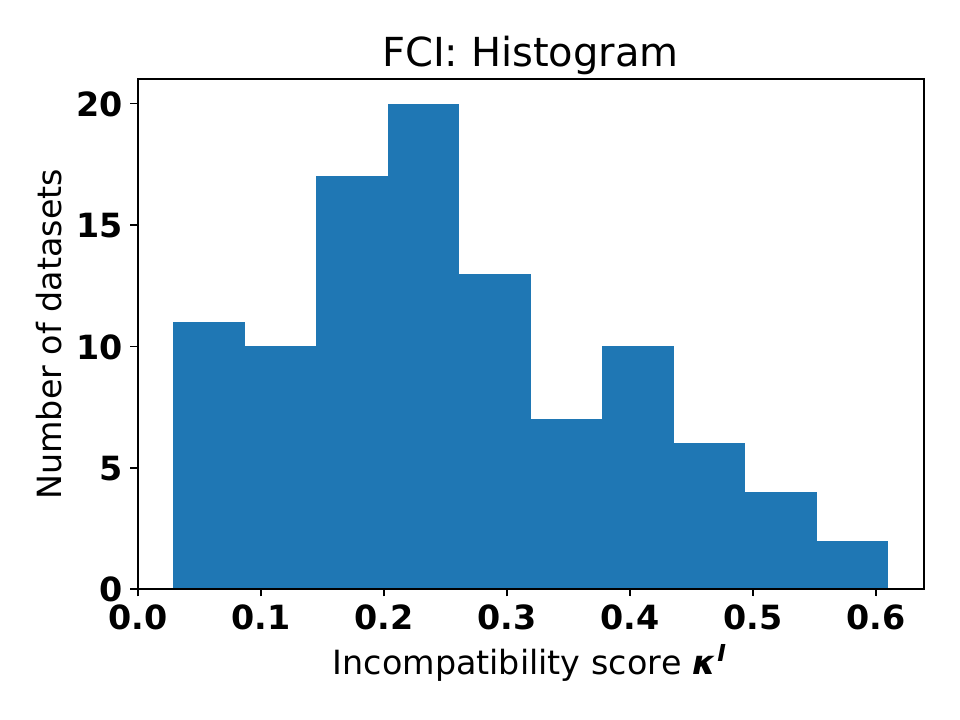}
		\includegraphics[width=.4\textwidth]{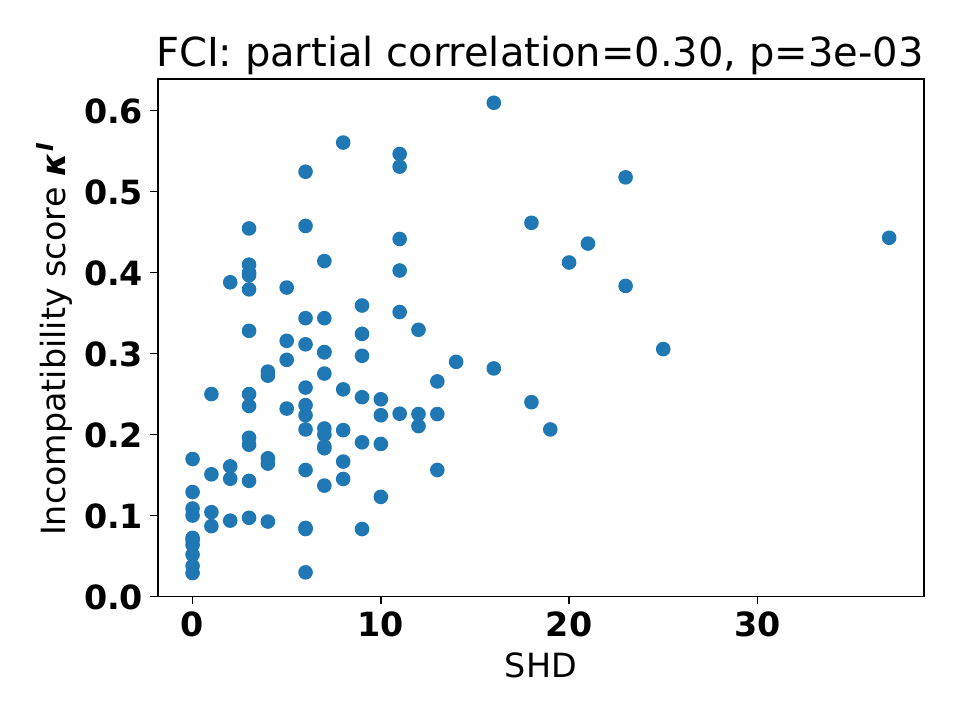}
		\caption{(Left) A histogram of $\kappa^I$ values for FCI on 100 linear Gaussian datasets. (Right) The structural Hamming distance of estimated graphs $\hat G$ to the respective true graph $G$ is on the $x$-axis and on the $y$-axis the incompatibility score $\kappa^I$. The figure shows a significant correlation between $\kappa^I$ and the SHD.}
		
		\label{fig:shd_vs_sb_fci}
	\end{figure}

	In \cref{sec:experiments} we have already seen that $\kappa^I$ is correlated with the SHD to the ground truth graph for RCD on linear non-Gaussian data.
    As a next step, we repeated this experiment with the FCI algorithm, where we used linear Gaussian data and the Fisher $Z$ test for conditional independence.
    In \cref{fig:shd_vs_sb_fci} we can see that this also yields a significant partial correlation (given the average node degree of the ground truth graph), albeit not as strong as for RCD.
    Recall, that we considered the partial correlation given the average node degree of the ground truth graph, as we suspect the density of the ground truth graph to affect both, the SHD and the incompatibility score.

  	\begin{figure}
		\centering
		\includegraphics[width=\textwidth]{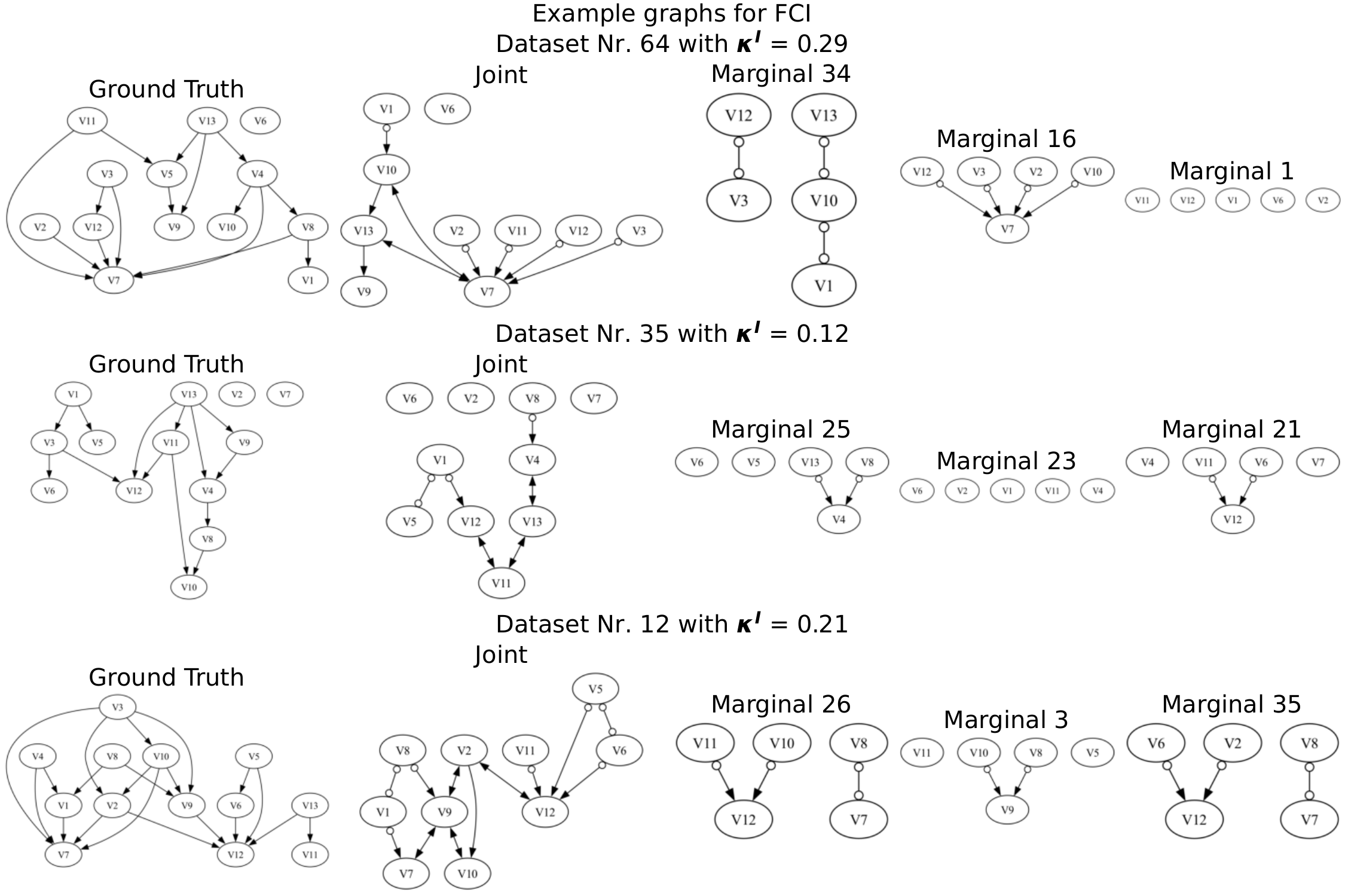}
		\caption{Randomly drawn example graphs from the experiment shown in \cref{fig:shd_vs_sb_fci}. The figure shows for three randomly picked datasets the ground truth graph (including hidden variables), the joint graph found by the algorithm on all variables and some randomly drawn marginal graphs, i.e. graphs that the algorithm found on subsets of variables.}
		
		\label{fig:example_graphs_fci}
	\end{figure}
\begin{figure}
		\centering
		\includegraphics[width=.4\textwidth]{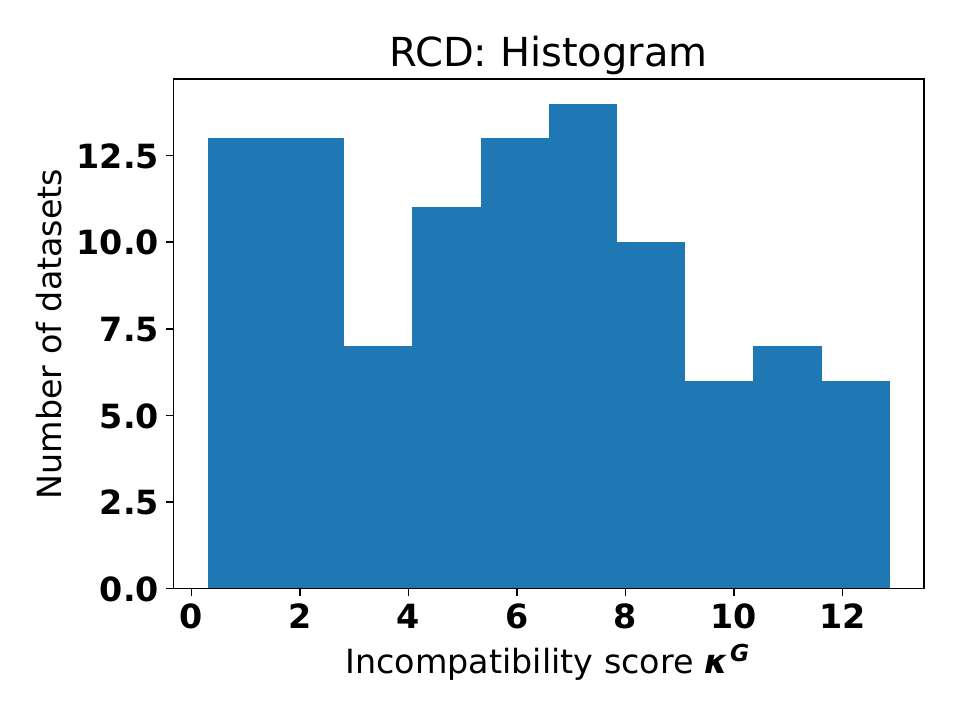}
		\includegraphics[width=.4\textwidth]{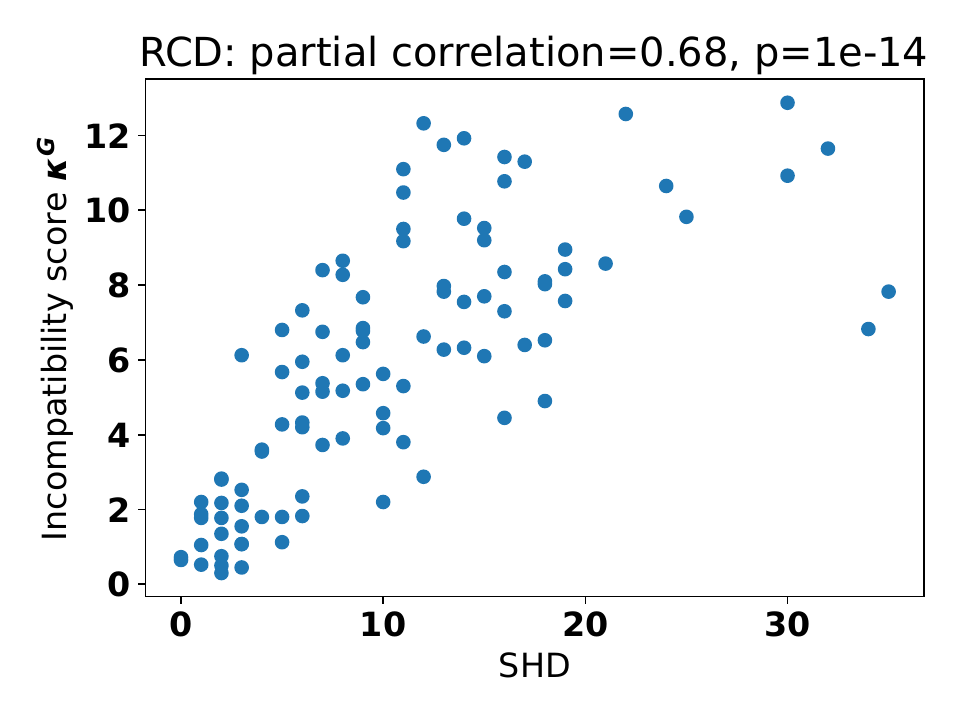}
		\caption{(Left) A histogram of $\kappa^G$ values for RCD on 100 datasets with linear models and uniform noise. (Right) The structural Hamming distance of estimated graphs $\hat G$ to the respective true graph $G$ is on the $x$-axis and on the $y$-axis the incompatibility score $\kappa^G$. The figure shows a significant correlation between $\kappa^G$ and the SHD.}
		
		\label{fig:shd_vs_sb_rcd_kappa_g}
	\end{figure}
\begin{figure}
		\centering
		\includegraphics[width=.4\textwidth]{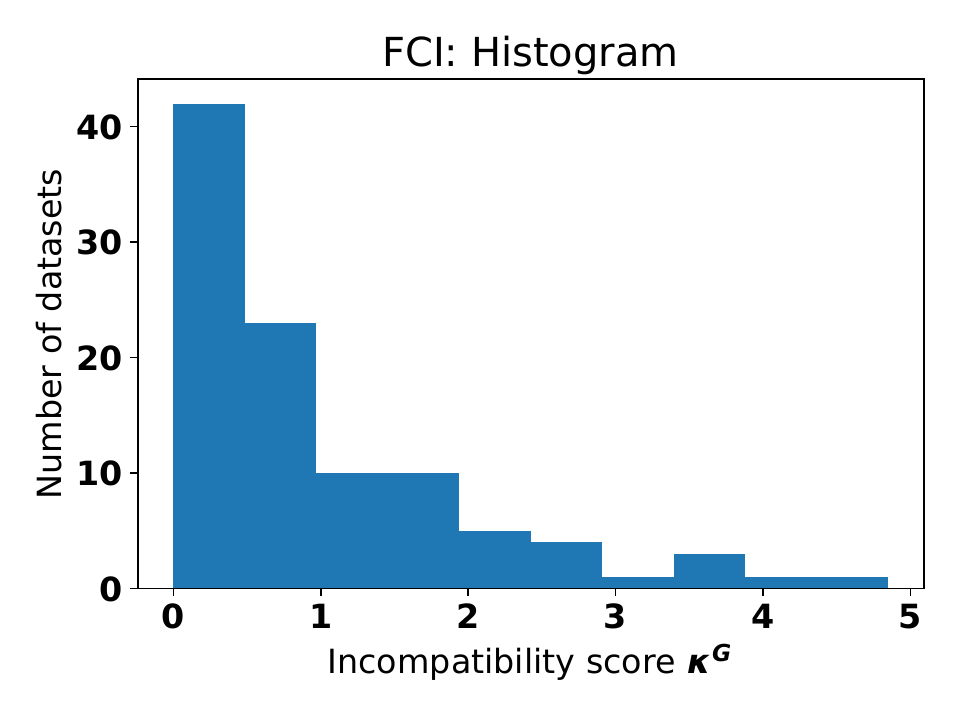}
		\includegraphics[width=.4\textwidth]{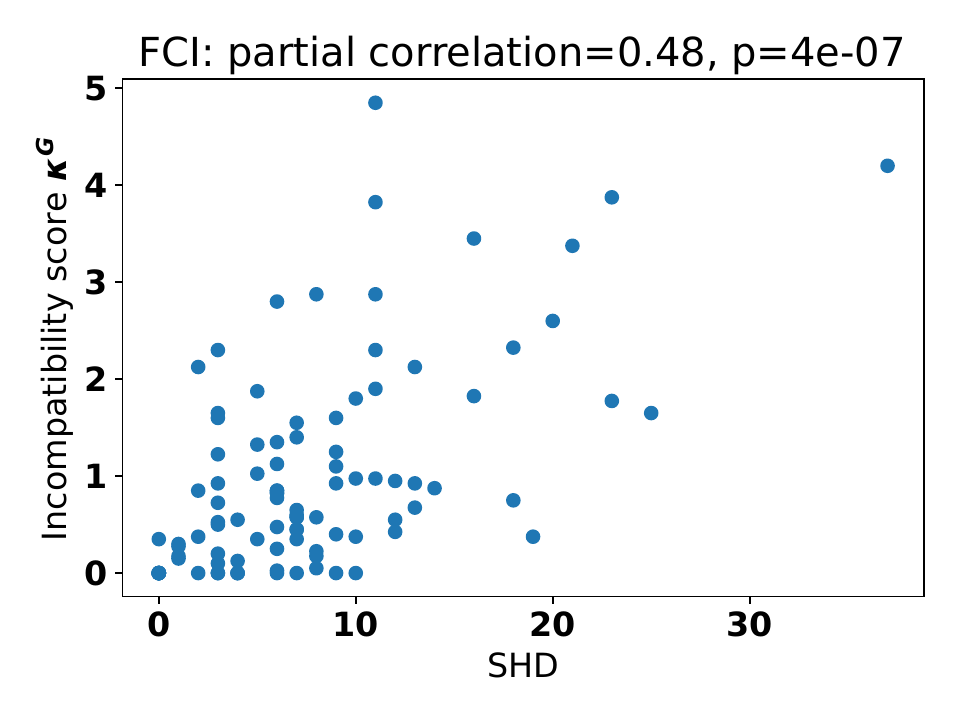}
		\caption{(Left) A histogram of $\kappa^G$ values for FCI on 100 datasets with linear models and Gaussian noise. (Right) The structural Hamming distance of estimated graphs $\hat G$ to the respective true graph $G$ is on the $x$-axis and on the $y$-axis the incompatibility score $\kappa^G$. The figure shows a significant correlation between $\kappa^G$ and the SHD.}
		
		\label{fig:shd_vs_sb_fci_kappa_g}
	\end{figure}

We also repeated the experiments with the graphical score $\kappa^G$.
\Cref{fig:shd_vs_sb_rcd_kappa_g,fig:shd_vs_sb_fci_kappa_g} show that we also get a significant correlation for the graphical criterion.

\begin{figure}
		\centering
		\includegraphics[width=.4\textwidth]{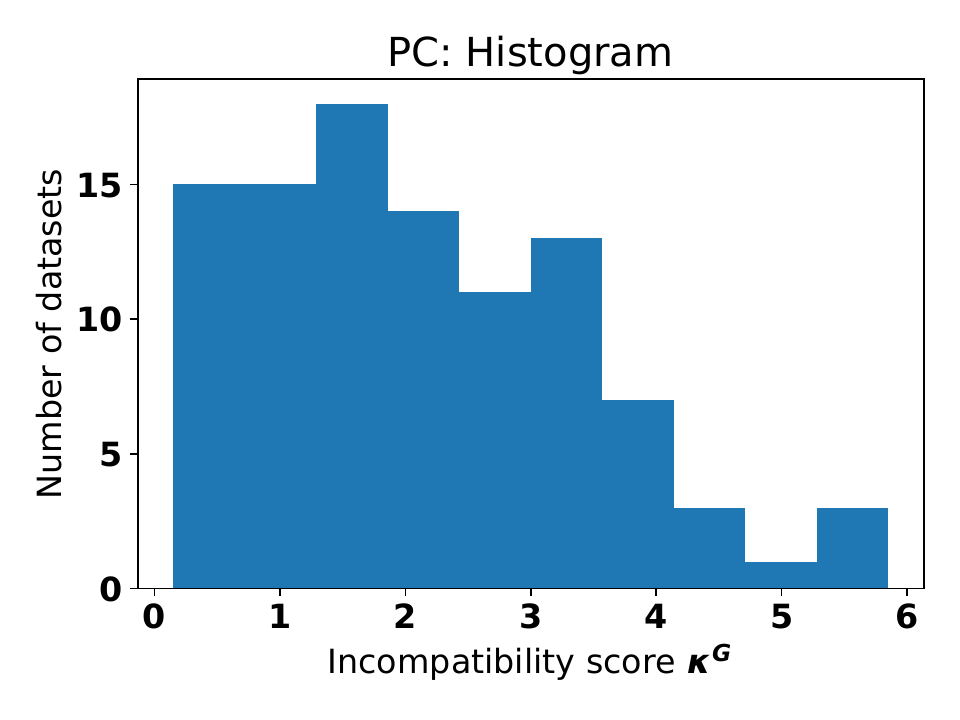}
		\includegraphics[width=.4\textwidth]{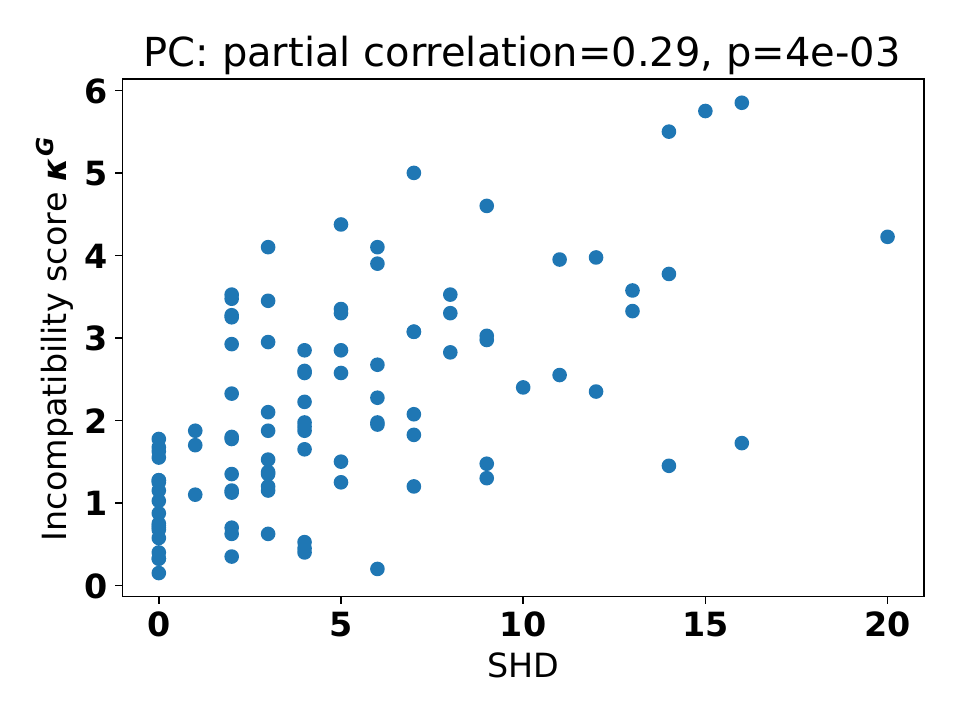}
		\caption{The structural Hamming distance of estimated graphs $\hat G$ to the respective true graph $G$ is on the $x$-axis and on the $y$-axis the incompatibility score $\kappa^G$ for PC.} 
        
		\label{fig:shd_vs_sb}
	\end{figure}

\begin{figure}
		\centering
		\includegraphics[width=.4\textwidth]{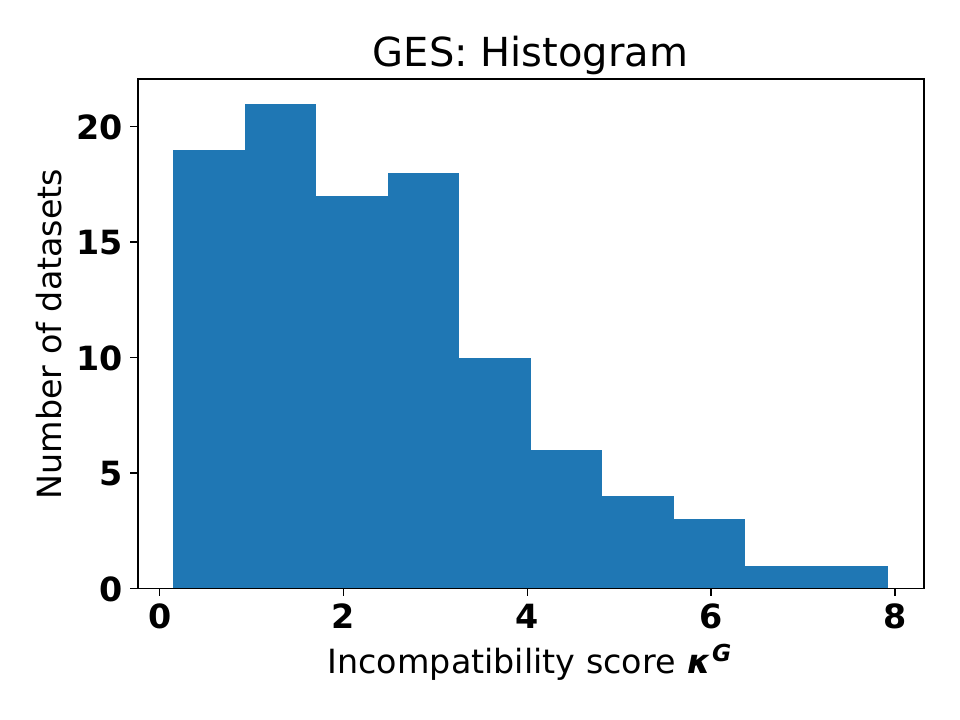}
		\includegraphics[width=.4\textwidth]{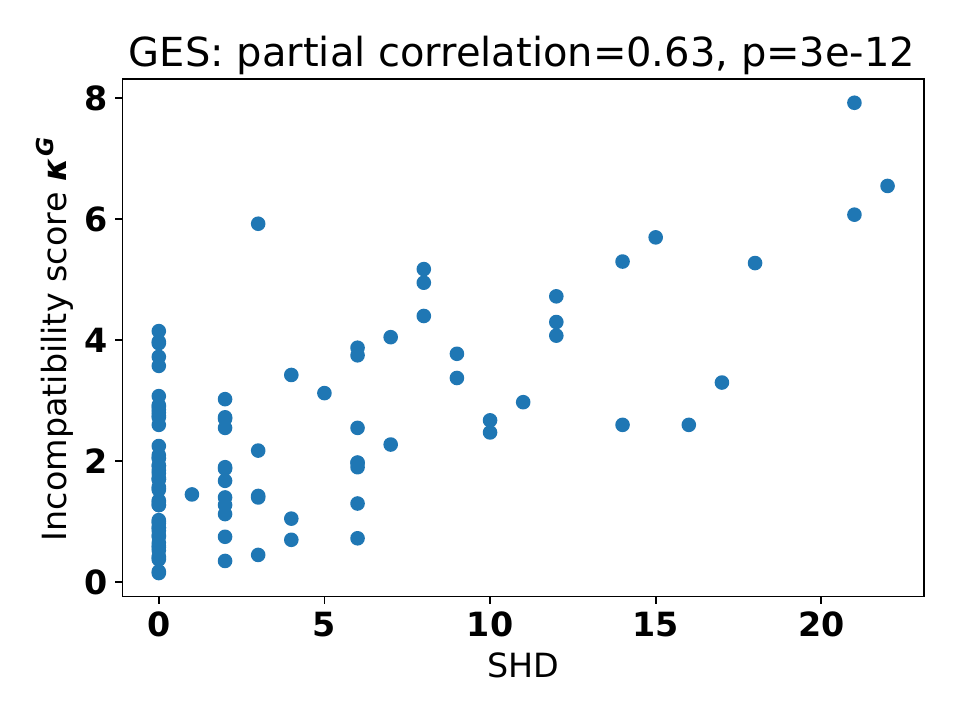}
		\caption{(Left) A histogram of $\kappa^G$ values for GES on 100 datasets with linear models and Gaussian noise. (Right) The structural Hamming distance of estimated graphs $\hat G$ to the respective true graph $G$ is on the $x$-axis and on the $y$-axis the incompatibility score $\kappa^G$.} 
        
		\label{fig:shd_vs_sb_ges}
	\end{figure}

\begin{figure}
		\centering
		\includegraphics[width=.4\textwidth]{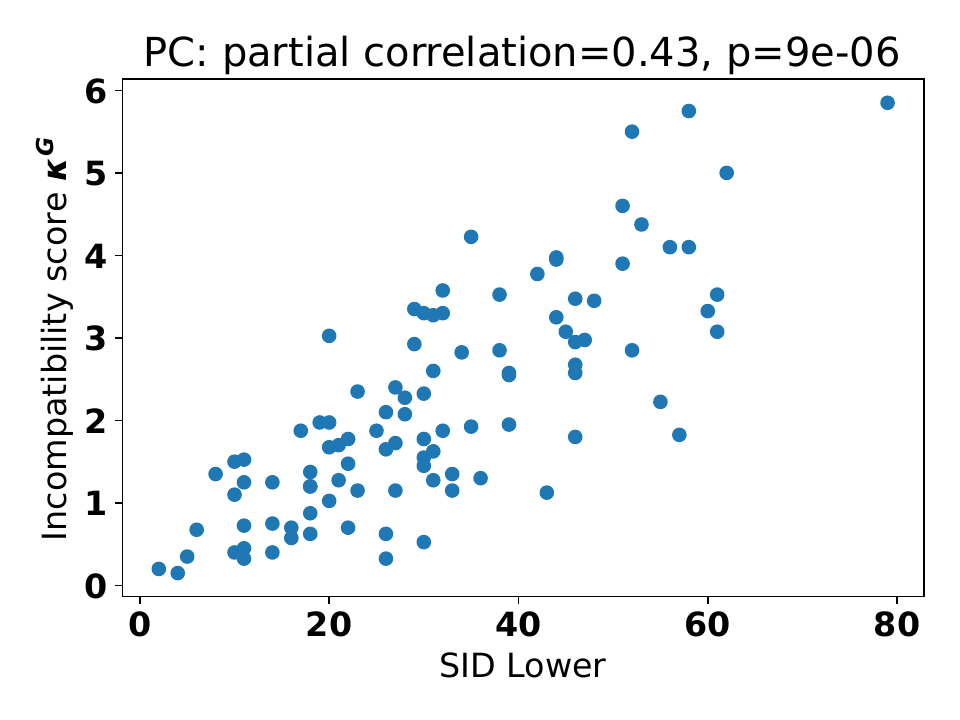}
		\includegraphics[width=.4\textwidth]{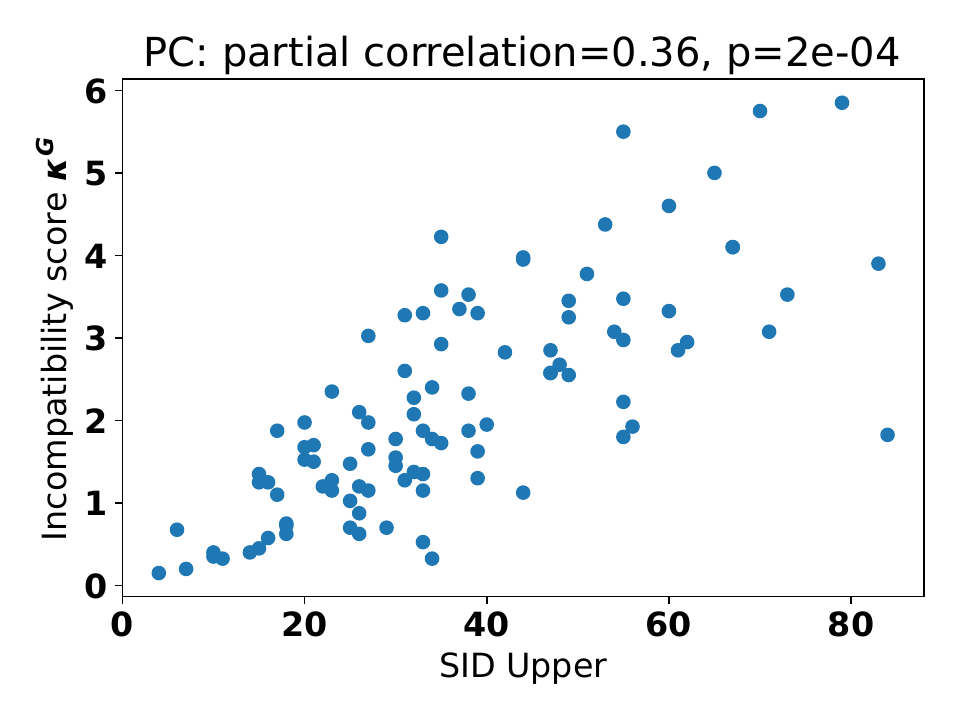}
		\caption{The bounds on the structural interventional  distance of estimated CPDAGs $\hat G$ to the respective true graph $G$ is on the $x$-axis and on the $y$-axis the incompatibility score $\kappa^G$ for PC. The plot shows a significant correlation between SID and $\kappa^G$.}
        
		\label{fig:shd_vs_sb_sid_pc}
	\end{figure}

\begin{figure}
		\centering
		\includegraphics[width=.4\textwidth]{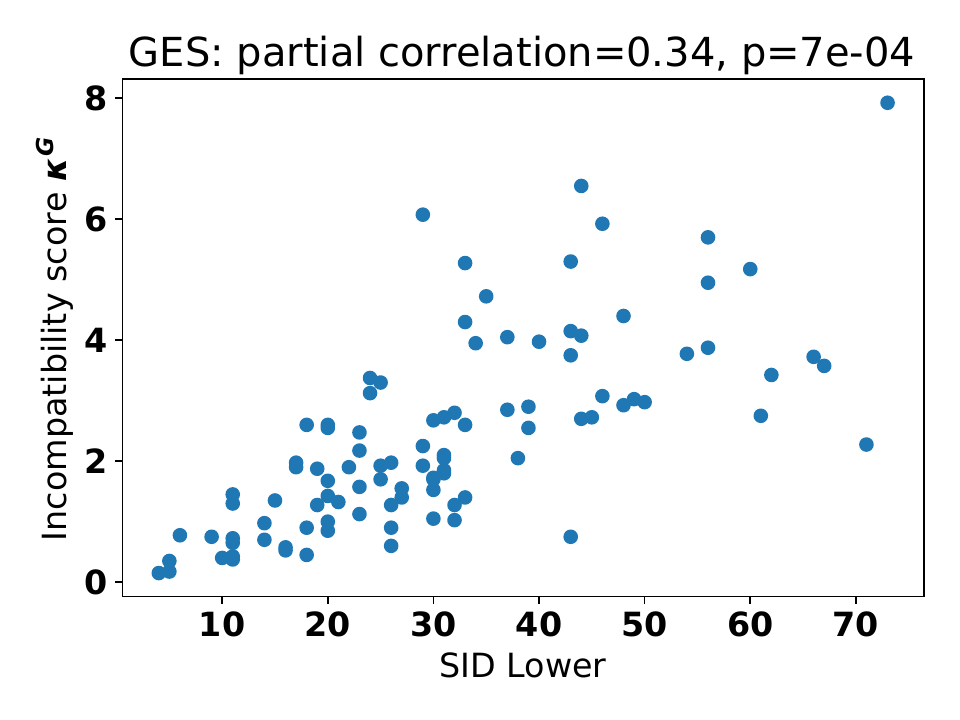}
		\includegraphics[width=.4\textwidth]{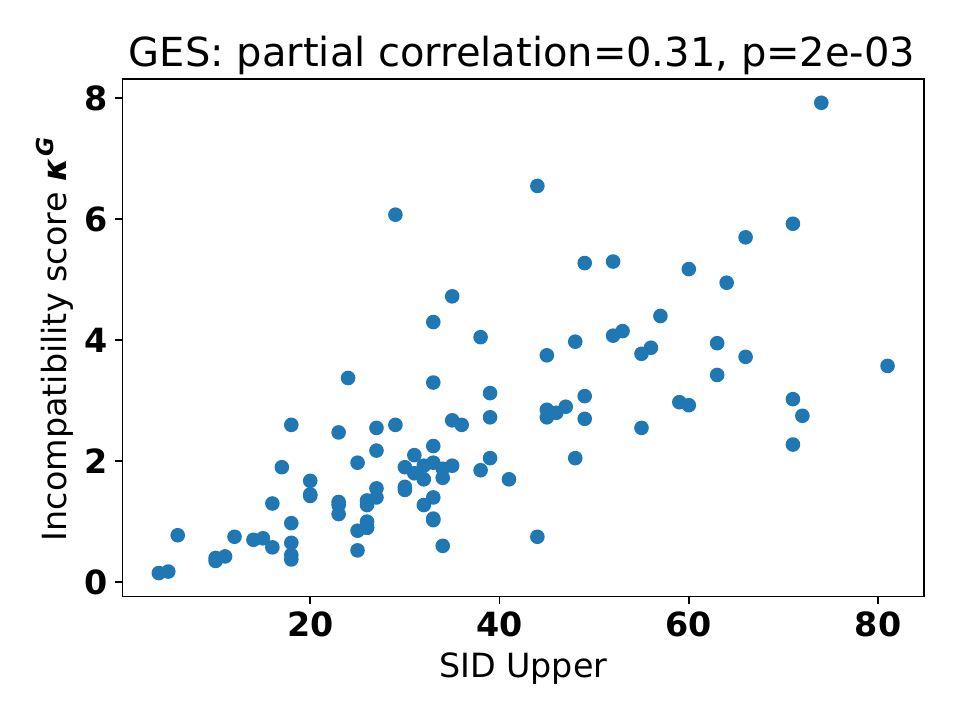}
		\caption{The bounds on the structural interventional  distance of estimated CPDAGs $\hat G$ to the respective true graph $G$ is on the $x$-axis and on the $y$-axis the incompatibility score $\kappa^G$ for GES. The plot shows a significant correlation between SHD and $\kappa^G$.}
        
		\label{fig:shd_vs_sb_sid_ges}
	\end{figure}

\begin{figure}
		\centering
		\includegraphics[width=.4\textwidth]{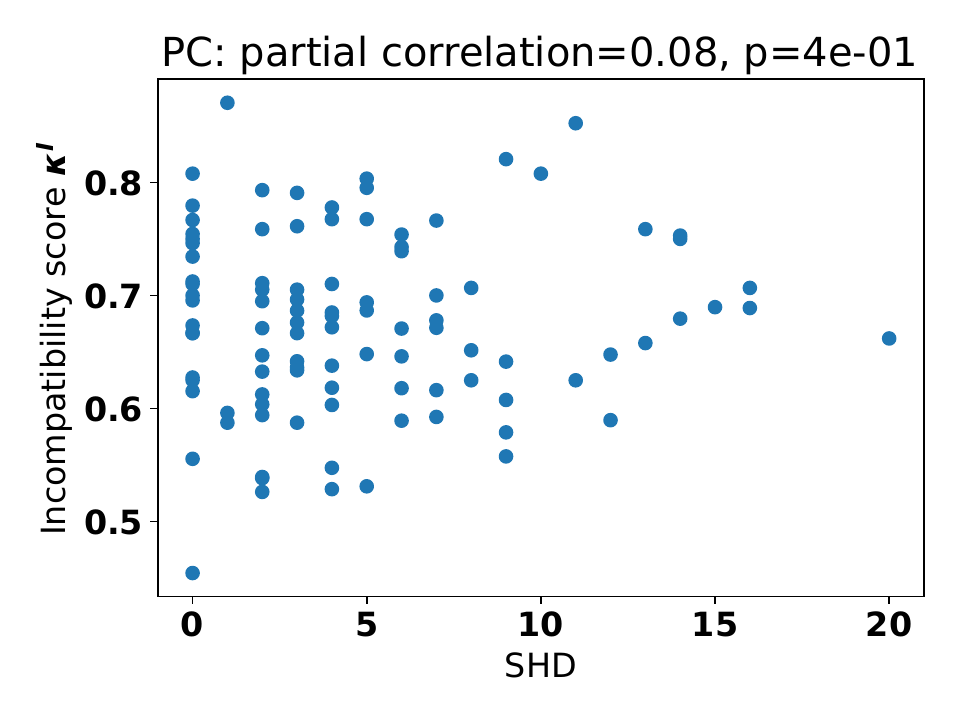}
		\includegraphics[width=.4\textwidth]{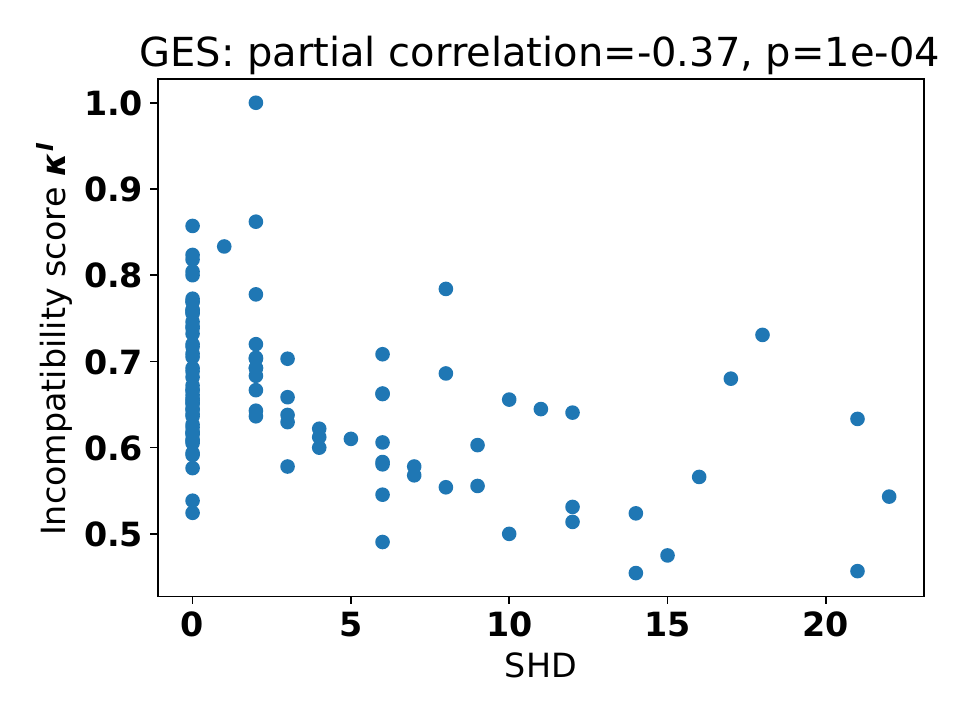}
		\caption{The structural Hamming distance of estimated graphs $\hat G$ to the respective true graph $G$ is on the $x$-axis and on the $y$-axis the incompatibility score $\kappa^I$ for PC and GES. In contrast to e.g. \cref{fig:rcd,fig:shd_vs_sb_fci} we do not see a positive correlation.} 
        
		\label{fig:shd_vs_sb_interv}
	\end{figure}

	In the next line of experiments we wanted to see whether this correlation also occurs in the setting where algorithms assume causal sufficiency and with the graphical criterion $\kappa^G$.  
	We therefore generated 100 datasets (as described above) and estimated  graphs $\hat G$ with two different causal discovery methods,
	namely PC and GES. 
	In \cref{fig:shd_vs_sb,fig:shd_vs_sb_ges} we see again that this yields a significant partial correlation. 
 In \cref{fig:shd_vs_sb_sid_pc,fig:shd_vs_sb_sid_ges} we can see that also the correlation between $\kappa^G$ and the bounds in the structural Interventional distance (SID) \citep{peters2015structural} is significant. 
 (Recall that SID is only defined for DAGs and \citet{peters2015structural} proposed to calculate the bounds on the SID over all DAGs in the equivalence class described by a CPDAG).
 Yet, \cref{fig:shd_vs_sb_interv} shows no correlation between SHD and $\kappa^I$ (in contrast to our experiments with RCD and FCI in \cref{fig:rcd,fig:shd_vs_sb_fci}).
    This seems to suggest that $\kappa^I$ might not be suitable to be used with algorithms that assume causal sufficiency without further modifications.


   	\begin{figure}
		\centering
		\includegraphics[width=\textwidth]{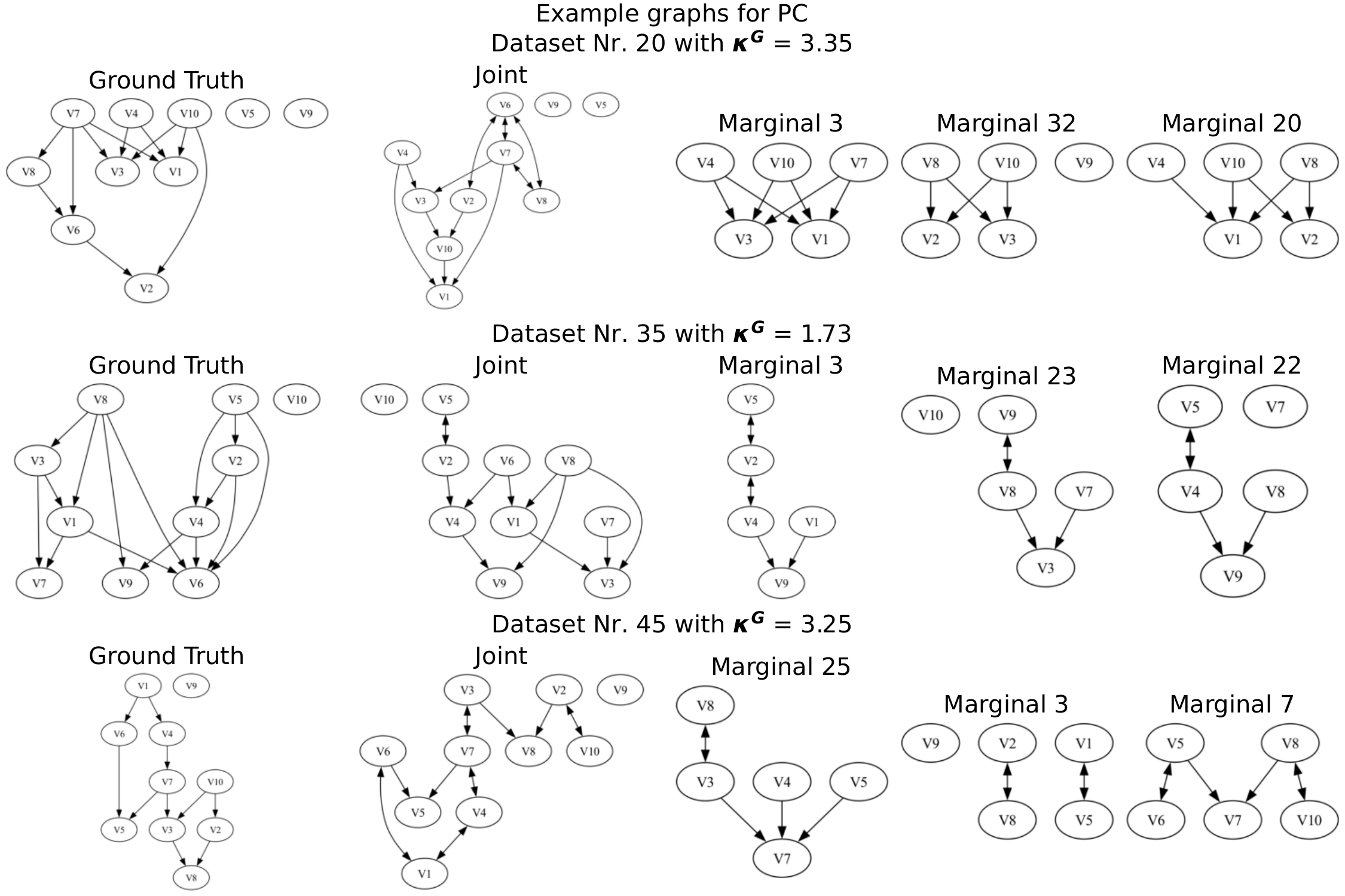}
		\caption{Randomly drawn example graphs from the experiment shown in \cref{fig:shd_vs_sb}. The figure shows for three randomly picked datasets the ground truth graph, the joint graph found by the algorithm on all variables and some randomly drawn marginal graphs, i.e. graphs that the algorithm found on subsets of variables.}
		
		\label{fig:example_graphs_pc}
	\end{figure}

   	\begin{figure}
		\centering
		\includegraphics[width=\textwidth]{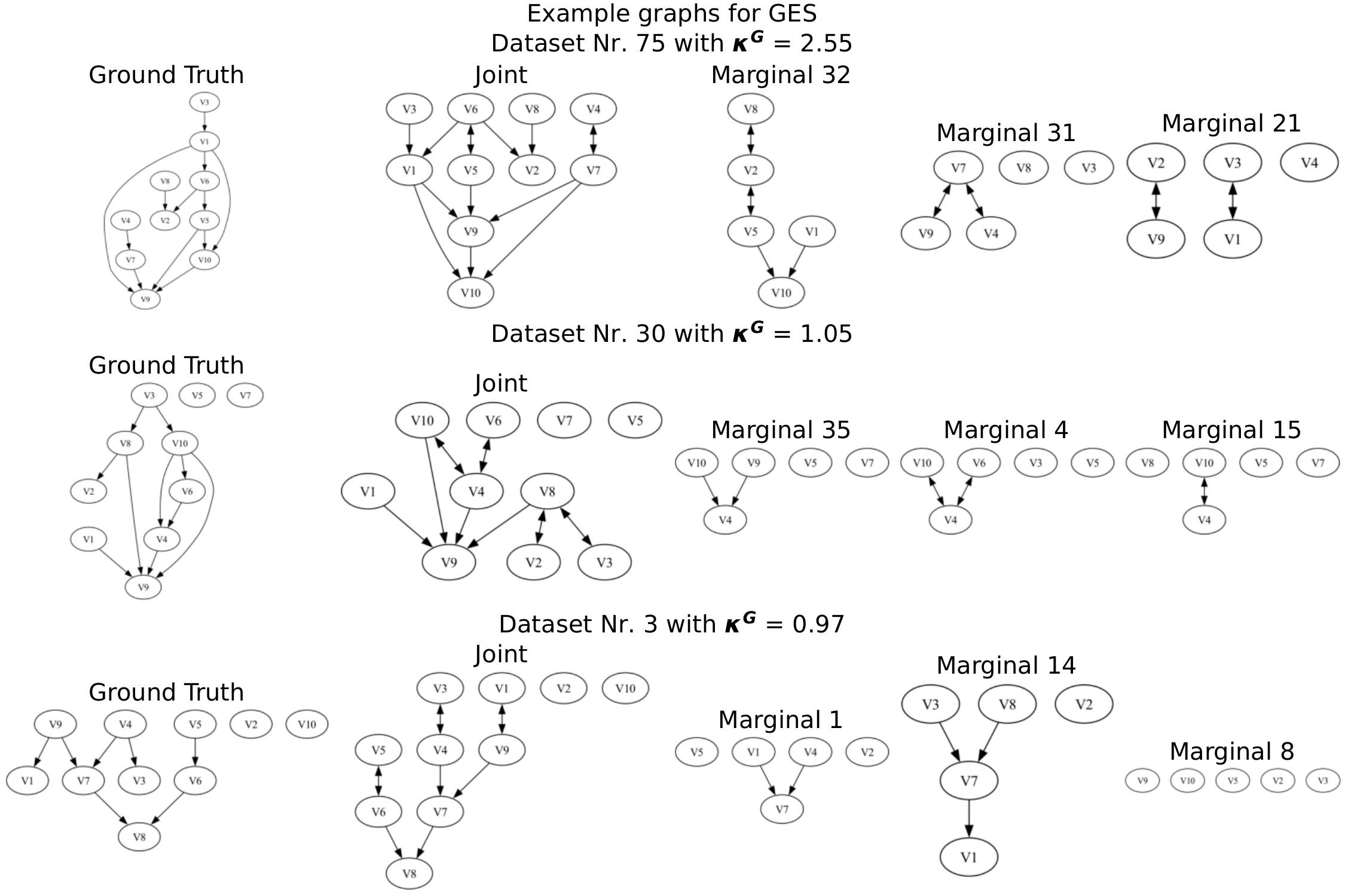}
		\caption{Randomly drawn example graphs from the experiment shown in \cref{fig:shd_vs_sb}. The figure shows for three randomly picked datasets the ground truth graph, the joint graph found by the algorithm on all variables and some randomly drawn marginal graphs, i.e. graphs that the algorithm found on subsets of variables.}
		
		\label{fig:example_graphs_ges}
	\end{figure}
	
	\subsection{Model Selection}
\begin{figure}
		\centering
		\includegraphics[width=.4\textwidth]{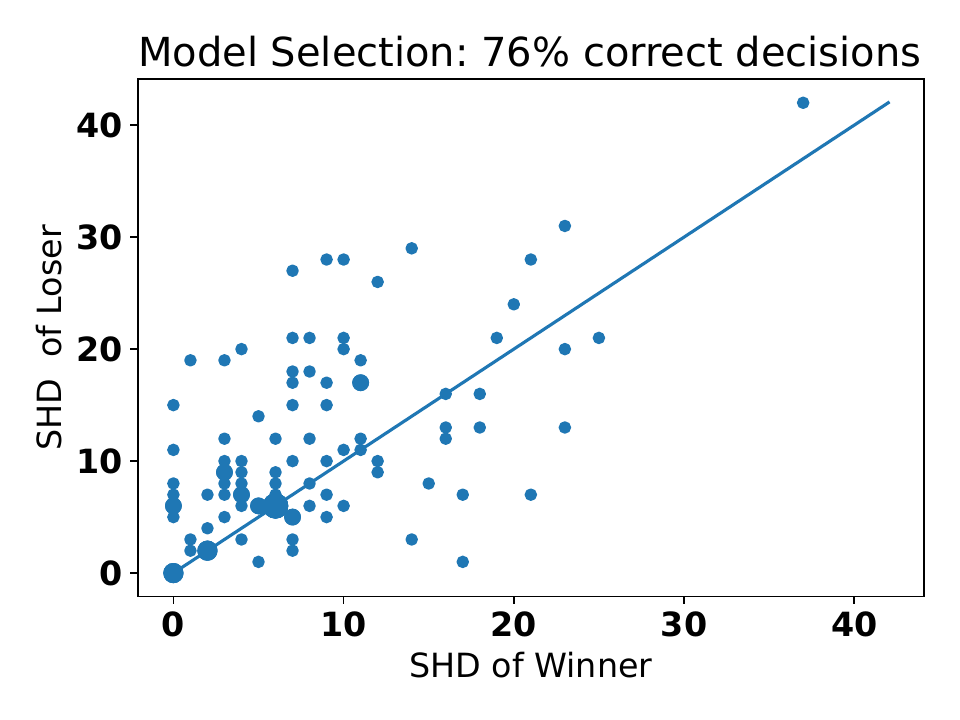}
		\includegraphics[width=.4\textwidth]{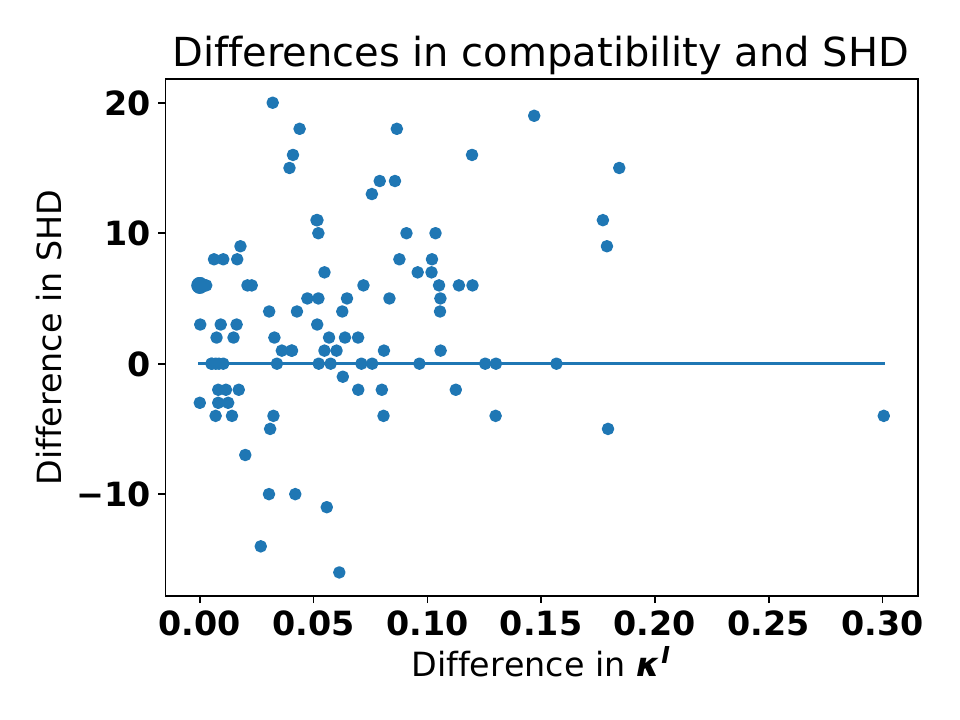}
		\caption{The incompatibility score $\kappa^I$ as metric for model selection FCI with $\alpha=0.1$ and $\alpha=0.001$ analogously to the main paper. We picked the model with the better $\kappa^I$ score as winner and report the SHD of the winner and of the loser on $x$-axis and $y$-axis respectively. In 62\% of datasets we picked the strictly better model, while 24\% are strictly worse.}
		\label{fig:model_selection_fci}
	\end{figure}

        We repeated the experiment from \cref{fig:model_selection} with the FCI algorithm. The FCI algorithm has exactly one hyperparameter, namely the threshold $\alpha$ below which the p-values of the conditional independence tests lead to a rejection of the null hypothesis. Again, we chose between $\alpha=0.1$ and $\alpha=0.001$.
        \Cref{fig:model_selection_fci} shows that we picked the better (or equally good) parameter in 76 \% of the datasets. Precisely, 62\% are strictly better, while 24\% are strictly worse.

\begin{figure}
		\centering
		\includegraphics[width=.4\textwidth]{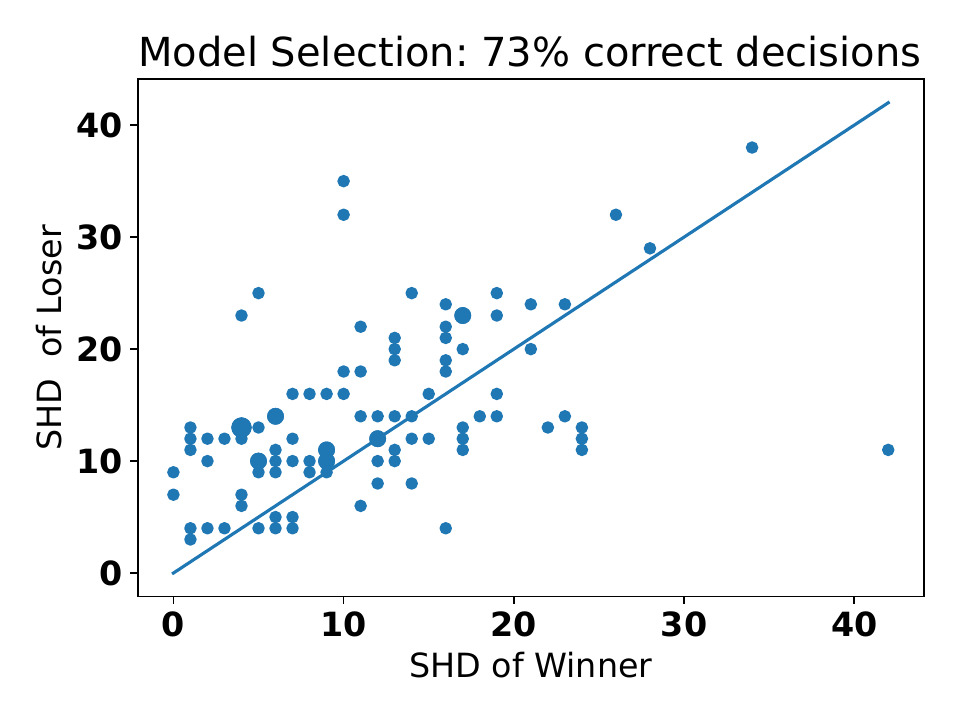}
		\includegraphics[width=.4\textwidth]{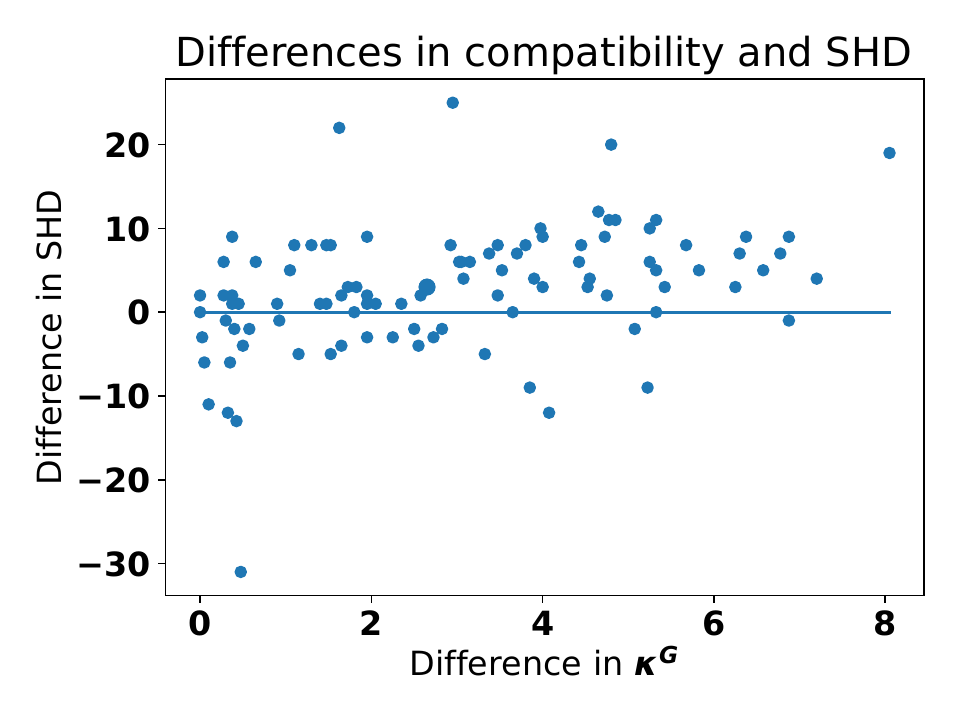}
		\caption{The incompatibility score $\kappa^G$ as metric for model selection RCD with $\alpha=0.1$ and $\alpha=0.001$ analogously to the main paper. We picked the model with the better $\kappa^G$ score as winner and report the SHD of the winner and of the loser on $x$-axis and $y$-axis respectively. In 69\% of datasets we picked the strictly better model, while 27\% are strictly worse.}
		\label{fig:model_selection_rcd_kappa_g}
	\end{figure}

    \Cref{fig:model_selection_rcd_kappa_g} shows the same experiment as in \cref{fig:model_selection} but this time with the graphical score $\kappa^G$.
    In this setting we pick the strictly better parameters in 69\% of the datasets and a strictly worse parameters in 27\%.

\begin{figure}
		\centering
		\includegraphics[width=.4\textwidth]{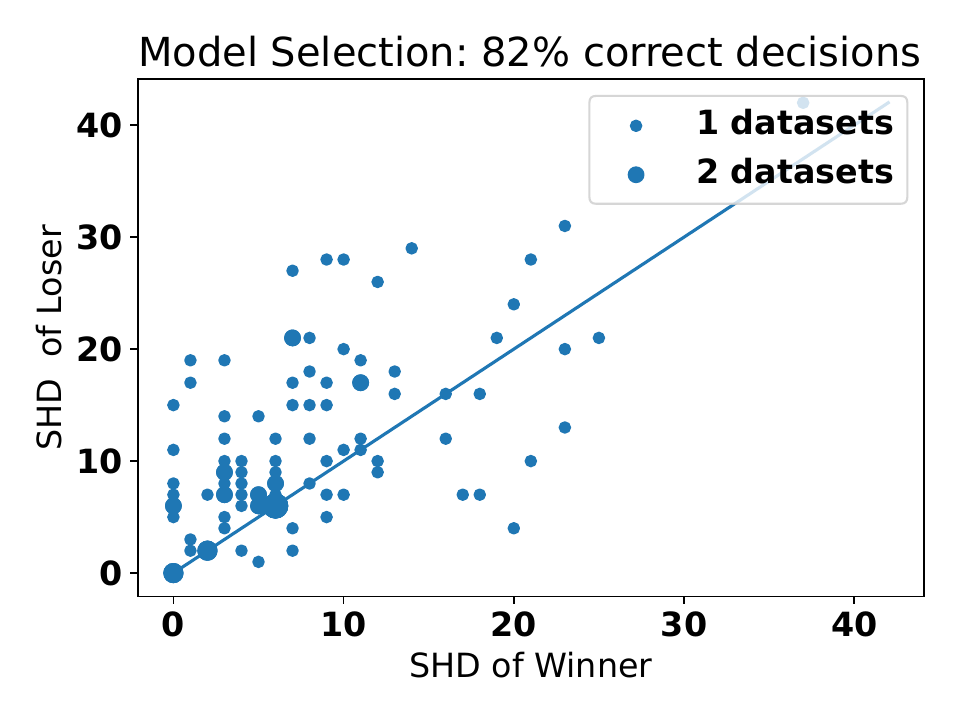}
		\includegraphics[width=.4\textwidth]{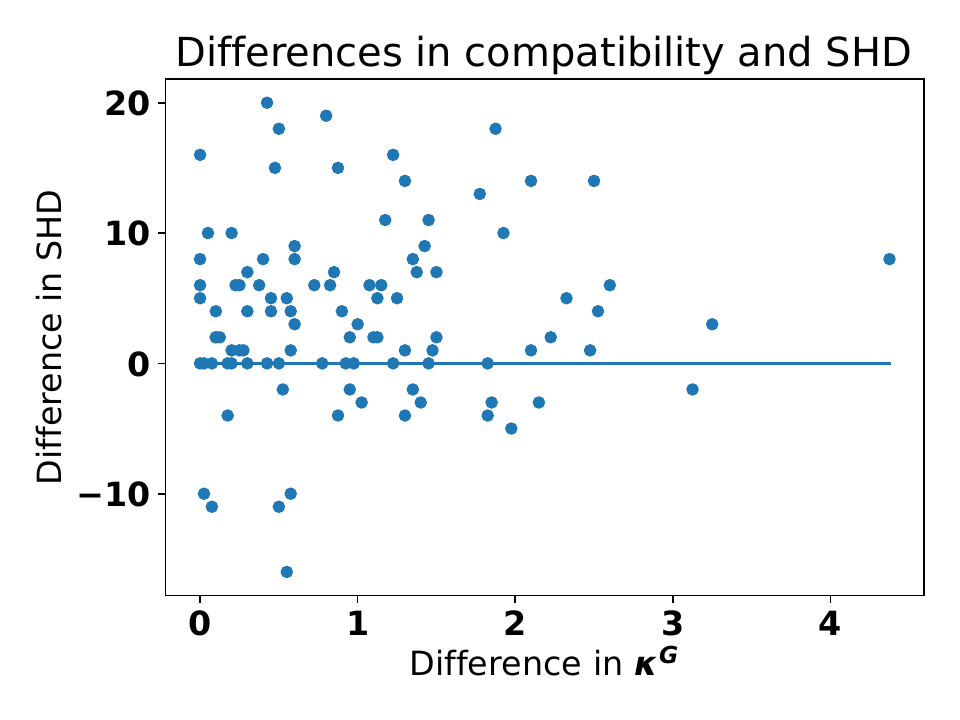}
		\caption{The incompatibility score $\kappa^G$ as metric for model selection FCI with $\alpha=0.1$ and $\alpha=0.001$ analogously to the main paper. We picked the model with the better $\kappa^G$ score as winner and report the SHD of the winner and of the loser on $x$-axis and $y$-axis respectively. In 68\% of datasets we picked the strictly better model, while 18\% are strictly worse. The dot size indicates that several points overlap.}
		\label{fig:model_selection_fci_kappa_g}
	\end{figure}

    In \cref{fig:model_selection_fci_kappa_g} we can see the same experiment with FCI and $\kappa^G$. 
    Here we got the strictly better parameter in 68\% of datasets and the strictly worse parameter in 18\%.
        
\begin{figure}
		\centering
		\includegraphics[width=.4\textwidth]{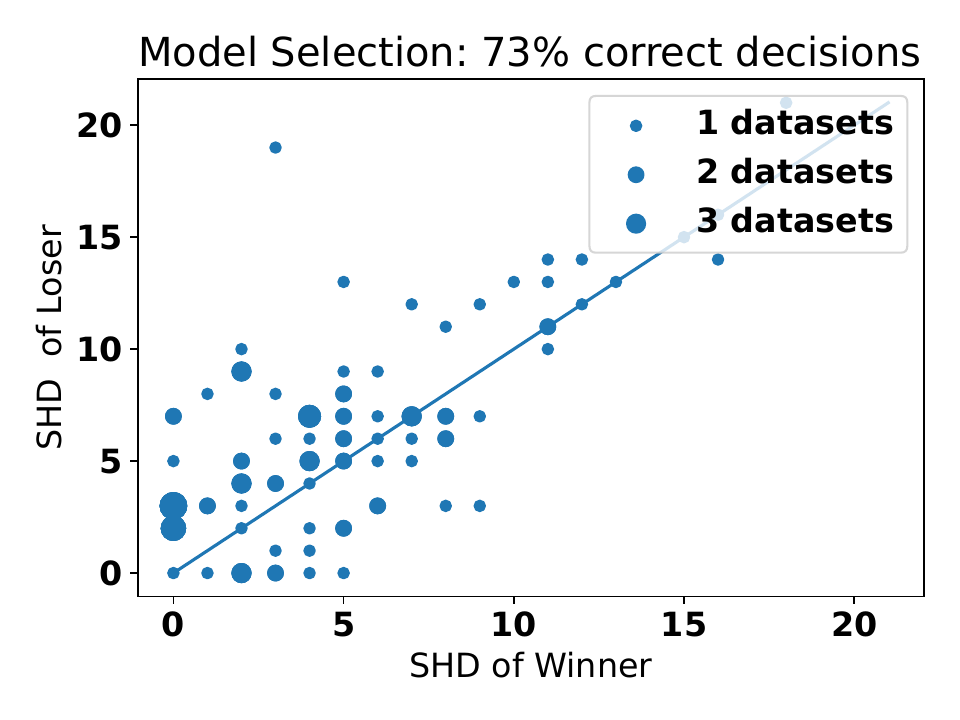}
		\includegraphics[width=.4\textwidth]{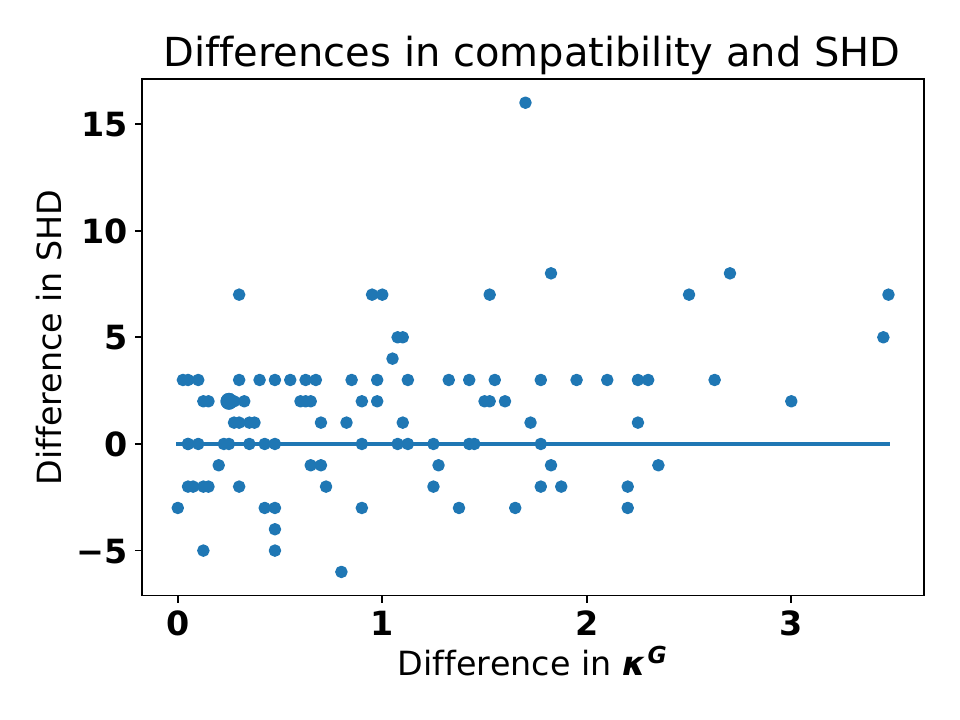}
		\caption{We used the incompatibility score $\kappa^G$ for model selection of the PC algorithm with  $\alpha=0.1$ and $\alpha=0.001$. We picked the strictly better parameter in 58\% of datasets and the strictly worse in 27\%. The dot size indicates that several points overlap.
		}
		\label{fig:model_selection_alpha}
	\end{figure}

	 Again, we also used PC and GES as algorithms that assume causal sufficiency.
	   Analogously to FCI, PC has only the $\alpha$-threshold as parameter.  
	We repeated an analogous experiment as above with the PC algorithm with  $\alpha=0.1$ and $\alpha=0.001$.
	The plots in \cref{fig:model_selection_alpha} suggest that again, the incompatibility score is an effective selection criterion and here we even make correct decisions 73\% of the cases.
 	\begin{figure}
		\centering
		\includegraphics[width=.4\textwidth]{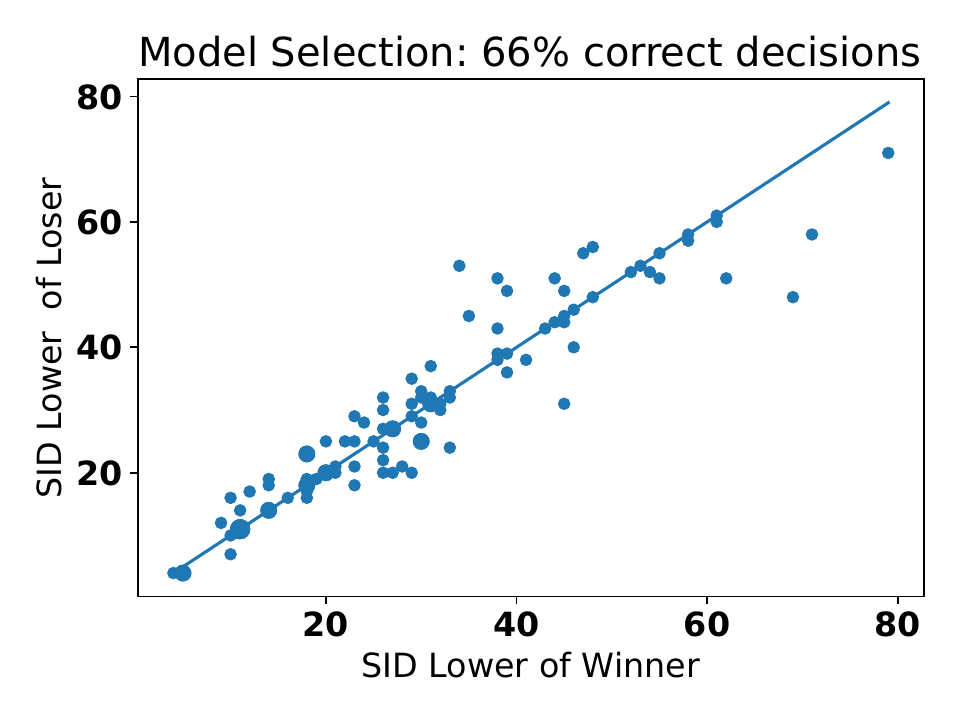}
		\includegraphics[width=.4\textwidth]{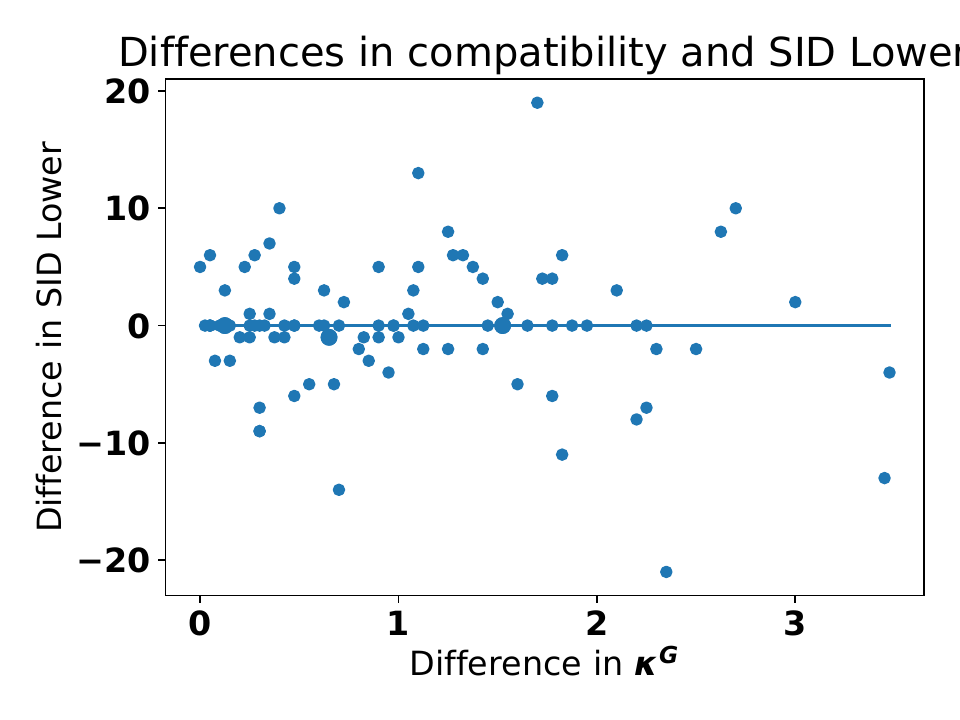}
		\caption{We used the incompatibility score $\kappa^G$ for model selection of the PC algorithm with  $\alpha=0.1$ and $\alpha=0.001$. We picked the strictly better parameter w.r.t. the lower bound on the SID in 33\% of datasets and the strictly worse in 34\%. 
		}
		\label{fig:model_selection_alpha_sid_lower}
	\end{figure}

  	\begin{figure}
		\centering
		\includegraphics[width=.4\textwidth]{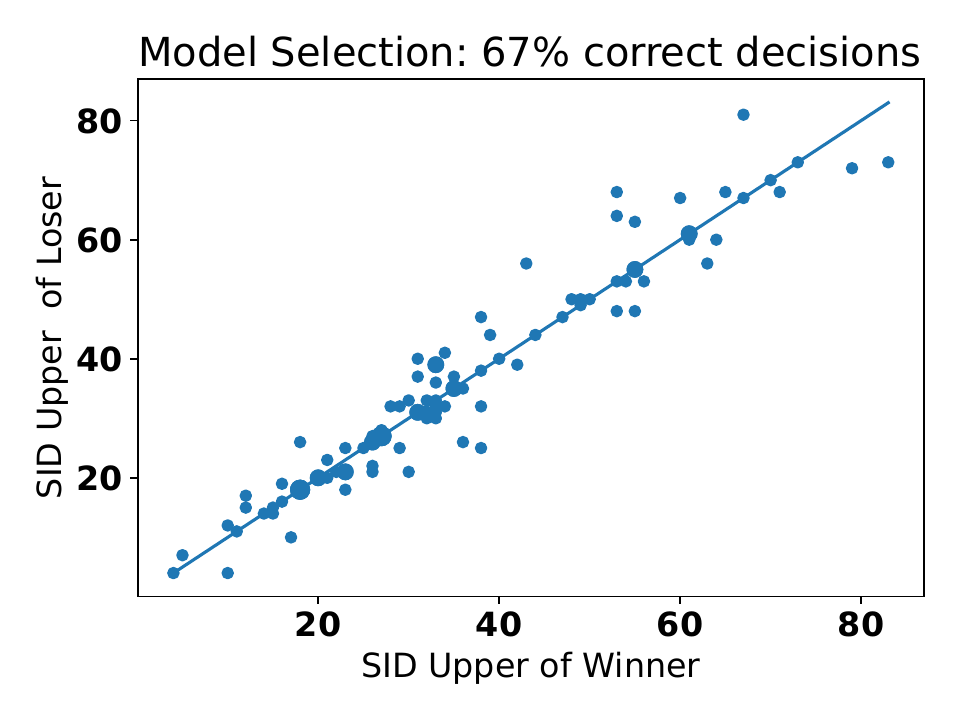}
		\includegraphics[width=.4\textwidth]{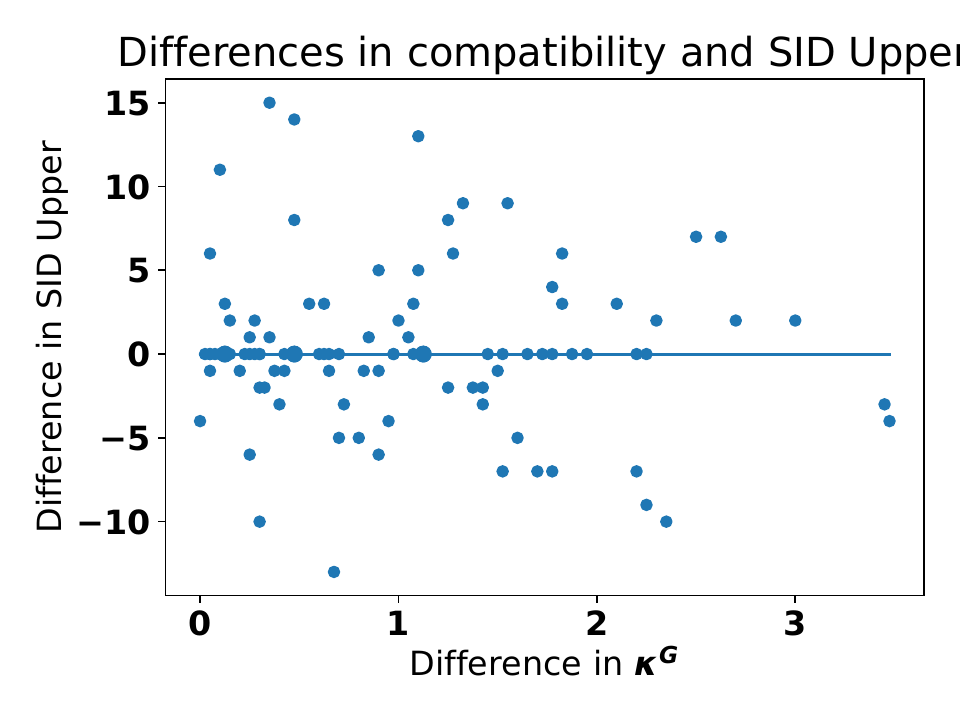}
		\caption{We used the incompatibility score $\kappa^G$ for model selection of the PC algorithm with  $\alpha=0.1$ and $\alpha=0.001$. We picked the strictly better parameter w.r.t. the upper bound on the SID in 32\% of datasets and the strictly worse in 33\%.
		}
		\label{fig:model_selection_alpha_sid_upper}
	\end{figure}

\begin{figure}
		\centering
		\includegraphics[width=.4\textwidth]{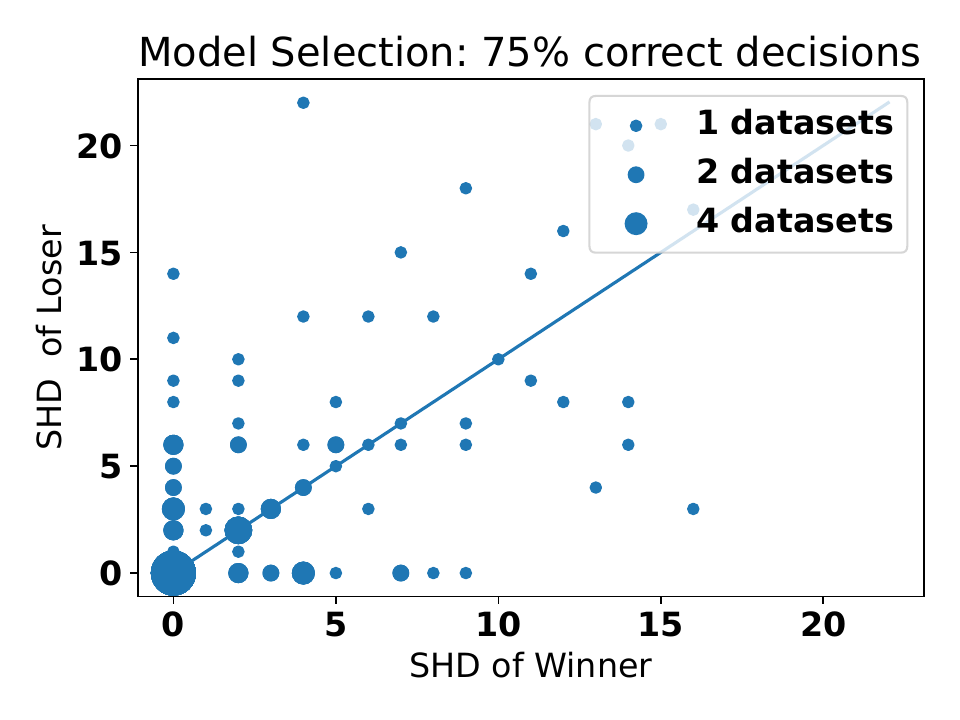}
		\includegraphics[width=.4\textwidth]{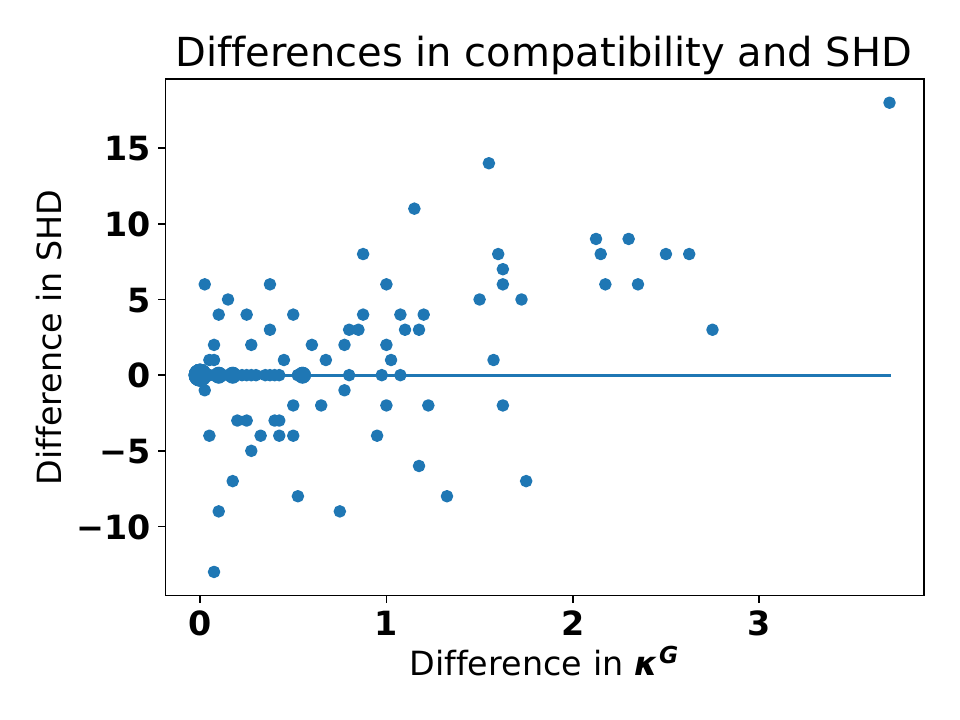}
		\caption{The incompatibility score $\kappa^G$ as metric for model selection between PC and GES analogously to the main paper. We picked the model with the better $\kappa^G$ score as winner and report the SHD of the winner and of the loser on $x$-axis and $y$-axis respectively. We picked the strictly better parameter w.r.t. SHD in 43\% of datasets and the strictly worse in 25\%. The dot size indicates that several points overlap.}
		\label{fig:model_selection_graphical}
	\end{figure}

\begin{figure}
		\centering
		\includegraphics[width=.4\textwidth]{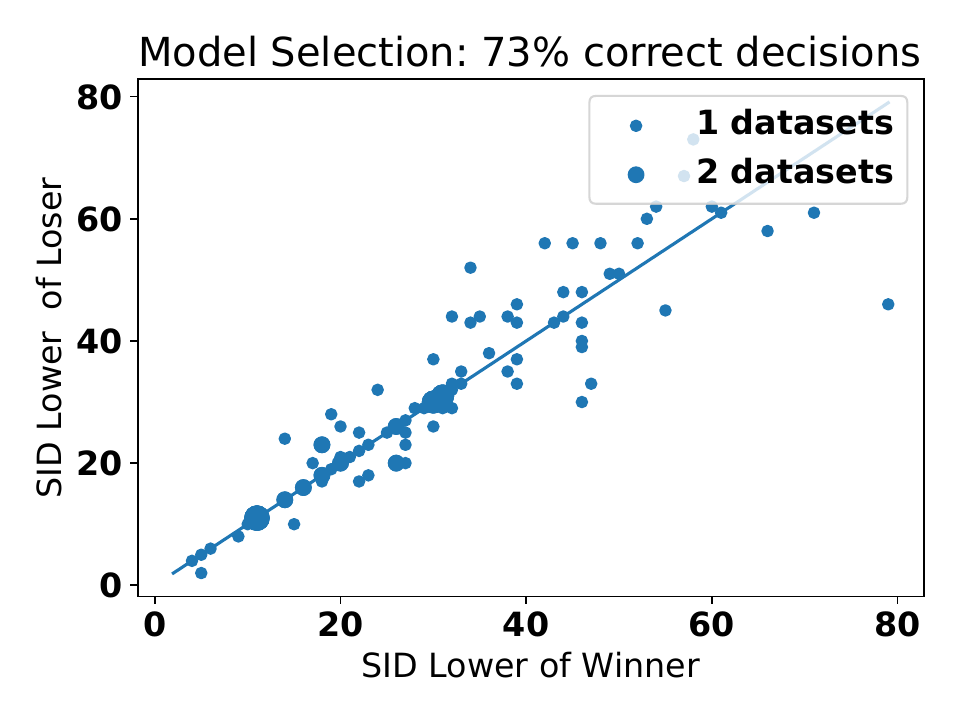}
		\includegraphics[width=.4\textwidth]{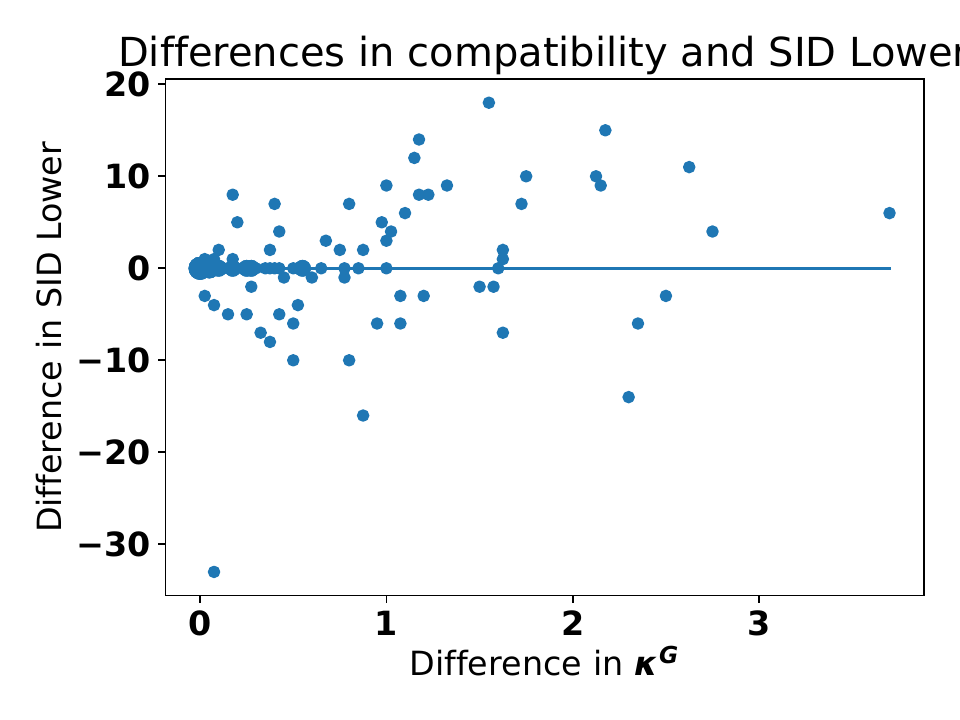}
		\caption{The incompatibility score $\kappa^G$ as metric for model selection between PC and GES analogously to the main paper. We picked the model with the better $\kappa^G$ score as winner and report the lower bound for the SID of the winner CPDAG and of the loser on $x$-axis and $y$-axis respectively. We picked the strictly better parameter w.r.t. the lower bound on the SID in 34\% of datasets and the strictly worse in 27\%. The dot size indicates that several points overlap.}
		\label{fig:model_selection_graphical_sid_lower}
	\end{figure}

\begin{figure}
		\centering
		\includegraphics[width=.4\textwidth]{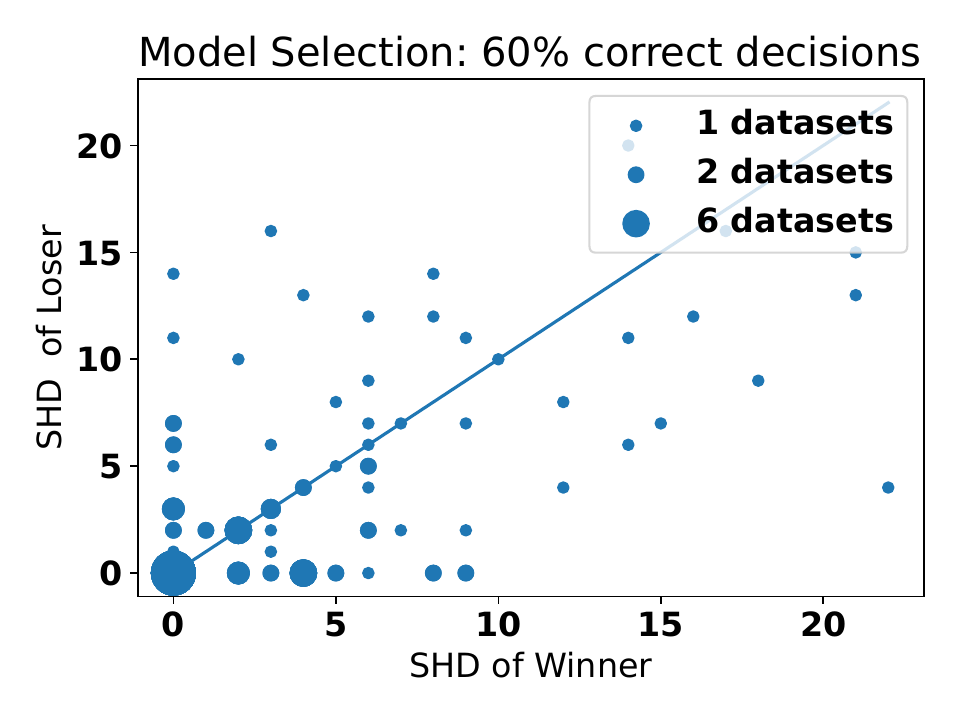}
		\includegraphics[width=.4\textwidth]{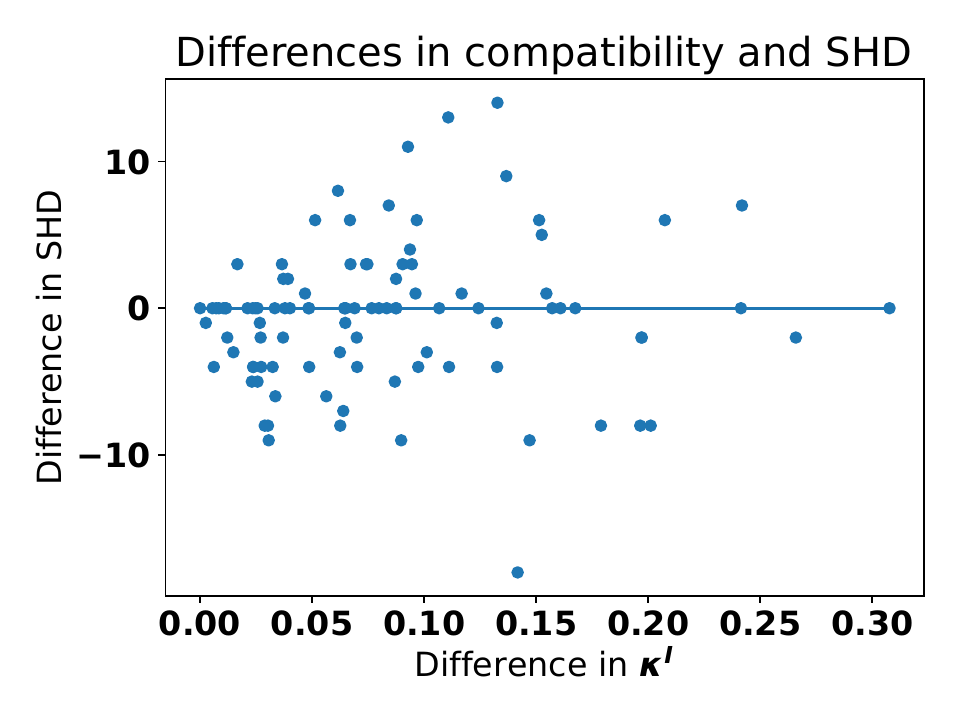}
		\caption{The incompatibility score $\kappa^I$ as metric for model selection between PC and GES analogously to the main paper. We picked the strictly better parameter w.r.t. to SHD in 28\% of datasets and the strictly worse in 40\%. The dot size indicates that several points overlap.}
		\label{fig:model_selection_interv}
	\end{figure}

    Eventually, we used the graphical score $\kappa^G$ to select between different \emph{algorithms}, in contrast to the previous experiments where we picked hyperparameters.
    Note, that such a comparison would not be possible between FCI and RCD, as the scores of PAGs and ADMGs are not a priori comparable.
    But PC and GES both output CPDAGs, which is why we chose them for these experiments.
	\Cref{fig:model_selection_graphical} does not show a similarly good performance as before, as we chose the strictly better or algorithm in 43\% of datasets, while we picked the worse algorithm in 25\%. 
    Analogously, \cref{fig:model_selection_graphical_sid_lower} shows that we picked the strictly better model w.r.t. the lower bound on the SID given by the CPDAG in 34\% of the cases and the strictly worse model in 27\% of the cases. 
    Although w.r.t. to the upper bound on the SID we picked a worse model in 31\% of the cases and the better model only in 28\% of the datasets.
    \Cref{fig:model_selection_interv} seems to suggest that $\kappa^I$ cannot be used for model selection with algorithms that assume causal sufficiency (without further modifications): 60\% of the points are on or above the line, but only 28\% are strictly above the line and 40\% are strictly below the line.
    This is in line with what we expected after the experiment shown in \cref{fig:shd_vs_sb_interv}.
 	\begin{figure}
		\centering
		\includegraphics[width=.4\textwidth]{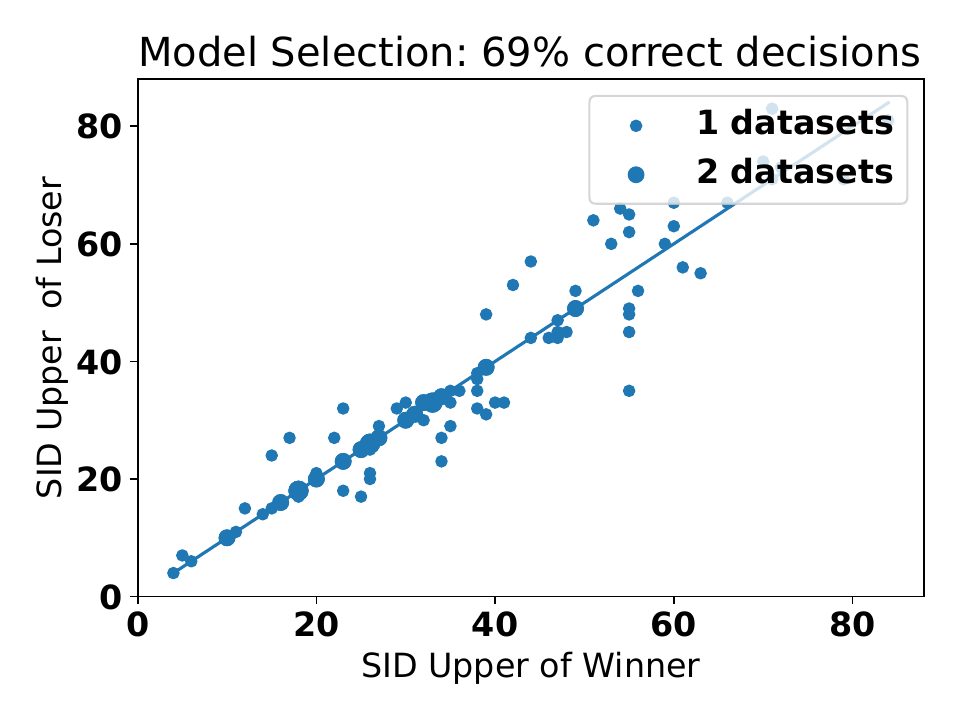}
		\includegraphics[width=.4\textwidth]{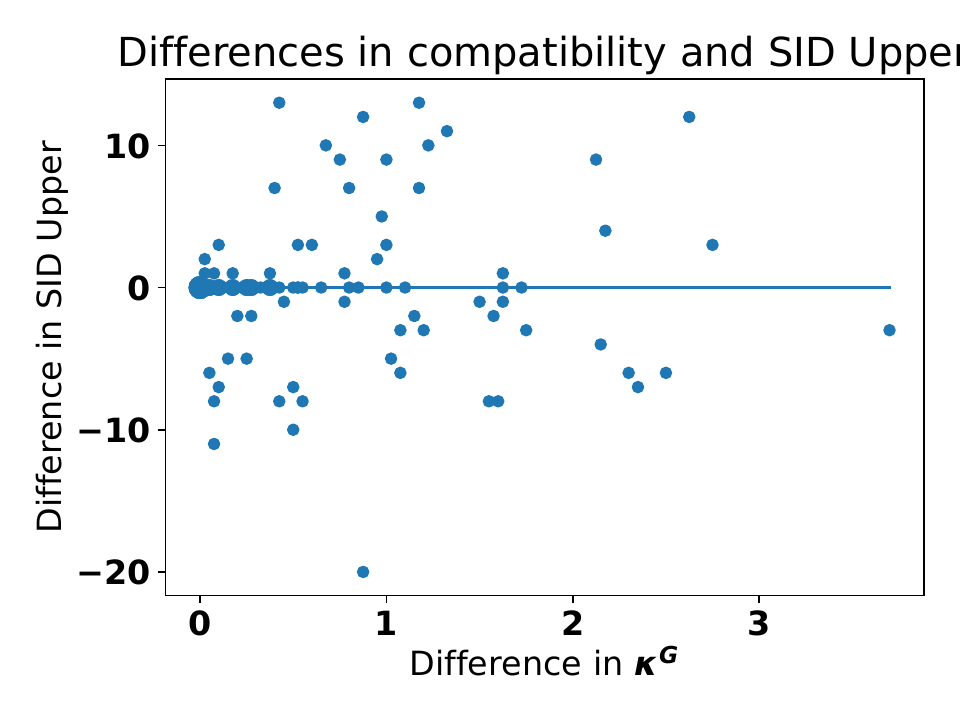}
		\caption{The incompatibility score $\kappa^G$ as metric for model selection between PC and GES analogously to the main paper. We picked the model with the better $\kappa^G$ score as winner and report the upper bound for the SID of the winner CPDAG and of the loser on $x$-axis and $y$-axis respectively. We picked the strictly better parameter w.r.t. the upper bound on the SID in 28\% of datasets and the strictly worse in 31\%. The dot size indicates that several points overlap.}
		\label{fig:model_selection_graphical_sid_upper}
	\end{figure}

	\subsection{Additional Real Data}
	We applied our incompatibility scores to the dataset presented by \citet{sachs2005causal}.
	
	We compared \dominik{be consistent regarding past tense or present tense} the SHD of $\cA(\bX)$ for all algorithms $\cA$ and also the $F_1$ score with respect to the existence of an edge in the skeleton of the resulting graph.
	It is worth noting that the SHD (and also the incompatibility scores) of different model types like ADMGs, PAGs and CPDAGs are not directly comparable and can only give an intuition about the relative performance of the algorithms.
    For the algorithms using the kernel independence test (KCI), we also randomly subsampled 1000 datapoints to speed up the computation.
    For all algorithms we picked $\alpha=0.01$.
	
\begin{figure}
		\centering
		\includegraphics[width=\textwidth]{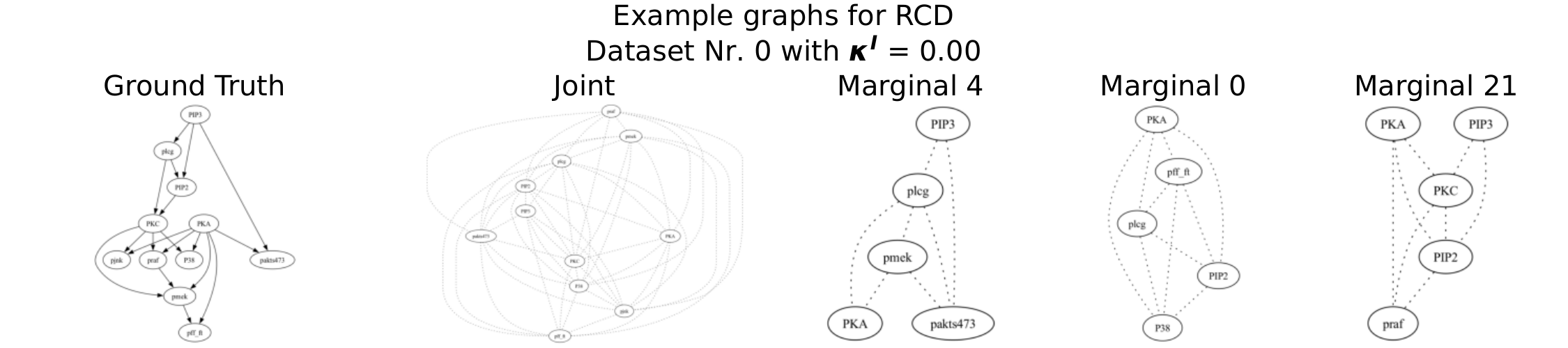}
  \includegraphics[width=\textwidth]{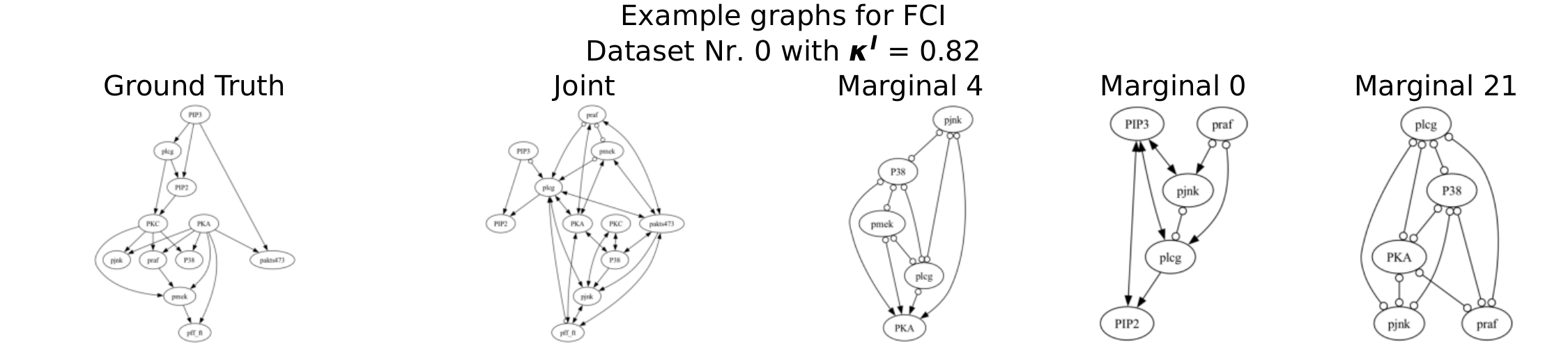}
  \includegraphics[width=\textwidth]{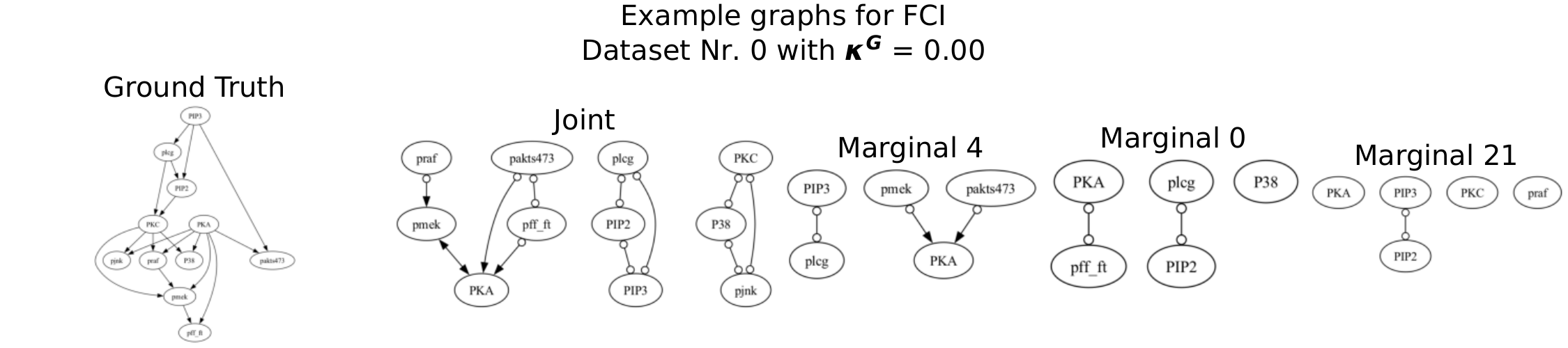}
  \includegraphics[width=\textwidth]{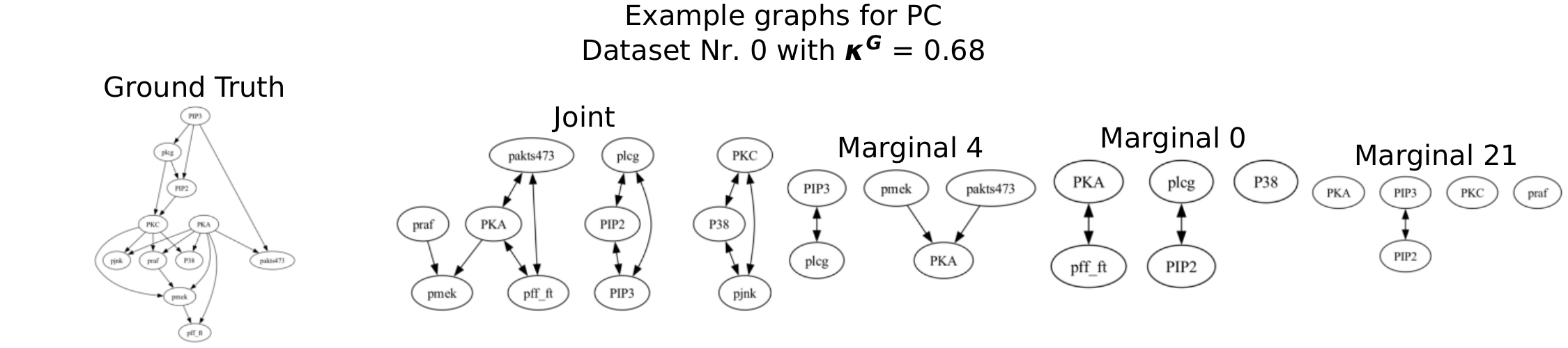}
		\caption{Randomly drawn example graphs from the experiment on the cell dataset. The figure shows the ground truth graph, the joint graph found by the algorithm on all variables and some randomly drawn marginal graphs, i.e. graphs that the algorithm found on subsets of variables. The algorithms are (from top to bottom) RCD, FCI with Fisher $Z$ test, FCI with kernel independence test and PC with kernel independence test.}
		
		\label{fig:example_graphs_sachs}
	\end{figure}

	 We suspected the real dataset might contain confounding.
    Therefore we started by using an algorithm that does not assume causal sufficiency.
  As the LiNGAM-based method does not merely return a Markov equivalence class, 
  we picked the RCD algorithm first.
  We used both incompatibility scores $\kappa^I$ and $\kappa^G$.
	As we can see in \cref{tab:results_sachs}, 
	the graphical incompatibility score is in a medium to high range, compared to the results in \cref{fig:shd_vs_sb_rcd_kappa_g} (i.e. compared to the setting where we know the true SHD).
	This lead us to the conclusion that possibly either linearity or the non-Gaussian additive noise assumption are violated.
    (Note, that the interventional score is 0, i.e. the best score possible. Looking at \cref{fig:example_graphs_sachs}, this is probably as in the joint graph, no interventional probability is identifiable. This shows that compatibility alone does not suffice as criterion, but one also needs to account for how much an algorithm \enquote{commits} to falsifiable statements.)
	We therefore tried FCI with the correlation-based Fisher $Z$ test and used again the interventional score $\kappa^I$ and the graphical score $\kappa^G$.
	Again, the graphical score is not really low and now additionally the interventional score seems to be quite high
 , compared to the histograms in \cref{fig:additional_main_plots,fig:shd_vs_sb_fci_kappa_g}.
	So as a third attempt, we tried FCI with the kernel-based independence test proposed by \citet{zhang2011kernel} and (as we cannot use $\kappa^I$ in a non-linear setting\footnote{The goal of our self-compatibility is to find out if the assumptions of a causal discovery algorithm are violated to the extent that the output of the algorithm is changed non-negligibly. So if we already use an algorithm that does not assume linear dependencies, it is not clear what information we would gain from conducting a test that relies on linearity.}) we only report $\kappa^G$.
	This yielded an incompatibility score of zero.
	We additionally wanted to try the PC algorithm, despite the fact that it assumes causal sufficiency. 
	This again led to a good score, although not as good as the result of FCI with KCI.
    So in this case, the incompatibility scores would have directed us towards the models with the best $F_1$ score and the one with the second best SHD.
    But as the best SHD of 20 (which is still comparably high) shows, the good incompatibility score is not enough to \emph{guarantee} a good performance.
 	\begin{table}[htbp]
		\centering
		\caption{Comparison of causal discovery algorithms on cell dataset}
		\label{tab:comparison}
		\begin{tabular}{@{}lcccc@{}}
			\toprule
			& \textbf{RCD} & \textbf{FCI + Fisher $Z$} & \textbf{FCI + KCI} & \textbf{PC + KCI} \\ \midrule
			$\kappa^I$ & 0.0 & 0.82 & - & - \\
			$\kappa^G$ & 6.65 & 6.5 & 0.0 & 0.68 \\
			SHD & 84 & 62 & 24 & \textbf{20} \\
			Skeleton $F_1$ & 0.5 & 0.49 & \textbf{0.62} & \textbf{0.62} \\ \bottomrule
		\end{tabular}
		\label{tab:results_sachs}
	\end{table}
	
	\label{subsec:additional_real_data}

	\ifSubfilesClassLoaded{
		\bibliographystyle{abbrvnat}
		\bibliography{../aistats}
	}{}
\end{document}

%% file: aistats.bbl
\begin{thebibliography}{60}
\providecommand{\natexlab}[1]{#1}
\providecommand{\url}[1]{\texttt{#1}}
\expandafter\ifx\csname urlstyle\endcsname\relax
  \providecommand{\doi}[1]{doi: #1}\else
  \providecommand{\doi}{doi: \begingroup \urlstyle{rm}\Url}\fi

\bibitem[Bollen and Ting(1993)]{bollen1993confirmatory}
K.~A. Bollen and K.-f. Ting.
\newblock Confirmatory tetrad analysis.
\newblock \emph{Sociological methodology}, pages 147--175, 1993.

\bibitem[Bongers et~al.(2021)Bongers, Forr{\'e}, Peters, and
  Mooij]{bongers2021foundations}
S.~Bongers, P.~Forr{\'e}, J.~Peters, and J.~M. Mooij.
\newblock Foundations of structural causal models with cycles and latent
  variables.
\newblock \emph{The Annals of Statistics}, 49\penalty0 (5):\penalty0
  2885--2915, 2021.

\bibitem[Bousquet and Elisseeff(2002)]{bousquet2002stability}
O.~Bousquet and A.~Elisseeff.
\newblock Stability and generalization.
\newblock \emph{The Journal of Machine Learning Research}, 2:\penalty0
  499--526, 2002.

\bibitem[Bousquet et~al.(2020)Bousquet, Klochkov, and
  Zhivotovskiy]{bousquet2020sharper}
O.~Bousquet, Y.~Klochkov, and N.~Zhivotovskiy.
\newblock Sharper bounds for uniformly stable algorithms.
\newblock In \emph{Conference on Learning Theory}, pages 610--626. PMLR, 2020.

\bibitem[Chickering(2002)]{chickering2002optimal}
D.~M. Chickering.
\newblock Optimal structure identification with greedy search.
\newblock \emph{Journal of machine learning research}, 3\penalty0
  (Nov):\penalty0 507--554, 2002.

\bibitem[Claassen and Heskes(2012)]{claassen2012bayesian}
T.~Claassen and T.~Heskes.
\newblock A bayesian approach to constraint based causal inference.
\newblock \emph{arXiv preprint arXiv:1210.4866}, 2012.

\bibitem[Daley et~al.(2022)Daley, Resch, and Spekkens]{daley2022experimentally}
P.~J. Daley, K.~J. Resch, and R.~W. Spekkens.
\newblock Experimentally adjudicating between different causal accounts of
  bell-inequality violations via statistical model selection.
\newblock \emph{Physical Review A}, 105\penalty0 (4):\penalty0 042220, 2022.

\bibitem[Daniusis et~al.(2010)Daniusis, Janzing, Mooij, Zscheischler, Steudel,
  Zhang, and Sch{\"o}lkopf]{deterministic}
P.~Daniusis, D.~Janzing, J.~M. Mooij, J.~Zscheischler, B.~Steudel, K.~Zhang,
  and B.~Sch{\"o}lkopf.
\newblock Inferring deterministic causal relations.
\newblock In \emph{Proceedings of the 26th Annual Conference on {U}ncertainty
  in {A}rtificial {I}ntelligence ({UAI})}, pages 143--150. AUAI Press, 2010.

\bibitem[Darmois(1953)]{Darmois}
G.~Darmois.
\newblock Analyse g\'en\'erale des liaisons stochastiques.
\newblock \emph{Rev. Inst. Internationale Statist.}, 21:\penalty0 2--8, 1953.

\bibitem[Dawid(2021)]{Dawid2021}
P.~Dawid.
\newblock Decision-theoretic foundations for statistical causality.
\newblock \emph{Journal of Causal Inference}, 9\penalty0 (1):\penalty0 39--77,
  2021.
\newblock \doi{doi:10.1515/jci-2020-0008}.
\newblock URL \url{https://doi.org/10.1515/jci-2020-0008}.

\bibitem[Eulig et~al.(2023)Eulig, Mastakouri, Blöbaum, Hardt, and
  Janzing]{eulig2023falsifying}
E.~Eulig, A.~A. Mastakouri, P.~Blöbaum, M.~Hardt, and D.~Janzing.
\newblock Toward falsifying causal graphs using a permutation-based test, 2023.

\bibitem[Evans and Richardson(2014)]{evans2014markovian}
R.~J. Evans and T.~S. Richardson.
\newblock Markovian acyclic directed mixed graphs for discrete data.
\newblock 2014.

\bibitem[Gettier(1963)]{gettier1963knowledge}
E.~Gettier.
\newblock Is knowledge justified true belief?
\newblock \emph{Analysis}, 23\penalty0 (6):\penalty0 121--123, 1963.

\bibitem[Gresele et~al.(2022)Gresele, Von~K{\"u}gelgen, K{\"u}bler, Kirschbaum,
  Sch{\"o}lkopf, and Janzing]{gresele2022causal}
L.~Gresele, J.~Von~K{\"u}gelgen, J.~K{\"u}bler, E.~Kirschbaum,
  B.~Sch{\"o}lkopf, and D.~Janzing.
\newblock Causal inference through the structural causal marginal problem.
\newblock In \emph{International Conference on Machine Learning}, pages
  7793--7824. PMLR, 2022.

\bibitem[Guo et~al.(2023)Guo, Wildberger, and Sch{\"o}lkopf]{guo2023out}
S.~Guo, J.~Wildberger, and B.~Sch{\"o}lkopf.
\newblock Out-of-variable generalization.
\newblock \emph{arXiv preprint arXiv:2304.07896}, 2023.

\bibitem[Heckerman(1995)]{Heckerman1995b}
D.~Heckerman.
\newblock A {Bayesian} approach to learning causal networks.
\newblock In \emph{Proceedings of the 11th Conference on Uncertainty in
  Artificial Intelligence}, pages 285--295, San Francisco, CA, 1995. Morgan
  Kaufmann Publishers.

\bibitem[Hoyer et~al.(2008{\natexlab{a}})Hoyer, Janzing, Mooij, Peters, and
  Sch{\"o}lkopf]{hoyer2008nonlinear}
P.~Hoyer, D.~Janzing, J.~M. Mooij, J.~Peters, and B.~Sch{\"o}lkopf.
\newblock Nonlinear causal discovery with additive noise models.
\newblock \emph{Advances in neural information processing systems}, 21,
  2008{\natexlab{a}}.

\bibitem[Hoyer et~al.(2008{\natexlab{b}})Hoyer, Shimizu, Kerminen, and
  Palviainen]{hoyer2008estimation}
P.~O. Hoyer, S.~Shimizu, A.~J. Kerminen, and M.~Palviainen.
\newblock Estimation of causal effects using linear non-gaussian causal models
  with hidden variables.
\newblock \emph{International Journal of Approximate Reasoning}, 49\penalty0
  (2):\penalty0 362--378, 2008{\natexlab{b}}.

\bibitem[Huang et~al.(2021)Huang, Kleindessner, Munishkin, Varshney, Guo, and
  Wang]{huang2021benchmarking}
Y.~Huang, M.~Kleindessner, A.~Munishkin, D.~Varshney, P.~Guo, and J.~Wang.
\newblock Benchmarking of data-driven causality discovery approaches in the
  interactions of arctic sea ice and atmosphere.
\newblock \emph{Frontiers in big Data}, 4:\penalty0 642182, 2021.

\bibitem[Imbens(2020)]{Imbens2020}
G.~W. Imbens.
\newblock Potential outcome and directed acyclic graph approaches to causality:
  Relevance for empirical practice in economics.
\newblock \emph{Journal of Economic Literature}, 58\penalty0 (4):\penalty0
  1129--79, December 2020.

\bibitem[Janzing(2019)]{misconceptions}
D.~Janzing.
\newblock The cause-effect problem: Motivation, ideas, and popular
  misconceptions.
\newblock In I.~Guyon, R.~Statnikov, and B.~Bakir~Batu, editors, \emph{Cause
  Effect Pairs in Machine Learning}, pages 3--26. Springer, 2019.

\bibitem[Janzing et~al.(2023)Janzing, Faller, and
  Vankadara]{janzing2023reinterpreting}
D.~Janzing, P.~M. Faller, and L.~C. Vankadara.
\newblock Reinterpreting causal discovery as the task of predicting unobserved
  joint statistics.
\newblock \emph{arXiv preprint arXiv:2305.06894}, 2023.

\bibitem[Lam et~al.(2022)Lam, Andrews, and Ramsey]{lam2022greedy}
W.-Y. Lam, B.~Andrews, and J.~Ramsey.
\newblock Greedy relaxations of the sparsest permutation algorithm.
\newblock In J.~Cussens and K.~Zhang, editors, \emph{Proceedings of the
  Thirty-Eighth Conference on Uncertainty in Artificial Intelligence}, volume
  180 of \emph{Proceedings of Machine Learning Research}, pages 1052--1062.
  PMLR, 01--05 Aug 2022.

\bibitem[Lu and White(2014)]{Lu2014}
X.~Lu and H.~White.
\newblock Robustness checks and robustness tests in applied economics.
\newblock \emph{Journal of Econometrics}, 178:\penalty0 194--206, 2014.
\newblock Annals Issue: Misspecification Test Methods in Econometrics.

\bibitem[Maeda and Shimizu(2020)]{maeda2020rcd}
T.~N. Maeda and S.~Shimizu.
\newblock Rcd: Repetitive causal discovery of linear non-gaussian acyclic
  models with latent confounders.
\newblock In \emph{International Conference on Artificial Intelligence and
  Statistics}, pages 735--745. PMLR, 2020.

\bibitem[Marx and Vreeken(2017)]{Marx2017}
A.~Marx and J.~Vreeken.
\newblock Telling cause from effect using mdl-based local and global
  regression.
\newblock In \emph{2017 {IEEE} International Conference on Data Mining, {ICDM}
  2017, New Orleans, LA, USA, November 18-21, 2017}, pages 307--316, 2017.

\bibitem[Mukherjee et~al.(2006)Mukherjee, Niyogi, Poggio, and
  Rifkin]{mukherjee2006learning}
S.~Mukherjee, P.~Niyogi, T.~Poggio, and R.~Rifkin.
\newblock Learning theory: stability is sufficient for generalization and
  necessary and sufficient for consistency of empirical risk minimization.
\newblock \emph{Advances in Computational Mathematics}, 25:\penalty0 161--193,
  2006.

\bibitem[Oster(2019)]{Oster2017}
E.~Oster.
\newblock Unobservable selection and coefficient stability: Theory and
  evidence.
\newblock \emph{Journal of Business \& Economic Statistics}, 37\penalty0
  (2):\penalty0 187--204, 2019.

\bibitem[Pearl(2009)]{pearl2009causality}
J.~Pearl.
\newblock \emph{Causality}.
\newblock Cambridge university press, 2009.

\bibitem[Pearl and Mackenzie(2018)]{Pearl2018}
J.~Pearl and J.~Mackenzie.
\newblock \emph{The book of why}.
\newblock Basic Books, USA, 2018.

\bibitem[Pearl and Verma(1995)]{pearl1995theory}
J.~Pearl and T.~S. Verma.
\newblock A theory of inferred causation.
\newblock In \emph{Studies in Logic and the Foundations of Mathematics}, volume
  134, pages 789--811. Elsevier, 1995.

\bibitem[Perkovi{\'c} et~al.(2015)Perkovi{\'c}, Textor, Kalisch, and
  Maathuis]{perkovic2015complete}
E.~Perkovi{\'c}, J.~Textor, M.~Kalisch, and M.~H. Maathuis.
\newblock A complete generalized adjustment criterion.
\newblock In \emph{Proceedings of the Thirty-First Conference on Uncertainty in
  Artificial Intelligence}, pages 682--691, 2015.

\bibitem[Peters and B{\"u}hlmann(2015)]{peters2015structural}
J.~Peters and P.~B{\"u}hlmann.
\newblock Structural intervention distance for evaluating causal graphs.
\newblock \emph{Neural computation}, 27\penalty0 (3):\penalty0 771--799, 2015.

\bibitem[Peters et~al.(2017)Peters, Janzing, and
  Sch{\"o}lkopf]{peters2017elements}
J.~Peters, D.~Janzing, and B.~Sch{\"o}lkopf.
\newblock \emph{Elements of causal inference: foundations and learning
  algorithms}.
\newblock The MIT Press, 2017.

\bibitem[Popper(1959)]{Popper1959}
K.~Popper.
\newblock \emph{The logic of scientific discovery}.
\newblock Routledge, London, 1959.

\bibitem[Ramsey et~al.(2012)Ramsey, Zhang, and Spirtes]{ramsey2012adjacency}
J.~Ramsey, J.~Zhang, and P.~L. Spirtes.
\newblock Adjacency-faithfulness and conservative causal inference.
\newblock \emph{arXiv preprint arXiv:1206.6843}, 2012.

\bibitem[Reisach et~al.(2021)Reisach, Seiler, and Weichwald]{Reisach2021}
A.~Reisach, C.~Seiler, and S.~Weichwald.
\newblock Beware of the simulated dag! causal discovery benchmarks may be easy
  to game.
\newblock In M.~Ranzato, A.~Beygelzimer, Y.~Dauphin, P.~Liang, and J.~W.
  Vaughan, editors, \emph{Advances in Neural Information Processing Systems},
  volume~34, pages 27772--27784. Curran Associates, Inc., 2021.

\bibitem[Reynolds et~al.(2022)Reynolds, Scott, Turner, Iwaniec, Bouxsein,
  Sanders, and Antonsen]{reynolds2022validating}
R.~J. Reynolds, R.~T. Scott, R.~T. Turner, U.~T. Iwaniec, M.~L. Bouxsein, L.~M.
  Sanders, and E.~L. Antonsen.
\newblock Validating causal diagrams of human health risks for spaceflight: An
  example using bone data from rodents.
\newblock \emph{Biomedicines}, 10\penalty0 (9):\penalty0 2187, 2022.

\bibitem[Richardson(2003)]{richardson2003markov}
T.~Richardson.
\newblock Markov properties for acyclic directed mixed graphs.
\newblock \emph{Scandinavian Journal of Statistics}, 30\penalty0 (1):\penalty0
  145--157, 2003.

\bibitem[Richardson et~al.(2023)Richardson, Evans, Robins, and
  Shpitser]{richardson2023nested}
T.~S. Richardson, R.~J. Evans, J.~M. Robins, and I.~Shpitser.
\newblock Nested markov properties for acyclic directed mixed graphs.
\newblock \emph{The Annals of Statistics}, 51\penalty0 (1):\penalty0 334--361,
  2023.

\bibitem[Sachs et~al.(2005)Sachs, Perez, Pe'er, Lauffenburger, and
  Nolan]{sachs2005causal}
K.~Sachs, O.~Perez, D.~Pe'er, D.~A. Lauffenburger, and G.~P. Nolan.
\newblock Causal protein-signaling networks derived from multiparameter
  single-cell data.
\newblock \emph{Science}, 308\penalty0 (5721):\penalty0 523--529, 2005.

\bibitem[Shalev-Shwartz et~al.(2010)Shalev-Shwartz, Shamir, Srebro, and
  Sridharan]{shalev2010learnability}
S.~Shalev-Shwartz, O.~Shamir, N.~Srebro, and K.~Sridharan.
\newblock Learnability, stability and uniform convergence.
\newblock \emph{The Journal of Machine Learning Research}, 11:\penalty0
  2635--2670, 2010.

\bibitem[Shimizu et~al.(2006)Shimizu, Hoyer, Hyv{\"a}rinen, Kerminen, and
  Jordan]{shimizu2006linear}
S.~Shimizu, P.~O. Hoyer, A.~Hyv{\"a}rinen, A.~Kerminen, and M.~Jordan.
\newblock A linear non-gaussian acyclic model for causal discovery.
\newblock \emph{Journal of Machine Learning Research}, 7\penalty0 (10), 2006.

\bibitem[Shimizu et~al.(2011)Shimizu, Inazumi, Sogawa, Hyvarinen, Kawahara,
  Washio, Hoyer, Bollen, and Hoyer]{shimizu2011directlingam}
S.~Shimizu, T.~Inazumi, Y.~Sogawa, A.~Hyvarinen, Y.~Kawahara, T.~Washio, P.~O.
  Hoyer, K.~Bollen, and P.~Hoyer.
\newblock Directlingam: A direct method for learning a linear non-gaussian
  structural equation model.
\newblock \emph{Journal of Machine Learning Research-JMLR}, 12\penalty0
  (Apr):\penalty0 1225--1248, 2011.

\bibitem[Skitovic(1962)]{Skitovic}
V.~Skitovic.
\newblock Linear combinations of independent random variables and the normal
  distribution law.
\newblock \emph{Select. Transl. Math. Stat. Probab.}, \penalty0 (2):\penalty0
  211--228, 1962.

\bibitem[Spirtes and Scheines(2004)]{spirtes2004}
P.~Spirtes and R.~Scheines.
\newblock Causal inference of ambiguous manipulations.
\newblock \emph{Philosophy of Science}, 71\penalty0 (5):\penalty0 833–845,
  2004.
\newblock \doi{10.1086/425058}.

\bibitem[Spirtes et~al.(2000)Spirtes, Glymour, Scheines, and
  Heckerman]{spirtes2000causation}
P.~Spirtes, C.~N. Glymour, R.~Scheines, and D.~Heckerman.
\newblock \emph{Causation, prediction, and search}.
\newblock MIT press, 2000.

\bibitem[Strobl et~al.(2016)Strobl, Spirtes, and
  Visweswaran]{strobl2016estimating}
E.~V. Strobl, P.~L. Spirtes, and S.~Visweswaran.
\newblock Estimating and controlling the false discovery rate for the pc
  algorithm using edge-specific p-values.
\newblock \emph{arXiv preprint arXiv:1607.03975}, 2016.

\bibitem[Su and Henckel(2022)]{su2022robustness}
Z.~Su and L.~Henckel.
\newblock A robustness test for estimating total effects with covariate
  adjustment.
\newblock In \emph{Uncertainty in Artificial Intelligence}, pages 1886--1895.
  PMLR, 2022.

\bibitem[Textor et~al.(2017)Textor, van~der Zander, Gilthorpe, Liśkiewicz, and
  Ellison]{textor2017robust}
J.~Textor, B.~van~der Zander, M.~S. Gilthorpe, M.~Liśkiewicz, and G.~T.
  Ellison.
\newblock {Robust causal inference using directed acyclic graphs: the R package
  ‘dagitty’}.
\newblock \emph{International Journal of Epidemiology}, 45\penalty0
  (6):\penalty0 1887--1894, 01 2017.
\newblock ISSN 0300-5771.
\newblock \doi{10.1093/ije/dyw341}.
\newblock URL \url{https://doi.org/10.1093/ije/dyw341}.

\bibitem[Tian and Pearl(2002)]{tian2002general}
J.~Tian and J.~Pearl.
\newblock A general identification condition for causal effects.
\newblock In \emph{Aaai/iaai}, pages 567--573, 2002.

\bibitem[Triantafillou and Tsamardinos(2016)]{triantafillou2016score}
S.~Triantafillou and I.~Tsamardinos.
\newblock Score-based vs constraint-based causal learning in the presence of
  confounders.
\newblock In \emph{Cfa@ uai}, pages 59--67, 2016.

\bibitem[Tsamardinos et~al.(2012)Tsamardinos, Triantafillou, and
  Lagani]{tsamardinos2012towards}
I.~Tsamardinos, S.~Triantafillou, and V.~Lagani.
\newblock Towards integrative causal analysis of heterogeneous data sets and
  studies.
\newblock \emph{The Journal of Machine Learning Research}, 13\penalty0
  (1):\penalty0 1097--1157, 2012.

\bibitem[Verma and Pearl(2022)]{verma2022equivalence}
T.~S. Verma and J.~Pearl.
\newblock Equivalence and synthesis of causal models.
\newblock In \emph{Probabilistic and Causal Inference: The Works of Judea
  Pearl}, pages 221--236. 2022.

\bibitem[Vorob'ev(1962)]{Vorobev1962}
N.~Vorob'ev.
\newblock Consistent families of measures and their extensions.
\newblock \emph{Theory Probab. Appl}, 7\penalty0 (2):\penalty0 147--163, 1962.

\bibitem[Walter and Tiemeier(2009)]{walter2009variable}
S.~Walter and H.~Tiemeier.
\newblock Variable selection: current practice in epidemiological studies.
\newblock \emph{European journal of epidemiology}, 24:\penalty0 733--736, 2009.

\bibitem[Zhang(2008)]{zhang2008causal}
J.~Zhang.
\newblock Causal reasoning with ancestral graphs.
\newblock \emph{Journal of Machine Learning Research}, 9:\penalty0 1437--1474,
  2008.

\bibitem[Zhang and Hyv\"arinen(2009)]{Zhang_UAI}
K.~Zhang and A.~Hyv\"arinen.
\newblock On the identifiability of the post-nonlinear causal model.
\newblock In \emph{Proceedings of the 25th Conference on Uncertainty in
  Artificial Intelligence}, Montreal, Canada, 2009.

\bibitem[Zhang et~al.(2011)Zhang, Peters, Janzing, and
  Sch{\"o}lkopf]{zhang2011kernel}
K.~Zhang, J.~Peters, D.~Janzing, and B.~Sch{\"o}lkopf.
\newblock Kernel-based conditional independence test and application in causal
  discovery.
\newblock In \emph{27th Conference on Uncertainty in Artificial Intelligence
  (UAI 2011)}, pages 804--813. AUAI Press, 2011.

\bibitem[Zheng et~al.(2023)Zheng, Huang, Chen, Ramsey, Gong, Cai, Shimizu,
  Spirtes, and Zhang]{causallearn}
Y.~Zheng, B.~Huang, W.~Chen, J.~Ramsey, M.~Gong, R.~Cai, S.~Shimizu,
  P.~Spirtes, and K.~Zhang.
\newblock Causal-learn: Causal discovery in python.
\newblock \emph{arXiv preprint arXiv:2307.16405}, 2023.

\end{thebibliography}
